\documentclass{article}
\usepackage{fullpage}
\usepackage[svgnames,dvipsnames,color,table]{xcolor}
\usepackage{hyperref}
\hypersetup{colorlinks,linkcolor={blue},citecolor={blue},urlcolor={RoyalBlue}} 

\usepackage{amsthm}
\usepackage{amsmath} %
\usepackage{nicefrac}
\usepackage{bm}
\usepackage{booktabs}
\usepackage{amsfonts}
\usepackage{bbm}
\usepackage{mleftright}
\usepackage{dsfont}
\usepackage{amscd,amssymb,amsmath,amsthm,bbold}
\usepackage{graphicx}
\usepackage{float}
\usepackage{multirow}
\usepackage[numbers]{natbib}

\usepackage{pifont}
\usepackage{microtype}
\usepackage{enumitem}
\usepackage{xfrac}
\usepackage{parskip}
\usepackage{numprint}
\usepackage{booktabs} %

\usepackage{dblfloatfix}
\usepackage{transparent}

\usepackage{algorithm}
\usepackage{listings}

\usepackage[textsize=scriptsize,textwidth=1.9cm]{todonotes}
\usepackage{xspace}
\usepackage{subcaption}
\usepackage{etoolbox}
\usepackage{comment}
\usepackage{etoc}
\usepackage{wrapfig}
\usepackage{placeins}
\usepackage{makecell}
\usepackage{lipsum,xcolor}
\newcommand\dunderline[3][-1pt]{{%
  \sbox0{#3}%
  \ooalign{\copy0\cr\rule[\dimexpr#1-#2\relax]{\wd0}{#2}}}}
\newcommand{\customfootnotetext}[2]{{%
  \renewcommand{\thefootnote}{#1}%
  \footnotetext[0]{#2}}}%

\DeclareMathOperator*{\argmax}{arg\,max}
\DeclareMathOperator*{\argmin}{arg\,min}

\usepackage{cleveref}
\crefformat{section}{\S#2#1#3} %
\crefformat{subsection}{\S#2#1#3}
\crefformat{subsubsection}{\S#2#1#3}

\usepackage{adjustbox}
\usepackage{array}

\newcolumntype{R}[2]{%
    >{\adjustbox{angle=#1,lap=\width-(#2)}\bgroup}%
    l%
    <{\egroup}%
}

\begin{document}

\date{}
\title{Robust fine-tuning of zero-shot models}
\author{
        Mitchell Wortsman$^{*\dag}$ \\
        \and 
        Gabriel Ilharco$^{*\dag}$\\
        \and 
        Jong Wook Kim$^\S$\\
        \and 
        Mike Li$^{\ddag}$\\
        \and 
        Simon Kornblith$^\diamond$ \\
        \and 
        Rebecca Roelofs$^\diamond$ \\
        \and 
        Raphael Gontijo-Lopes$^\diamond$ \\
        \and 
        \setcounter{footnote}{8}
        Hannaneh Hajishirzi$^{\dag\circ}$ \hspace*{-0.25cm}\\
        \and 
        Ali Farhadi$^{\star\dag}$ \hspace*{-0.25cm}\\
        \and 
        Hongseok Namkoong$^{\star\ddag}$ \hspace*{-0.25cm}\\
        \and 
        Ludwig Schmidt$^{\dag\vartriangle}$ %
}

\maketitle
\customfootnotetext{$*\star$}{These authors contributed equally.}

\customfootnotetext{$\dag$}
{University of Washington\enspace
$^\S$OpenAI\enspace
$^\ddag$Columbia University\enspace
$^\diamond$Google Research, Brain Team\enspace

\enspace\hspace{-0.05cm}$^\circ$Allen Institute for Artificial Intelligence\enspace
$^\vartriangle$Toyota Research Institute

\enspace Code provided at \url{https://github.com/mlfoundations/wise-ft}.\vspace*{-0.5cm}}

\newif\ifcomments
\commentsfalse

\ifcomments
    \newcommand\gamaga[1]{{\todo[color=teal]{GI: {#1}}}}
    \newcommand\ali[1]{{\todo[color=olive]{AF: {#1}}}}
    \newcommand\hanna[1]{{\todo[color=yellow]{HH: {#1}}}}
    \newcommand\mw[1]{{\todo[color=purple]{MW: {#1}}}}
    \newcommand\ludwig[1]{{\todo[color=orange]{LS: {#1}}}}
    \newcommand\hong[1]{{\todo[color=green]{HN: {#1}}}}
    \newcommand\mike[1]{{\todo[color=blue]{ML: {#1}}}}
\else
    \providecommand{\mw}[1]{}
    \providecommand{\gamaga}[1]{}
    \providecommand{\ali}[1]{}
    \providecommand{\hanna}[1]{}
    \providecommand{\ludwig}[1]{}
    \providecommand{\hong}[1]{}
    \providecommand{\mike}[1]{}
\fi

\newcommand\ALPHA{mixing coefficient}

\makeatletter
\renewcommand\paragraph{\@startsection{paragraph}{4}{\z@}                                     {1.05ex \@plus1ex \@minus.2ex}                                {-.5em}
{\normalfont\normalsize\bfseries}}
\makeatother
\newcommand{\smallpara}[1]{\textbf{#1}}
\newcommand{\beforesec}{\vspace*{0em}}
\newcommand{\postsec}{\vspace*{0em}}

\vspace*{-1cm}
\begin{abstract}
Large pre-trained models such as CLIP or ALIGN offer consistent accuracy across a range of data distributions when performing zero-shot inference (i.e., without fine-tuning on a specific dataset).  
Although existing fine-tuning methods substantially improve accuracy on a given target distribution, they often reduce robustness to distribution shifts.  
We address this tension by introducing a simple and effective method for improving robustness while fine-tuning:  ensembling the weights of the zero-shot and fine-tuned models (WiSE-FT). 
Compared to standard fine-tuning, WiSE-FT provides large accuracy improvements under distribution shift, while  preserving high accuracy on the target distribution.  
On ImageNet and five derived distribution shifts, WiSE-FT improves accuracy under distribution shift by 4 to 6 percentage points (pp) over prior work while increasing ImageNet accuracy by 1.6 pp. 
WiSE-FT achieves similarly large robustness gains (2 to 23 pp) on a diverse set of six further distribution shifts, and accuracy gains of 0.8 to 3.3 pp compared to standard fine-tuning on seven commonly used transfer learning datasets. 
These improvements come at no additional computational cost during fine-tuning or inference.
\end{abstract}

\setcounter{footnote}{0}
\section{Introduction}
\postsec

A foundational goal of machine learning is to develop models that work reliably across a broad range of data distributions.
Over the past few years, researchers have proposed a variety of distribution shifts on which current algorithmic approaches to enhance robustness yield little to no gains \cite{taori2020measuring, miller21b}.
While these negative results highlight the difficulty of learning robust models, large pre-trained models such as CLIP \cite{radford2021learning}, ALIGN \cite{jia2021scaling} and BASIC \cite{pham2021scaling} have recently demonstrated unprecedented robustness to these challenging distribution shifts.
The success of these models points towards pre-training on large, heterogeneous datasets as a promising direction for increasing robustness.
However, an important caveat is that these robustness improvements are largest in the zero-shot setting, i.e., when the model performs inference without fine-tuning on a specific target distribution.

In a concrete application, a zero-shot model can be fine-tuned on extra application-specific data, which often yields large performance gains on the target distribution.
However, in the experiments of \citet{radford2021learning} and \citet{pham2021scaling}, fine-tuning comes at the cost of robustness: across several natural distribution shifts, the accuracy of their fine-tuned models is lower than that of the original zero-shot model.
This leads to a natural question:
\begin{center}
    \emph{Can zero-shot models be fine-tuned without reducing accuracy under distribution shift?}
\end{center}

\begin{figure*}
    \centering
    \includegraphics[width=\textwidth]{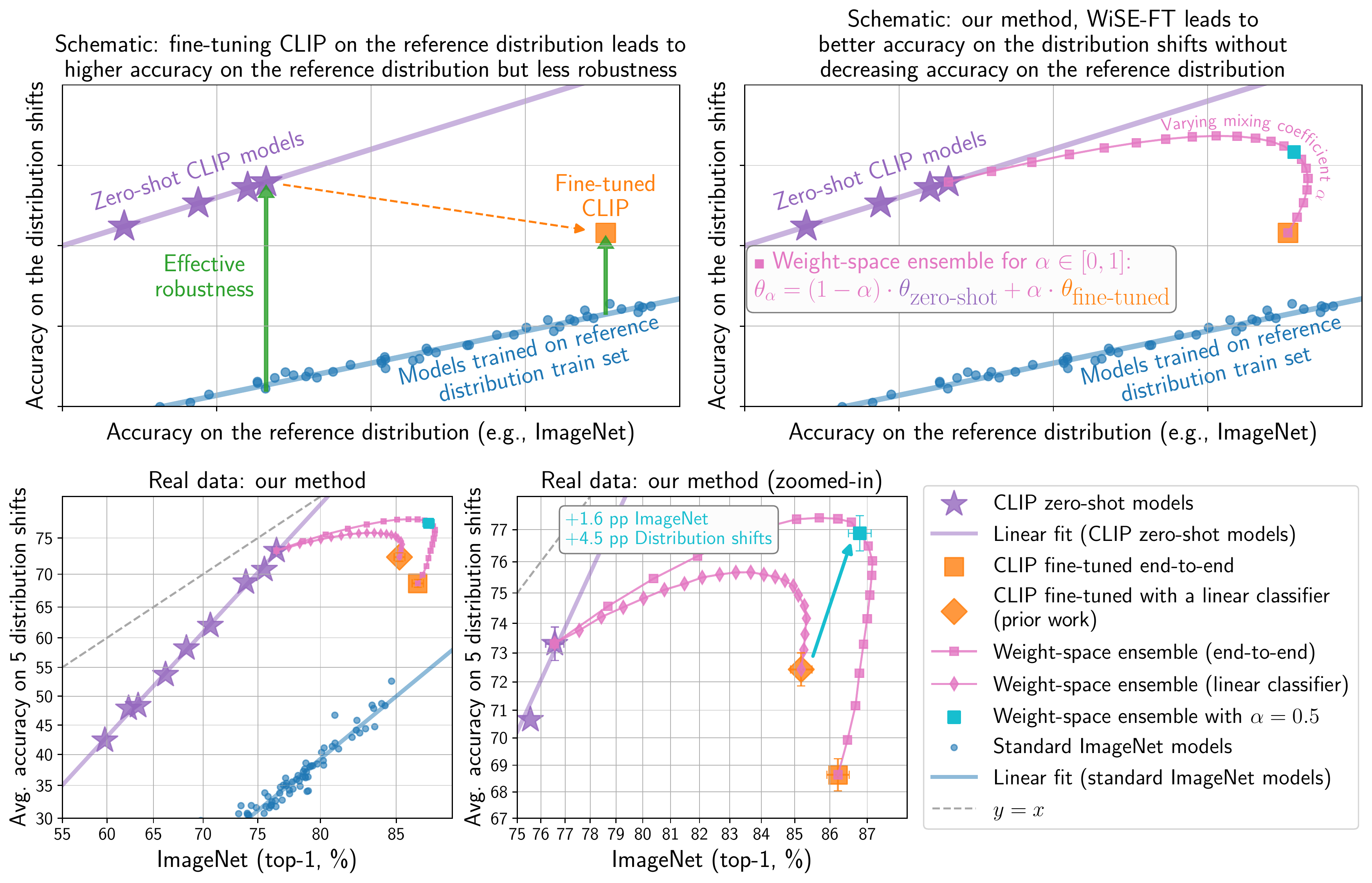}
    \caption{
    \textbf{(Top left)} Zero-shot CLIP models exhibit moderate accuracy on the reference distribution ($x$-axis, the target for fine-tuning) and high effective robustness (accuracy on the distribution shifts beyond the baseline models).
    In contrast, standard fine-tuning---either end-to-end or with a linear classifier (final layer)---attains higher accuracy on the reference distribution but less effective robustness.
    \textbf{(Top right)} Our method linearly interpolates between the zero-shot and fine-tuned models with a mixing coefficient $\alpha \in [0,1]$. \textbf{(Bottom)} On five distribution shifts derived from ImageNet (ImageNetV2, ImageNet-R, ImageNet Sketch, ObjectNet, and ImageNet-A), WiSE-FT improves average accuracy relative to both the zero-shot and fine-tuned models while maintaining or improving accuracy on ImageNet.
    }
    \label{fig:fig1}
\end{figure*}

As pre-trained models are becoming a cornerstone of machine learning, techniques for fine-tuning them on downstream applications are increasingly important.
Indeed, the question of robustly fine-tuning pre-trained models has recently also been raised as an open problem by several authors 
\cite{andreassen2021evolution,bommasani2021opportunities,radford2021learning,pham2021scaling}.
\citet{andreassen2021evolution} explored several fine-tuning approaches but found that none yielded models with improved robustness at high accuracy.
Furthermore, \citet{taori2020measuring} demonstrated that no current algorithmic robustness interventions provide consistent gains across the distribution shifts where zero-shot models excel.

In this paper, we conduct an empirical investigation to understand and improve fine-tuning of zero-shot models from a distributional robustness perspective.
We begin by measuring how different fine-tuning approaches (last-layer vs.\ end-to-end fine-tuning, hyperparameter changes, etc.) affect the accuracy under distribution shift of the resulting fine-tuned models.
Our empirical analysis uncovers two key issues in the standard fine-tuning process.
First, the robustness of fine-tuned models varies substantially under even small changes in hyperparameters, but the best hyperparameters cannot be inferred from accuracy on the target distribution alone.
Second, more aggressive fine-tuning (e.g., using a larger learning rate) yields larger accuracy improvements on the target distribution, but can also reduce accuracy under distribution shift by a large amount.

Motivated by the above concerns, we propose a robust way of fine-tuning zero-shot models that addresses the aforementioned trade-off and achieves the best of both worlds: increased performance under distribution shift while maintaining or even improving accuracy on the target distribution relative to standard fine-tuning.
In addition, our method simplifies the choice of hyperparameters in the fine-tuning process.

Our method (Figure \ref{fig:fig1}) has two steps: first, we fine-tune the zero-shot model on the target distribution.
Second, we combine the original zero-shot and fine-tuned models by linearly interpolating between their weights, which we refer to as weight-space ensembling.
Interpolating model parameters is a classical idea in convex optimization dating back decades (e.g., see~\cite{ruppert1988efficient, polyak1990new}).
Here, we empirically study model interpolation for non-convex models from the perspective of distributional robustness.
Interestingly, linear interpolation in weight-space still succeeds despite the non-linearity in the activation functions of the neural networks.

Weight-space ensembles for fine-tuning (WiSE-FT) substantially improve accuracy under distribution shift compared to prior work while maintaining high performance on the target distribution.
Concretely, on ImageNet \cite{deng2009imagenet} and five of the natural distribution shifts studied by \citet{radford2021learning}, WiSE-FT applied to standard end-to-end fine-tuning improves accuracy under distribution shift by 4 to 6 percentage points (pp) over prior work while maintaining or improving the ImageNet accuracy of the fine-tuned CLIP model.
Relative to the zero-shot model, WiSE-FT improves accuracy under distribution shift by 1 to 9 pp.
Moreover, WiSE-FT improves over a range of alternative approaches such as regularization and evaluating at various points throughout fine-tuning.
These robustness gains come at no additional computational cost during fine-tuning or inference.

While our investigation centers around CLIP, we observe similar trends for other zero-shot models including ALIGN~\cite{jia2021scaling}, BASIC~\cite{pham2021scaling}, and a ViT model pre-trained on JFT~\cite{dosovitskiy2021an}.
For instance, WiSE-FT improves the ImageNet accuracy of a fine-tuned BASIC-L model by 0.4 pp, while improving average accuracy under distribution shift by 2 to 11 pp.

To understand the robustness gains of WiSE-FT, we first study WiSE-FT when fine-tuning a linear classifier (last layer) as it is more amenable to analysis. In this linear case, our procedure is equivalent to ensembling the outputs of two models, and experiments point towards the complementarity of model predictions as a key property.
For end-to-end fine-tuning, we connect our observations  to earlier work on the phenomenology of deep learning.
\citet{neyshabur2020being} found that end-to-end fine-tuning the same model twice yielded two different solutions that were connected via a linear path in weight-space along which error remains low, known as linear mode connectivity \cite{frankle2020linear}.
Our observations suggest a similar phenomenon along the path generated by WiSE-FT, but the exact shape of the loss landscape and connection between error on the target and shifted distributions are still open problems.

In addition to the aforementioned ImageNet distribution shifts, WiSE-FT consistently improves robustness on a diverse set of six additional distribution shifts including:
(i) geographic shifts in satellite imagery and wildlife recognition (WILDS-FMoW, WILDS-iWildCam) \cite{wilds2021, christie2018functional, beery2021iwildcam},
(ii) reproductions of the popular image classification dataset CIFAR-10 with a distribution shift (CIFAR-10.1 and CIFAR-10.2)  \cite{pmlr-v97-recht19a, lu2020harder}, and
(iii) datasets with distribution shift induced by temporal perturbations in videos (ImageNet-Vid-Robust and YTBB-Robust) \cite{vidrobust}.
Beyond the robustness perspective, WiSE-FT also improves accuracy compared to standard fine-tuning, reducing the relative error rate by 4-49\% on a range of seven datasets: ImageNet, CIFAR-10, CIFAR-100 \cite{krizhevsky2009learning}, Describable Textures \cite{dtd}, Food-101 \cite{food101}, SUN397 \cite{sun397}, and Stanford Cars \cite{cars}.
Even when fine-tuning data is scarce, reflecting many application scenarios, we find that WiSE-FT improves performance.

Overall, WiSE-FT is simple, universally applicable in the problems we studied, and can be implemented in a few lines of code. 
Hence we encourage its adoption for fine-tuning zero-shot models.

 \begin{figure*}[t]
    \centering
    \includegraphics[width=0.9\textwidth]{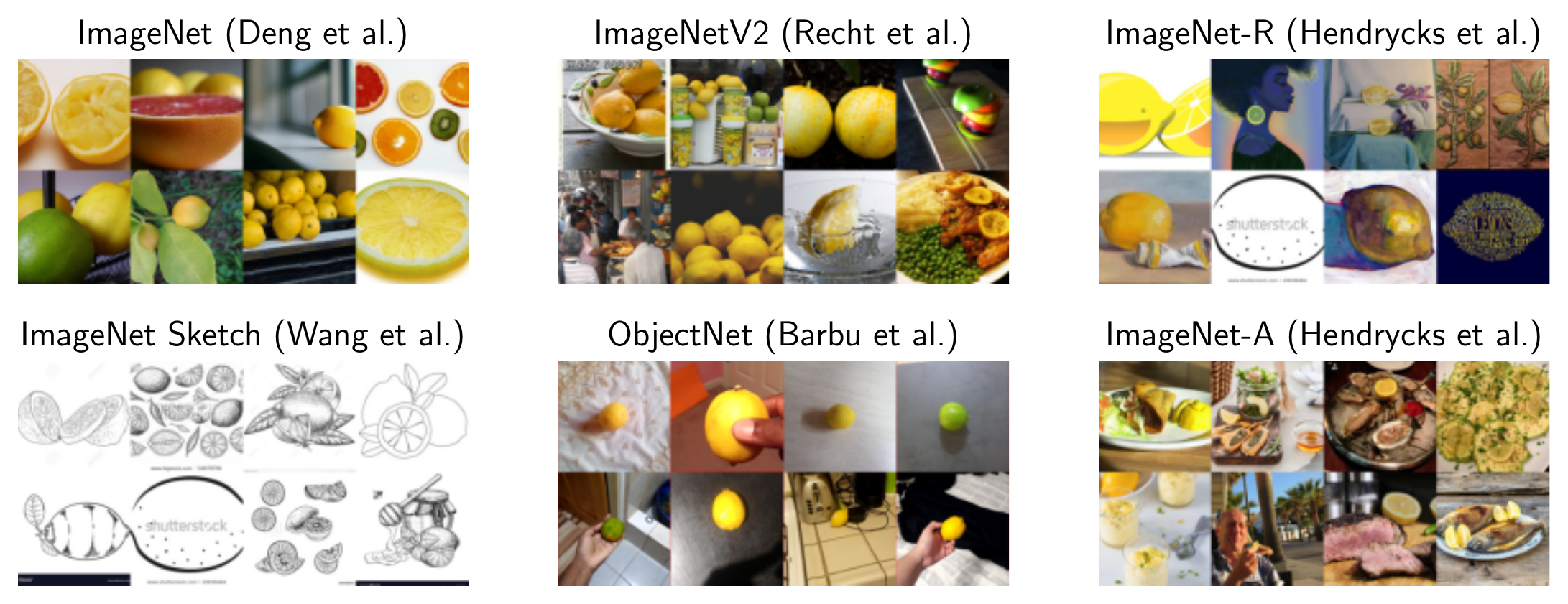}
    \caption{Samples of the class \textit{lemon}, from the reference distribution ImageNet \cite{deng2009imagenet} and the derived distribution shifts considered in our main experiments: ImageNet-V2 \cite{pmlr-v97-recht19a}, ImageNet-R \cite{imagenetr}, ImageNet Sketch \cite{imagenetsketch}, ObjectNet \cite{objectnet}, and ImageNet-A \cite{imageneta}.}
    \label{fig:data_samples}
\end{figure*}
\beforesec
\section{Background and experimental setup}\label{sec:expsetup}
\postsec

Our experiments compare the performance of zero-shot models, corresponding fine-tuned models, and models produced by WiSE-FT. 
To measure robustness, we contrast model accuracy on two related but different distributions, a reference distribution $\mathcal{D}_{\mathrm{ref}}$ which is the target for fine-tuning, and shifted distribution $\mathcal{D}_{\mathrm{shift}}$.\footnote{$\mathcal{D}_{\mathrm{ref}}$ and $\mathcal{D}_{\mathrm{shift}}$ are sometimes referred to as \textit{in-distribution} (ID) and \textit{out-of-distribution} (OOD). In this work, we include evaluations of zero-shot models, which are \textit{not} trained on data from the reference distribution, so referring to $\mathcal{D}_{\mathrm{ref}}$ would be imprecise. For clarity, we avoid the ID/OOD terminology.}
We assume both distributions have test sets for evaluation, and $\mathcal{D}_{\mathrm{ref}}$ has an associated training set $\mathcal{S}_{\mathrm{ref}}^\text{tr}$ which is typically used for training or fine-tuning.
The goal for a model is to achieve both high accuracy and consistent performance on the two distributions $\mathcal{D}_{\mathrm{ref}}$ and $\mathcal{D}_{\mathrm{shift}}$.
This is a natural goal as humans often achieve similar accuracy across the distribution shifts in our study \cite{shankar2020evaluating}.

For a model $f$, we let $\mathsf{Acc}_{\mathrm{ref}}(f)$ and $\mathsf{Acc}_{\mathrm{shift}}(f)$ refer to classification accuracy on the reference and shifted test sets, respectively. 
We consider $k$-way image classification, where $x_i$ is an image with corresponding label $y_i \in \{1,...,k\}$. 
The outputs of $f$ are $k$-dimensional vectors of non-normalized class scores.

\smallpara{Distribution shifts.}
\citet{taori2020measuring} categorized distribution shifts into two broad categories:
(i) \emph{synthetic}, e.g., $\ell_\infty$-adversarial examples or artificial changes in image contrast, brightness, etc. \cite{imagenetc,biggio2013evasion,biggio2018wild,geirhos2018generalisation,alcorn2019strike}; and 
(ii) \emph{natural}, where samples are not perturbed after acquisition and changes in data distributions arise through naturally occurring variations in lighting, geographic location, crowdsourcing process, image styles, etc. \cite{taori2020measuring, pmlr-v97-recht19a, imagenetr, imageneta, wilds2021}.
Following \citet{radford2019language}, our focus here is on natural distribution shifts as they are more representative of the real world when no active adversary is present. 
Specifically, we present our key results for five natural distribution shifts derived from ImageNet (i.e., $\mathcal{S}^\text{tr}_\text{ref}$ is ImageNet):
\begin{itemize}
 \item
ImageNet-V2 (IN-V2) \cite{pmlr-v97-recht19a}, a reproduction of the ImageNet test set with distribution shift
\item ImageNet-R (IN-R) \cite{imagenetr}, renditions (e.g., sculptures, paintings) for 200 ImageNet classes
\item ImageNet Sketch (IN-Sketch) \cite{imagenetsketch}, which contains sketches instead of natural images
\item ObjectNet \cite{objectnet}, a test set of objects in various scenes with 113 classes overlapping with ImageNet
\item ImageNet-A (IN-A) \cite{imageneta}, a test set of natural images misclassified by a ResNet-50 \cite{he2016deep} for 200 ImageNet classes.
\end{itemize}
Figure \ref{fig:data_samples} illustrates the five distribution shifts.

\smallpara{Effective robustness and scatter plots.}
To compare the robustness of models with different accuracies on the reference distribution, we follow the \emph{effective robustness} framework introduced by \citet{taori2020measuring}.
Effective robustness quantifies robustness as accuracy \emph{beyond a baseline} trained only on the reference distribution.
A useful tool for studying (effective) robustness are scatter plots that illustrate model performance under distribution shift  \cite{pmlr-v97-recht19a, taori2020measuring}. 
These scatter plots display accuracy on the reference distribution on the $x$-axis and accuracy under distribution shift on the $y$-axis, i.e., a model $f$ is shown as a point $\left(\mathsf{Acc}_{\mathrm{ref}}(f),\ \mathsf{Acc}_{\mathrm{shift}}(f)\right)$.
Figure \ref{fig:fig1} exemplifies these scatter plots with both schematics and real data.
For the distribution shifts we study, accuracy on the reference distribution is a reliable predictor of accuracy under distribution shift \cite{taori2020measuring, miller21b}. 
In other words, there exists a function $\beta : [0,1] \rightarrow [0,1]$ such that $\mathsf{Acc}_\mathrm{shift}\mleft(f\mright)$ approximately equals $\beta\mleft(\mathsf{Acc}_\mathrm{ref}\mleft(f\mright)\mright)$ for models $f$ trained on the train set $\mathcal{S}^\text{tr}_\mathrm{ref}$.  
Effective robustness \cite{taori2020measuring} is accuracy beyond this baseline, defined formally as $\rho(f) = \mathsf{Acc}_\mathrm{shift}(f) -\beta\mleft(\mathsf{Acc}_\mathrm{ref}\mleft(f\mright)\mright)$.

In the corresponding scatter plots, effective robustness is vertical movement above expected accuracy under distribution shift (Figure \ref{fig:fig1}, top).
Effective robustness thereby disentangles accuracy changes on the reference distribution from the effect of robustness interventions.
When we say that a model is robust to distribution shift, we mean that effective robustness is positive. \citet{taori2020measuring} observed that no algorithmic robustness intervention consistently achieves substantial effective robustness across the distribution shifts in Figure~\ref{fig:data_samples}---the first method to do so was zero-shot CLIP.
Empirically, when applying logit (or probit) axis scaling, models trained on the reference distribution approximately lie on a linear trend \cite{taori2020measuring,miller21b}. As in \citet{taori2020measuring}, we apply logit axis scaling and show 95\% Clopper-Pearson confidence intervals for the accuracies of select points.

 \smallpara{Zero-shot models and CLIP.} We primarily explore CLIP models \cite{radford2021learning}, although we also investigate other zero-shot models including ALIGN \cite{jia2021scaling}, BASIC \cite{pham2021scaling} and a ViT model pre-trained on JFT \cite{dosovitskiy2021an}. Zero-shot models exhibit effective robustness and lie on a qualitatively different linear trend (Figure~\ref{fig:fig1}). 
CLIP-like models are pre-trained using image-caption pairs from the web.
Given a set of image-caption pairs $\{(x_1,s_1)...,(x_B,s_B)\}$, CLIP-like models train an image-encoder $g$ and text-encoder $h$ such that the similarity $\left\langle g(x_i), h(s_i)\right\rangle$ is maximized relative to unaligned pairs.
CLIP-like models perform zero-shot $k$-way classification given an image $x$ and class names $C = \{c_1,...,c_k\}$ by matching $x$ with potential captions.
For instance, using caption $s_i = \text{``a photo of a \{$c_i$\}''}$ for each class $i$, the zero-shot model predicts the class via $\argmax_j \left\langle g(x), h(s_j) \right\rangle$.\footnote{For improved accuracy, the embedding of a few candidate captions are averaged, e.g., $s_i^{(1)}{=}\text{``a \textit{photo} of a \{$c_i$\}''}$ and $s_i^{(2)}{=}\text{``a \textit{picture} of a \{$c_i$\}''}$ (referred to as prompt ensembling \cite{radford2021learning}).}
Equivalently, one can construct $\textbf{W}_{\text{zero-shot}} \in \mathbb{R}^{d \times k}$ with columns $h(s_j)$ and compute outputs $f(x) = g(x)^\top \textbf{W}_{\text{zero-shot}}$.
Unless explicitly mentioned, our experiments use the CLIP model \texttt{ViT-L/14@336px}, although all CLIP models are displayed in our scatter plots (additional details provided in Appendix~\ref{sec:morezs}).

\beforesec
\section{Weight-space ensembles for fine-tuning} 
\label{sec:main}
\postsec
This section describes and motivates our proposed method, WiSE-FT, which consists of two simple steps.
First, we fine-tune the zero-shot model on application-specific data. 
Second, we combine the original zero-shot and fine-tuned models by linearly interpolating between their weights, also referred to as weight-space ensembling.
WiSE-FT can be implemented in a few lines of PyTorch, and we provide example code in Appendix \ref{sec:pseudo-code}.

The zero-shot model excels under distribution shift while standard fine-tuning achieves high accuracy on the reference distribution.
Our motivation is to combine these two models into one that achieves the best of both worlds.
Weight-space ensembles are a natural choice as they ensemble without extra computational cost. 
Moreover,
previous work has suggested that interpolation in weight space
may improve performance when models share part of their optimization trajectory \cite{izmailov2018averaging, neyshabur2020being}.

\smallpara{Step 1: Standard fine-tuning.} 
As in Section~\ref{sec:expsetup}, we let $\mathcal{S}_{\mathrm{ref}}^\text{tr}$ denote the dataset used for fine-tuning and $g$ denote the image encoder used by CLIP.
We are now explicit in writing $g\mleft(x, \mathbf{V}_{\text{enc}} \mright)$ where $x$ is an input image and $\mathbf{V}_{\text{enc}}$ are the parameters of the encoder $g$.
Standard fine-tuning considers the model $f\mleft(x, \theta\mright) = g\left(x, \mathbf{V}_{\text{enc}}\right)^\top \textbf{W}_{\text{classifier}}$ where $\textbf{W}_{\text{classifier}} \in \mathbb{R}^{d \times k}$ is the classification head and $\theta = \left[\mathbf{V}_{\text{enc}}, \textbf{W}_{\text{classifier}} \right]$ are the parameters of $f$. 
We then solve $\argmin_{\theta} \left\{ \sum_{(x_i, y_i) \in \mathcal{S}^\text{tr}_\mathrm{ref}} \ell\mleft(f(x_i,\theta),\ y_i \mright) + \lambda R(\theta) \right\}$ where $\ell$ is the cross-entropy loss and $R$ is a regularization term (e.g., weight decay).
We consider the two most common variants of fine-tuning: end-to-end, where all values of $\theta$ are modified, and fine-tuning only a linear classifier, where $\mathbf{V}_{\text{enc}}$ is fixed at the value learned during pre-training. 
Appendices~\ref{sec:e2e-ft} and \ref{sec:moreclf} provide additional details. %

\smallpara{Step 2: Weight-space ensembling.}
For a \textit{\ALPHA{}}  $\alpha \in [0,1]$, we consider the \textit{weight-space ensemble} between the zero-shot model with parameters $\theta_0$ and the model obtained via standard fine-tuning with parameters $\theta_1$. 
The predictions of the weight-space ensemble $\mathsf{wse}$ are given by
\begin{align}\label{eqn:wse}
    \mathsf{wse}(x, \alpha) = f\mleft(x, (1-\alpha)\cdot\theta_0 + \alpha\cdot\theta_1 \mright)\ ,
\end{align}
i.e., we use the element-wise weighted average of the zero-shot and fined-tuned parameters. 
When fine-tuning only the linear classifier, weight-space ensembling is equivalent to the traditional output-space ensemble \cite{dietterich2000ensemble, breiman1996bagging, FREUND1997119} $(1-\alpha)\cdot f\mleft(x, \theta_0\mright)  + \alpha\cdot f\mleft(x, \theta_1 \mright)$ since Equation~\ref{eqn:wse} decomposes as $(1-\alpha)\cdot g\mleft(x, \mathbf{V}_{\text{enc}}\mright)^\top \textbf{W}_{\text{zero-shot}} + \alpha \cdot g\mleft(x, \mathbf{V}_{\text{enc}}\mright)^\top \textbf{W}_{\text{classifier}}$.
 
 As neural networks are non-linear with respect
 to their parameters, ensembling all layers---as we do when end-to-end fine-tuning---typically fails, achieving no better accuracy than a randomly initialized neural network \cite{frankle2020linear}. However, as similarly observed by previous work where part of the optimization trajectory is shared \cite{izmailov2018averaging,frankle2020linear, neyshabur2020being}, we find that the zero-shot and fine-tuned models are connected by a linear path in weight-space along which accuracy remains high (explored further in Section~\ref{sec:landscape}).

Remarkably, as we show in Section \ref{sec:results},  WiSE-FT improves accuracy under distribution shift while maintaining high performance on the reference distribution relative to fine-tuned models. 
These improvements come without any additional computational cost as a single set of weights is used.

\beforesec
\section{Results} 
\label{sec:results}
\postsec

This section presents our key experimental findings.
First, we show that WiSE-FT boosts the accuracy of a fine-tuned CLIP model on five ImageNet distribution shifts studied by \citet{radford2021learning}, while maintaining or improving ImageNet accuracy.
Next, we present additional experiments, including more distribution shifts, the effect of hyperparameters, accuracy improvements on the reference distribution, and experiments in the low-data regime. 
Finally, we demonstrate that our findings are more broadly applicable by exploring WiSE-FT for BASIC \cite{pham2021scaling}, ALIGN \cite{jia2021scaling}, and a ViT-H/14 \cite{dosovitskiy2021an} model pre-trained on JFT-300M \cite{sun2017revisiting}.

\begin{table*}
\setlength\tabcolsep{5.1pt}
\small
\begin{center}
\begin{tabular}{lc|ccccc|cc}
\toprule
{} &            &             \multicolumn{5}{c|}{Distribution shifts}             &                    Avg &                        Avg\\
{} &           IN (reference) &             IN-V2 &              IN-R &                 IN-Sketch &                 ObjectNet* &              IN-A &                    shifts &                        ref., shifts\\
\midrule
CLIP \texttt{ViT-L/14@336px} & & & & & & & &\\
\quad Zero-shot \cite{radford2021learning} &                    76.2 &                    70.1 &                    88.9 &                             60.2 &                             70.0 &                    77.2 &                             73.3 &                             74.8 \\
\quad  Fine-tuned LC \cite{radford2021learning} &                    85.4 &                    75.9 &                    84.2 &                             57.4 &                             66.2 &                    75.3 &                             71.8 &                             78.6 \\
\quad  Zero-shot (PyTorch)                   &                    76.6 &                    70.5 &                    89.0 &                             60.9 &                             69.1 &                    77.7 &                             73.4 &                             75.0 \\
\quad  Fine-tuned LC (ours)              &                    85.2 &                    75.8 &                    85.3 &                             58.7 &                             67.2 &                    76.1 &                             72.6 &                             78.9 \\
\quad  Fine-tuned E2E (ours)                     &                    86.2 &                    76.8 &                    79.8 &                             57.9 &                             63.3 &                    65.4 &                             68.6 &                             77.4 \\\midrule
\quad  WiSE-FT (ours) & & & & & & & &\\
\quad \quad LC, $\alpha{=}0.5$     &                    83.7 &                    76.3 &                    89.6 &  63.0 &  70.7 &  79.7 &  75.9 &  79.8 \\
\quad \quad LC, optimal $\alpha$        &                    85.3 &                    76.9 &                    89.8 &                             63.0 &                             70.7 &                    79.7 &                             75.9 &                             80.2 \\
\quad \quad E2E, $\alpha{=}0.5$ &                    86.8 &  \textbf{79.5} &                    89.4 &                    64.7 &                    71.1 &                    79.9 &                    76.9 &                    81.8 \\

\quad \quad E2E, optimal $\alpha$       &  \textbf{87.1} &  \textbf{79.5} &  \textbf{90.3} &  \textbf{65.0} &  \textbf{72.1} &  \textbf{81.0} & \textbf{77.4} &  \textbf{81.9} \\

\bottomrule
\end{tabular}
\caption{\label{tab:main} Accuracy of various methods on ImageNet and derived distribution shifts for CLIP \texttt{ViT-L/14@336px} \cite{radford2021learning}. E2E: end-to-end; LC: linear classifier. \textit{Avg shifts} displays the mean performance among the five distribution shifts, while \textit{Avg reference, shifts} shows the average of ImageNet (reference) and Avg shifts. For optimal $\alpha$, we choose the single mixing coefficient that maximizes the column. Results for additional models are provided in Appendix~\ref{sec:more-models}.
}

\end{center}
\end{table*}

\paragraph{Main results: ImageNet and associated distribution shifts.}
\label{sec:main_results}
As illustrated in Figure~\ref{fig:fig1}, when the \ALPHA{} $\alpha$ varies from $0$ to $1$, $\mathsf{wse}(\cdot, \alpha)$ is able to simultaneously improve accuracy on both the reference and shifted distributions. A breakdown for each dataset is shown in Appendix~\ref{sec:fig1breakdown}. 
Table \ref{tab:main} presents our main results on ImageNet and five derived distribution shifts. 
WiSE-FT (end-to-end, $\alpha{=}0.5$) outperforms numerous strong models in both average accuracy under distribution shift and the average accuracy on the reference and shifted distributions.
While future work may lead to more sophisticated strategies for choosing the mixing coefficient $\alpha$, $\alpha{=}0.5$ yields close to optimal performance across a range of experiments. Hence, we recommend $\alpha{=}0.5$ when no domain knowledge is available. Appendix \ref{sec:appendix_alpha} further explores the effect of $\alpha$. Moreover, results for twelve additional backbones are shown in Appendix \ref{sec:appendix_additional_exps}.

\paragraph{Robustness on additional distribution shifts.}
Beyond the five distribution shifts derived from ImageNet, WiSE-FT consistently improves robustness on a diverse set of further distributions shifts including geographic shifts in satellite imagery and  wildlife recognition (WILDS-FMoW \cite{wilds2021, christie2018functional}, WILDS-iWildCam \cite{wilds2021, beery2021iwildcam}), reproductions of the popular image classification dataset CIFAR-10 \cite{krizhevsky2009learning} with a distribution shift (CIFAR-10.1 \cite{pmlr-v97-recht19a} and CIFAR-10.2 \cite{lu2020harder}), and datasets with distribution shift induced by temporal perturbations in videos (ImageNet-Vid-Robust and YTBB-Robust \cite{shankar2020evaluating}). 
Concretely, WiSE-FT ($\alpha{=}0.5$) improves performance under distribution shift by 3.5, 6.2, 1.7, 2.1, 9.0 and 23.2 pp relative to the fine-tuned solution while decreasing performance on the reference distribution by at most 0.3 pp (accuracy on the reference distribution often improves).
In contrast to the ImageNet distribution shifts, the zero-shot model initially achieves less than 30\% accuracy on the WILDS distribution shifts, and WiSE-FT provides improvements regardless.
Appendix \ref{sec:moreshift} (Figure~\ref{fig:fig_more_shifts} and Table~\ref{tab:moreshifts}) includes more detailed results.

\begin{figure}
    \centering
    \includegraphics[width=\textwidth]{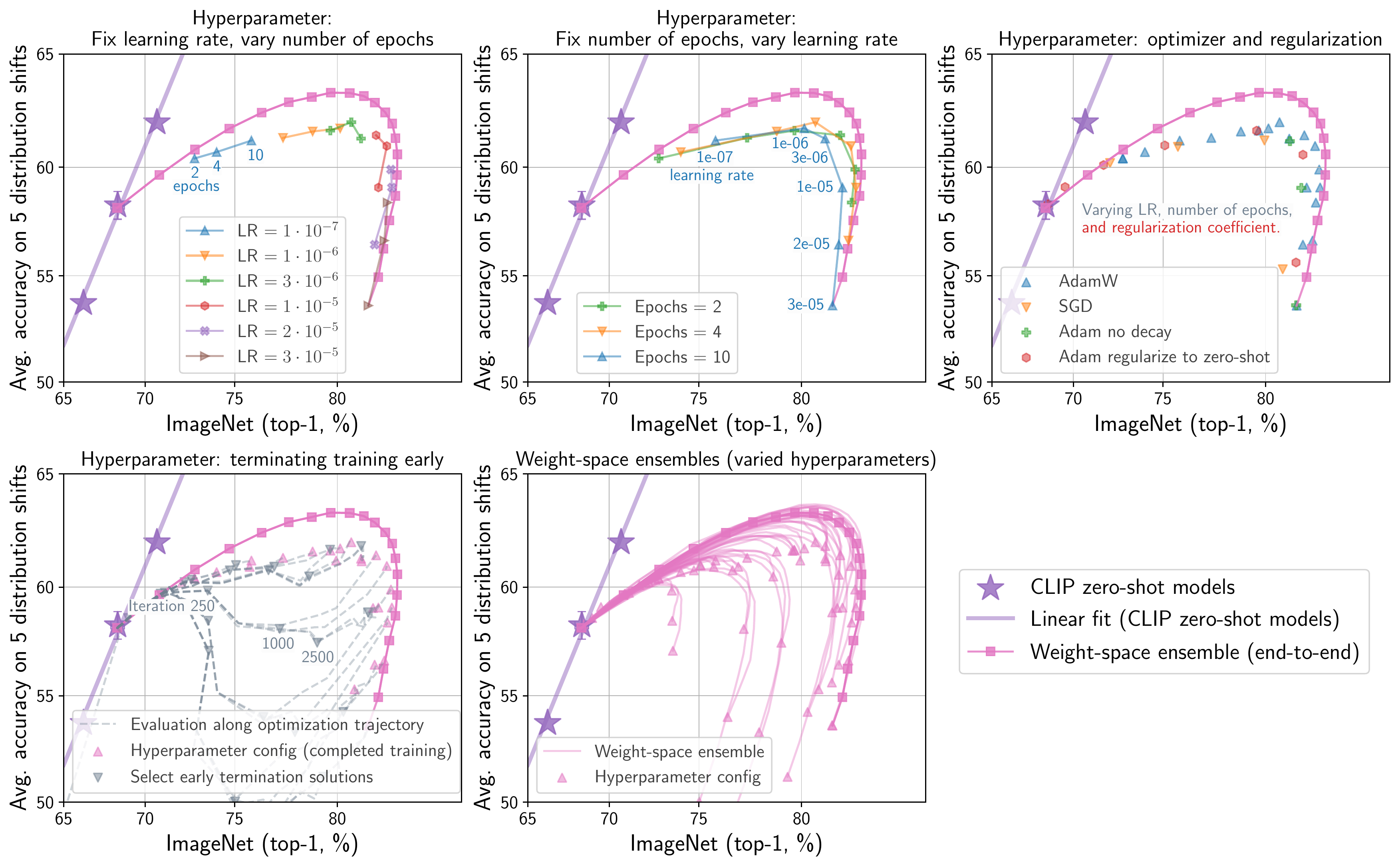}
    \caption{
    The robustness of fine-tuned models varies substantially under even small changes in hyperparameters.
    Applying WiSE-FT addresses this brittleness and can remove the trade-off between accuracy on the reference and shifted distributions.
    Results shown for CLIP \texttt{ViT-B/16} fine-tuned with cosine-annealing learning rate schedule and all models in the top left and top middle plots are fine-tuned with AdamW \cite{loshchilov2018decoupled}. Moreover, 
    \textit{regularize to zero-shot} appends the regularizer $\lambda \mleft\|\theta - \theta_0  \mright\|_2^2$ to the fine-tuning objective, where $\theta_0$ are the parameters of the zero-shot model.
    }
    \label{fig:hparams}
\end{figure}

\paragraph{Hyperparameter variation and alternatives.} As illustrated by Figure~\ref{fig:hparams}, moderate changes in standard hyperparameters such as the learning rate or the number of epochs can substantially affect performance under distribution shift. Moreover, these performance differences cannot be detected reliably from model performance on reference data alone.
For instance, while training for 10 epochs with learning rate $3 \cdot 10^{-5}$ and $3 \cdot 10^{-6}$ lead to a small accuracy difference on ImageNet (0.3 pp), accuracy under distribution shift varies by as much as 8 pp.

Furthermore, tuning hyperparameters on ImageNet data can also reduce robustness. For instance, while moving from small to moderate learning rates ($10^{-7}$ to $3 \cdot 10^{-5}$) improves performance on ImageNet by 5 pp, it also deteriorates accuracy under distribution shift by 8 pp.

WiSE-FT addresses this brittleness of hyperparameter tuning: even when using a learning rate $3 \cdot 10^{-5}$ where standard fine-tuning leads to low robustness, applying WiSE-FT removes the trade-off between accuracy on the reference and shifted distributions. 
The models which can be achieved by varying $\alpha$ are as good or better than those achievable by other hyperparameter configurations. Then, instead of searching over a wide range of hyperparameters, only $\alpha$ needs to be considered. Moreover, evaluating different values of $\alpha$ does not require training new models.

There is no hyperparameter in Figure~\ref{fig:hparams} which can be varied to match or exceed the optimal curve produced by WiSE-FT.
In our experiments, this frontier is reached only through methods that average model weights, either using WiSE-FT or
with a more sophisticated averaging scheme: keeping an exponential moving average of all model iterates (EMA, \cite{szegedy2016rethinking}).
Comparisons with EMA are detailed in Appendix~\ref{sec:ema}.

\customfootnotetext{$*$}{Although this table considers ImageNet class names, ObjectNet provides alternative class names which can improve the performance of zero-shot CLIP by 2.3 percentage points (Appendix \ref{sec:objectnet}).}

Additional comparisons are also presented in Appendix~\ref{sec:baselines-appendix}, including distillation, additional regularization, and CoOp \cite{coop}. Finally, Appendix~\ref{sec:dataaug} recreates Figure~\ref{fig:hparams} with stronger data augmentation and finds similar trends.

\paragraph{Accuracy gains on reference distributions.}
Beyond robustness to distribution shift, Table \ref{tab:id_gains} demonstrates that WiSE-FT also improves accuracy after fine-tuning on seven datasets. When fine-tuning end-to-end on ImageNet, CIFAR-10, CIFAR-100, Describable Textures, Food-101, SUN397, and Stanford Cars, WiSE-FT reduces relative error by 4 to 49\%.
Even though standard fine-tuning directly optimizes for high accuracy on the reference distribution, WiSE-FT achieves better performance.
Appendix \ref{sec:low-data} includes more details, including explorations in the low-data regime.

\begin{table}
\setlength\tabcolsep{3.5pt}
\begin{center}
\small
\begin{tabular}{l@{\hskip .2in}c@{\hskip .2in}ccccccc}
\toprule
{} &     ImageNet &      CIFAR10 &     CIFAR100 &         Cars &          DTD &       SUN397 &      Food101 \\          \midrule                                                                             Standard fine-tuning    &         86.2 &                                            98.6 &                                            92.2 &                                            91.6 &                                            81.9 &                                            80.7 &                                            94.4 \\                   WiSE-FT ($\alpha{=}0.5$) &  86.8 {\scriptsize\textcolor{LimeGreen}{(+0.6)}} &  99.3 {\scriptsize\textcolor{LimeGreen}{(+0.7)}} &  93.3 {\scriptsize\textcolor{LimeGreen}{(+1.1)}} &  93.3 {\scriptsize\textcolor{LimeGreen}{(+1.7)}} &  84.6 {\scriptsize\textcolor{LimeGreen}{(+2.8)}} &  83.2 {\scriptsize\textcolor{LimeGreen}{(+2.5)}} &  96.1 {\scriptsize\textcolor{LimeGreen}{(+1.6)}} \\                                                         WiSE-FT (opt. $\alpha$)  &  87.1 {\scriptsize\textcolor{LimeGreen}{(+0.9)}} &  99.5 {\scriptsize\textcolor{LimeGreen}{(+0.8)}} &  93.4 {\scriptsize\textcolor{LimeGreen}{(+1.2)}} &  93.6 {\scriptsize\textcolor{LimeGreen}{(+2.0)}} &  85.2 {\scriptsize\textcolor{LimeGreen}{(+3.3)}} &  83.3 {\scriptsize\textcolor{LimeGreen}{(+2.6)}} &  96.2 {\scriptsize\textcolor{LimeGreen}{(+1.8)}} \\  
\bottomrule
\end{tabular}
\caption{\label{tab:id_gains} Beyond robustness, WiSE-FT can improve accuracy after fine-tuning on several datasets.}%
\end{center}
\end{table}

\begin{figure}
    \centering
    \includegraphics[width=\textwidth]{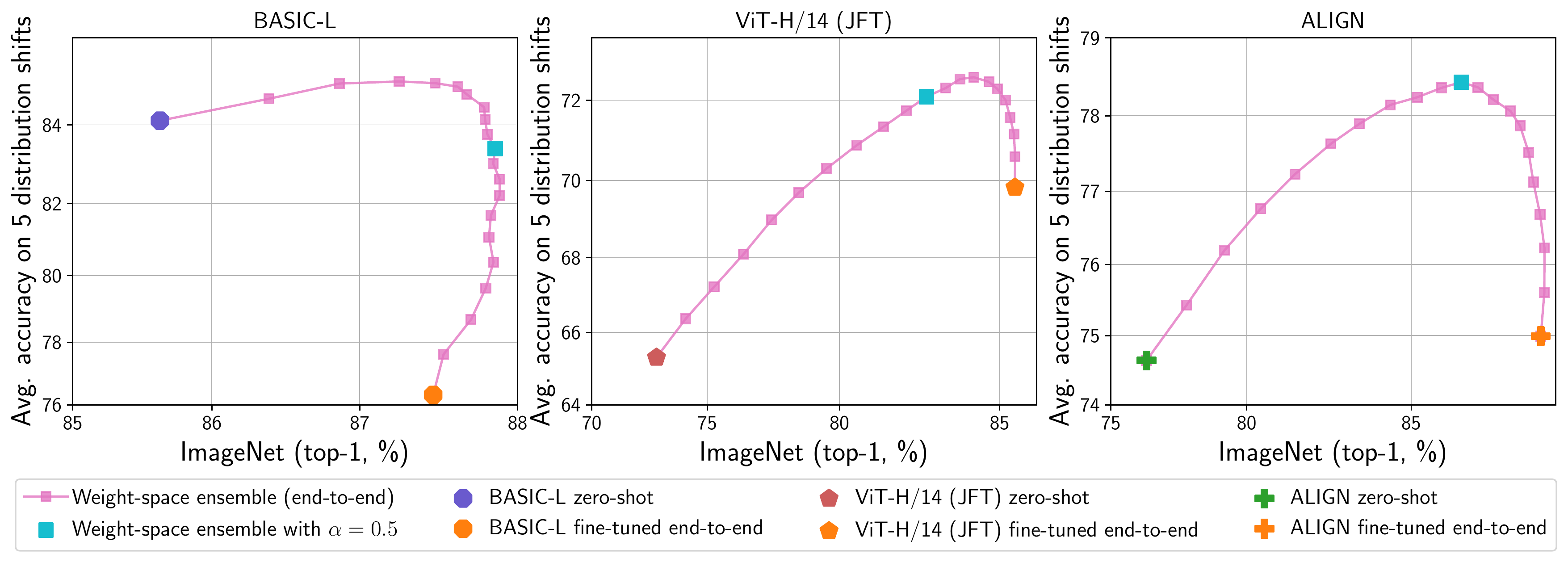}
    \caption{WiSE-FT applied to BASIC-L \cite{pham2021scaling}, a ViT-H/14 \cite{dosovitskiy2021an} model pre-trained on JFT-300M \cite{sun2017revisiting} and ALIGN~\cite{jia2021scaling}.}
    \label{fig:beyondclip}
    \vspace*{-0.4cm}
\end{figure}

\paragraph{Beyond CLIP.} Figure~\ref{fig:beyondclip} illustrates that WiSE-FT is generally applicable to zero-shot models beyond CLIP, and beyond models pre-trained contrastively with image-text pairs. First, we interpolate between the weights of the zero-shot and fine-tuned BASIC-L model \cite{pham2021scaling}, finding that $\alpha{=}0.5$ improves average accuracy on five distribution shifts derived from ImageNet by over 7 pp while improving ImageNet accuracy by 0.4 pp relative to the fine-tuned BASIC-L model (a per-dataset breakdown is provided in Figure~\ref{fig:basicbreakdown} and Table~\ref{tab:basic_l} of the Appendix).
As in \citet{pham2021scaling}, the model is fine-tuned using a contrastive loss and half of the ImageNet training data. 
WiSE-FT provides improvements on both reference and shifted distributions, despite these experimental differences.

Next, we consider the application of WiSE-FT to a ViT-H/14 model \cite{dosovitskiy2021an} pre-trained on JFT-300M \cite{sun2017revisiting},
where the zero-shot classifier is constructed by manually identifying a class correspondence (details provided in Section~\ref{sec:JFT}).
WiSE-FT improves performance under distribution shift over both the zero-shot and fine-tuned models. 
When $\alpha{=}0.8$, WiSE-FT outperforms the fine-tuned model by 2.2 pp on distribution shifts, while maintaining ImageNet performance within 0.2 pp of the fine-tuned model.
This result demonstrates that WiSE-FT can be successfully applied even to models which do not use contrastive image-text pre-training.

Finally, we apply WiSE-FT to the ALIGN model of \citet{jia2021scaling}, which is similar to CLIP but is pre-trained with a different dataset, finding similar trends.

\vspace*{-0.1cm}
\beforesec
\section{Discussion}
\label{sec:discussion}
\postsec
This section further analyzes the empirical phenomena we have observed so far.
We begin with the case where only the final linear layer is fine-tuned and predictions from the weight-space ensemble can be factored into the outputs of the zero-shot and fine-tuned model.
Next, we connect our observations regarding end-to-end fine-tuning with earlier work on the phenomenology of deep learning.%
\vspace*{-0.1cm}
\beforesec
\subsection{Zero-shot and fine-tuned models are complementary}
\label{sec:decomposing}
\postsec

In this section, we find that the zero-shot and fine-tuned models have diverse predictions, both on reference and shifted distributions. Moreover, while the fine-tuned models are more confident on the reference distribution, the reverse is true under distribution shift.  

\smallpara{Zero-shot and fine-tuned models are diverse.}
\label{sec:diversity}
In certain cases, ensemble accuracy is correlated with diversity among the constituents \cite{kuncheva2003measures, gontijo2021no}.
If two models make coincident mistakes, so will their ensemble, and no benefit will be gained from combining them. 
Here, we explore two measures of diversity: {\it prediction diversity,} which measures the fraction of examples for which two classifiers disagree but one is correct; and {\it Centered Kernel Alignment Complement,}  the complement of CKA \cite{kornblith2019similarity}. 
Additional diversity measures and details are provided in Appendix \ref{sec:diversity_defs}.
In Figure \ref{fig:diversity_and_confidence} (left), we show that the zero-shot and fine-tuned models are diverse both on the reference and shifted distributions, despite sharing the same backbone. 
As a point of comparison, we include avg. diversity measures between two linear classifiers fine-tuned with random splits on half of 
ImageNet,\footnote{Two linear classifiers fine-tuned on the same data converge to similar solutions, resulting in negligible diversity. 
As a stronger baseline, we fine-tune classifiers on different subsets of ImageNet, with half of the data.} denoted in orange in Figure \ref{fig:diversity_and_confidence}.

\begin{figure*}
    \centering
    \includegraphics[width=\textwidth]{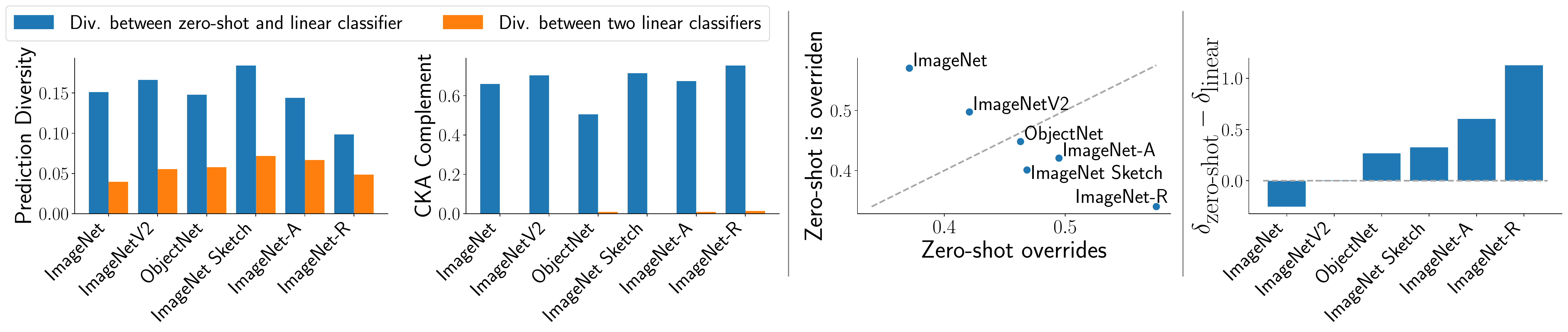}
    \caption{\textbf{(Left)} Zero-shot and fine-tuned models exhibit diversity in their predictions. \textbf{(Middle)} On most distribution shifts, the zero-shot model overrides the linear classifier more than it is overridden. The reverse is true for ImageNet (reference). \textbf{(Right)} Similarly, zero-shot models are more confident under distribution shift, while the reverse is true on the reference distribution. The margin $\delta_f$ measures the average difference between the largest and second largest unormalized output for classifier $f$}
    \label{fig:diversity_and_confidence}
\end{figure*}

\smallpara{Models are more confident where they excel.}
In order for the ensemble model to be effective, it should leverage each model's expertise based on which distribution the data is from.
Here, we empirically show that this occurs on a number of datasets we consider.
First, we examine the cases where the models being ensembled disagree. 
We say the zero-shot model \textit{overrides} the fine-tuned model if their predictions disagree and the zero-shot prediction matches that of the weight-space ensemble. 
Similarly, if models disagree and the linear classifier prediction matches the ensemble, we say the zero-shot is \textit{overridden}.
Figure \ref{fig:diversity_and_confidence} (middle) shows the fraction of samples where the zero-shot model overrides and is overridden by the fine-tuned linear classifier for $\alpha{=}0.5$. 
Other than ImageNetV2, which was collected to closely reproduce ImageNet, the zero-shot model overrides the linear classifier more than it is overridden on the distribution shifts.

Additionally, we are interested in measuring model confidence. Recall that we are ensembling quantities before a softmax is applied, so we avoid criteria that use probability vectors, e.g., \citet{guo2017calibration}.
Instead, we consider the margin $\delta$ between the largest and second largest output of each classifier. 
Figure \ref{fig:diversity_and_confidence} (right) shows that the zero-shot model is more confident in its predictions under distribution shift, while the reverse is true on the reference distribution.
\beforesec
\subsection{An error landscape perspective}
\label{sec:landscape}
\postsec
We now turn to empirical phenomena we observe when weight-space ensembling \emph{all} layers in the network.
Specifically, this section formalizes our observations and details related phenomena. 
Recall that the weight-space ensemble of $\theta_0$ and $\theta_1$ is given by $f\mleft(x, (1-\alpha)\cdot\theta_0 + \alpha\cdot\theta_1 \mright)$ (Equation~\ref{eqn:wse}).

For a distribution $\mathcal{D}$ and model $f$, let $\mathsf{Acc}_{\mathcal{D}, f}(\theta)$ denote the expected accuracy of $f$ evaluated with parameters $\theta$ on distribution $\mathcal{D}$.

\begin{figure*}
    \centering
    \includegraphics[width=\textwidth]{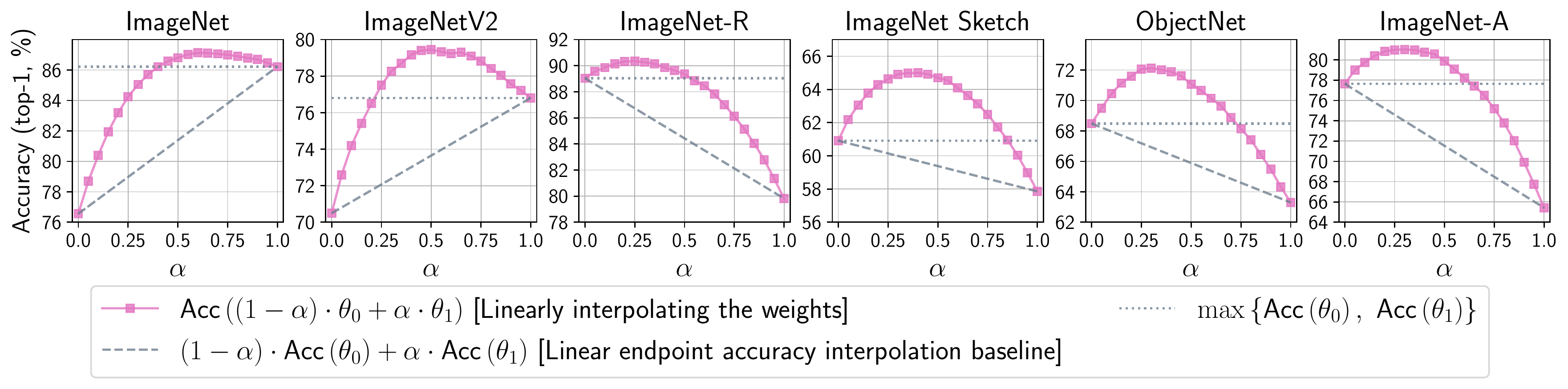}
    \caption{On ImageNet and the main distribution shifts we consider, linearly interpolating between the weights of $\theta_0$ and $\theta_1$ exceeds the baseline of linearly interpolating the accuracies of the two models for all $\alpha$ (Observation 1). Moreover, there exists an $\alpha$ for which WiSE-FT outperforms both the zero-shot and fine-tuned models (Observation 2).}
    \label{fig:def}
\end{figure*}

\textbf{Observation 1:} As illustrated in Figure~\ref{fig:def}, on ImageNet and the five associated distribution shifts we consider
\begin{equation} \label{eq:base}
\mathsf{Acc}_{\mathcal{D}, f}\mleft((1-\alpha)\cdot \theta_0 +  \alpha \cdot \theta_1\mright)\ \geq \ (1-\alpha) \cdot \mathsf{Acc}_{\mathcal{D}, f}\mleft(\theta_0\mright) + \alpha \cdot \mathsf{Acc}_{\mathcal{D}, f}\mleft(\theta_1\mright)
\end{equation}
for all $\alpha \in [0,1]$.

Note that equation~\ref{eq:base} uses the baseline of linearly interpolating between the accuracies of the two endpoints, 
which is always achievable by using weights
$\theta_1$ with probability $\alpha$ and using model $\theta_0$ otherwise.
In the case where the accuracy of both endpoints are similar,
Equation~\ref{eq:base} is equivalent to the definition of Linear Mode Connectivity of \citet{frankle2020linear}.

To assist in contextualizing Observation 1, we review related phenomena. Neural networks are nonlinear, hence weight-space ensembles only achieve good performance in exceptional cases---interpolating the weights of two networks trained from a random initialization results in no better accuracy than a random classifier \cite{frankle2020linear}. 
Linear mode connectivity has been observed by 
\citet{frankle2020linear}; \citet{izmailov2018averaging}  when part of the training trajectory is shared, and by \citet{neyshabur2020being} when two models are fine-tuned with a shared initialization.
In particular, the observations of \citet{neyshabur2020being} may elucidate why weight-space ensembles attain high 
accuracy in the setting we consider, as they suggest that fine-tuning remains in a region where solutions are connected by a linear path along which error remains low.
Instead of considering the weight-space ensemble of two fine-tuned models, we consider the weight-space ensemble of the 
\textit{pre-trained} and fine-tuned models. This is only possible for a pre-trained model capable of zero-shot inference such as CLIP.

\textbf{Observation 2:} As illustrated by Figure~\ref{fig:def}, on ImageNet and the five associated distribution shifts we consider, weight-space ensembling (end-to-end) may outperform both the zero-shot and fine-tuned models, i.e., there exists an $\alpha$ for which $\mathsf{Acc}_{\mathcal{D}, f}\left((1-\alpha) \cdot \theta_0 +  \alpha \cdot \theta_1\right) \ \geq \ \max\left\{  \mathsf{Acc}_{\mathcal{D}, f}\left(\theta_0\right) ,  \ \mathsf{Acc}_{\mathcal{D}, f}\left(\theta_1\right) \right\}$.

We are not the first to observe that when interpolating between models, the accuracy of models along the path may exceed that of either endpoint \cite{izmailov2018averaging, neyshabur2020being, pmlr-v139-wortsman21a}. \citet{neyshabur2020being}
conjecture that interpolation could produce solutions closer to the true center of a basin. 
In contrast to \citet{neyshabur2020being}, we interpolate between models which observe different data.

\beforesec
\section{Related work}\label{sec:related}
\postsec
\smallpara{Robustness.}
Understanding how models perform under distribution shift remains an important goal, as real world models may encounter data from new environments \cite{quinonero2009dataset, torralba2011unbiased}. 
Previous work has studied model behavior under synthetic \cite{imagenetc,tramer2017ensemble,madry2017towards,geirhos2018generalisation,eykholt2018robust,alcorn2019strike} and natural distribution shift \cite{imagenetr, wilds2021, imagenetsketch, objectnet, imageneta}.
Interventions used for synthetic shifts do not typically provide robustness to many natural distribution shifts \cite{taori2020measuring}. 
In contrast, accuracy on the reference distribution is often a reliable predictor for accuracy under distribution shift \cite{yadav2019cold, miller2020effect, taori2020measuring, sun2020scalability, miller21b}. On the other hand, \citet{d2020underspecification} show that accuracy under certain distribution shifts cannot be reliably inferred from accuracy on the reference distribution.
We observe a similar phenomenon when fine-tuning with different hyperparameters (Section~\ref{sec:results}, Figure~\ref{fig:hparams}).

\smallpara{Pre-training and transfer learning.}
Pre-training on large amounts of data is a powerful technique for building high-performing machine learning systems %
\cite{sharif2014cnn, dosovitskiy2021an,kolesnikov2020big,yalniz2019billion,radford2019language,brown2020language}.
One increasingly popular class of vision models are those pre-trained with auxiliary language supervision, which can be used for zero-shot inference \cite{desai2021virtex,sariyildiz2020learning,zhang2020contrastive,radford2021learning,jia2021scaling,pham2021scaling,zhai21lit}.
When pre-trained models are adapted to a specific distribution through standard fine-tuning, effective robustness deteriorates at convergence \cite{andreassen2021evolution}.
In natural language processing, previous work proposed stable fine-tuning methods that incur computational overhead \cite{jiang2019smart,zhu2019freelb}, alleviating problems such as representational collapse \cite{aghajanyan2021better}.
More generally, a variety of methods have attempted to mitigate catastrophic forgetting \cite{MCCLOSKEY1989109}. \citet{kirkpatrick2017overcoming}; \citet{zenke2017continual} explored weighted quadratic regularization for sequential learning. \citet{xuhong2018explicit} showed that, for fine-tuning, the simple quadratic regularization explored in Section~\ref{sec:results} performs best, while \citet{lubana2021quadratic} explored the connection between quadratic regularization and interpolation.
\citet{andreassen2021evolution} found that many approaches from continual learning do not provide robustness to multiple natural distribution shifts. Finally, \citet{li2020rethinking} investigate the effect of fine-tuning hyperparameters on performance.

\smallpara{Traditional (output-space) ensembles.}
Traditional ensemble methods, which we refer to as output-space ensembles, combine the predictions (outputs) of many classifiers \cite{dietterich2000ensemble, bauer1999empirical, breiman1996bagging,friedman2001elements, deepensembles, FREUND1997119}. 
Typically, output-space ensembles outperform individual classifiers and
provide uncertainty estimates under distribution shift that are more callibrated than baselines \cite{deepensembles,ovadia2019can,stickland2020diverse}. In contrast to these works, we consider the ensemble of two models which have observed different data.
Output-space ensembles require more computational resources as they require a separate pass through each model.
Compared to an ensemble of 15 models trained on the same dataset, \citet{mustafa2020deep} find an improvement of 0.8--1.6 pp  under distribution shift (on ImageNetV2, ImageNet-R, ObjectNet, and ImageNet-A) by ensembling a similar number of models pre-trained on different datasets.
In contrast, we see an improvement of 2--15 pp from ensembling two models. Moreover, as we ensemble in weight-space, no extra compute is required compared to a single model.

\paragraph{Weight-space ensembles.}

Weight-space ensembles linearly interpolate between the weights of different models \cite{frankle2020linear, lucas2021analyzing, goodfellow2014qualitatively, szegedy2016rethinking}. For example, \citet{izmailov2018averaging} average checkpoints saved throughout training for improved performance.
Indeed, averaging the weights along the training trajectory is a central method in optimization \cite{ruppert1988efficient, polyak1992acceleration, nichol2018first}.
For instance, \citet{zhang2019lookahead} propose optimizing with a set of fast and slow weights, where every $k$ steps, these two sets of weights are averaged and a new trajectory begins.
Here, we revisit these techniques from a distributional robustness perspective and consider the weight-space ensemble of models which have observed different data. 

\paragraph{Concurrent and subsequent work.} Topics including robust fine-tuning, ensembles for improved robustness, and interpolating the weights of fine-tuned models are studied in concurrent and subsequent work. \citet{kumar2021finetuning} observe that fine-tuning end-to-end often results in higher accuracy on the reference distribution but lower accuracy under distribution shift, compared to linear classifier fine-tuning. To address this, \citet{kumar2021finetuning} first fine-tune a linear classifier and use this as the initialization for end-to-end fine-tuning. We consider fine-tuning zero-shot models, and so we begin with a classifier (i.e., the zero-shot classifier) which we are using as the initialization for end-to-end fine-tuning. In a separate work, \citet{kumar2021calibrated} find that calibrated output-space ensembles can be used to mitigate accuracy trade-offs. In Figures \ref{fig:ose} and \ref{fig:beyond} of the Appendix, we observe that it is possible to mitigate accuracy trade-offs with output-space ensembles even without calibration.

\citet{hewitt2021ensembles} explore the application of output-space ensembles and distillation to mitigate accuracy trade-offs which arise in fine-tuning models for natural language generation. \citet{hewitt2021ensembles} observe that output-space ensembles mainly outperform distillation, which we observe for a separate domain in Figure~\ref{fig:baselines} of the Appendix.
\citet{lopes2021no} explore output-space ensembles of models across hyper-parameters, architectures, frameworks, and datasets.
They find that specializing in subdomains of data leads to high ensemble performance. 
Finally, \citet{matena2021merging} introduce a method of combining models in weight-space that goes beyond linear interpolation with a single mixing-coefficient as employed in WiSE-FT.
Specifically, \citet{matena2021merging} employ Fisher information as a measure of per-parameter importance.
While their experiments do not examine accuracy under distribution shift,
their goal of combining differing expertise into one shared model is well aligned with ours.
\beforesec
\section{Limitations, impact, and conclusion}
\label{sec:conclusion}
\postsec
\smallpara{Limitations.} While we expect our findings to be more broadly applicable to other domains such as natural language processing, our investigation here is limited to image classification.
Exploring fine-tuning for object detection and natural language processing are interesting directions for future work.
Moreover, although the interpolation parameter setting $\alpha{=}0.5$ provides good overall performance, we leave the question of finding the optimal $\alpha$ for specific target distributions to future work.

\smallpara{Impact.}
 \citet{radford2021learning} and \citet{brown2020language} extensively discuss the broader impact of large zero-shot models and identify potential causes of harm including
 model biases and potential malicious uses such as surveillance systems.
 WiSE-FT is a fine-tuning method that builds on such models, and thus may perpetuate their negative impact.
 
 \smallpara{Conclusion.}
WiSE-FT can substantially improve performance under distribution shift with minimal or no loss in accuracy on the target distribution compared to standard fine-tuning.
We view WiSE-FT as a first step towards more sophisticated fine-tuning schemes and anticipate that future work will continue to leverage the robustness of zero-shot models for building more reliable neural networks.

\subsection*{Acknowledgements}
We thank
Anders Andreassen,
Tim Dettmers,
Jesse Dodge,
Katie Everett,
Samir Gadre,
Ari Holtzman,
Sewon Min,
Mohammad Norouzi,
Nam Pho, 
Ben Poole,
Sarah Pratt,
Alec Radford,
Jon Shlens,
and Rohan Taori for helpful discussions and draft feedback, Hyak at UW for computing support, 
Rosanne Liu for fostering the collaboration,
and Basil Mustafa for providing an earlier version of the mapping between JFT and ImageNet classes.
This work is in part supported by NSF IIS 1652052, IIS 17303166, DARPA N66001-19-2-4031, DARPA W911NF-15-1-0543 and gifts from Allen Institute for Artificial Intelligence.

{\small
\bibliographystyle{plainnat}
\bibliography{main}
}

\clearpage

\appendix

\section{Pseudocode for WiSE-FT}
\label{sec:pseudo-code}

\begin{algorithm}[H]
\caption{Pytorch pseudocode for WiSE-FT}
\label{alg:code}
\definecolor{codeblue}{rgb}{0.25,0.5,0.5}
\definecolor{codeblue2}{rgb}{0,0,1}
\definecolor{mauve}{rgb}{0.58,0,0.82}
\lstset{
  backgroundcolor=\color{white},
  basicstyle=\fontsize{7.2pt}{7.2pt}\ttfamily\selectfont,
  columns=fullflexible,
  breaklines=true,
  captionpos=b,
  commentstyle=\fontsize{7.2pt}{7.2pt}\color{codeblue},
  keywordstyle=\fontsize{7.2pt}{7.2pt}\color{codeblue2},
  stringstyle=\color{mauve},
}
\begin{lstlisting}[language=python]
def wse(model, zeroshot_checkpoint, finetuned_checkpoint, alpha):
    # load state dicts from checkpoints
    theta_0 = torch.load(zeroshot_checkpoint)["state_dict"]
    theta_1 = torch.load(finetuned_checkpoint)["state_dict"]

    # make sure checkpoints are compatible
    assert set(theta_0.keys()) == set(theta_1.keys())

    # interpolate between all weights in the checkpoints
    theta = {
        key: (1-alpha) * theta_0[key] + alpha * theta_1[key]
        for key in theta_0.keys()
    }

    # update the model (in-place) according to the new weights
    model.load_state_dict(theta)

def wise_ft(model, dataset, zeroshot_checkpoint, alpha, hparams):
    # load the zero-shot weights
    theta_0 = torch.load(zeroshot_checkpoint)["state_dict"]
    model.load_state_dict(theta_0)

    # standard fine-tuning
    finetuned_checkpoint = finetune(model, dataset, hparams)

    # perform weight-space ensembling (in-place)
    wse(model, zeroshot_checkpoint, finetuned_checkpoint, alpha)
        
\end{lstlisting}
\end{algorithm}

\section{Mixing coefficient}
\label{sec:appendix_alpha}

Table \ref{tab:alpha} compares the performance of WiSE-FT using a fixed mixing coefficient $\alpha{=}0.5$ with the fixed optimal mixing coefficient. On ImageNet and the five derived distribution shifts, the average performance of the optimal $\alpha$ is $0$ to $0.4$ percentage points better than that of $\alpha{=}0.5$. 
Due to its simplicity and effectiveness, we recommend using $\alpha{=}0.5$ when no domain knowledge is available. Finding the optimal value of the mixing coefficient for any distribution is an interesting question for future work. Unlike other hyperparameters, no re-training is required to test different $\alpha$, so tuning is relatively cheap.

\section{Additional experiments}
\label{sec:appendix_additional_exps}

This section supplements the results of Section~\ref{sec:results}. First, in Section~\ref{sec:fig1breakdown} we provide a breakdown of Figure~\ref{fig:fig1} for each distribution shift.
Next, in Section~\ref{sec:moreshift} we provide effective robustness scatter plots for six additional distribution shifts, finding WiSE-FT to provide consistent improvements  under distribution shift without any loss in performance on the reference distribution.
Section~\ref{sec:baselines-appendix} compares WiSE-FT with additional alternatives including distillation and CoOp \cite{coop}.
Beyond robustness, Section~\ref{sec:low-data} demonstrates that WiSE-FT can provide accuracy improvements on reference data, with a focus on the low-data regime. Section~\ref{sec:scaling} showcases that the accuracy improvements under distribution shift are not isolated to large models, finding similar trends across scales of pre-training computes. Section \ref{sec:more-models} explores the application of WiSE-FT for additional models such as ALIGN \cite{jia2021scaling}, a ViT-H/14 model  pre-trained on JFT \cite{dosovitskiy2021an} and BASIC \cite{pham2021scaling}. Finally, Section~\ref{sec:beyond} ensembles zero-shot CLIP with an independently trained classifier.

\subsection{Breakdown of CLIP experiments on ImageNet}
\label{sec:fig1breakdown}

\begin{table*}
\setlength\tabcolsep{5.1pt}
\small
\begin{center}
\begin{tabular}{lc|ccccc|cc}
\toprule
{} &            &             \multicolumn{5}{c|}{Distribution shifts}             & Avg &     Avg\\
{} &           IN (ref.) &             IN-V2 &              IN-R &                 IN-Sketch &                 ObjectNet &              IN-A & shifts &     ref., shifts\\
\midrule
\texttt{ViT-B/16}, end-to-end & 0.9 & 0.4 & 1.4 & 0.2 & 0.4 & 2.4 & 0.5 & 0.0 \\
\texttt{ViT-B/16}, linear classifier & 1.8 & 0.6 & 1.2 & 0.1 & 0.2 & 0.6 & 0.1 & 0.2 \\
\texttt{ViT-L/14@336}, end-to-end & 0.3 & 0.0 & 0.9 & 0.3 & 1.0 & 1.1 & 0.5 & 0.1 \\
\texttt{ViT-L/14@336}, linear classifier & 1.6 & 0.6 & 0.2 & 0.0 & 0.0 & 0.0 & 0.0 & 0.4 \\
\bottomrule
\end{tabular}
\caption{\label{tab:alpha}
Difference in performance (percentage points) between WiSE-FT using the optimal mixing coefficient and a fixed value of $\alpha{=}0.5$ for CLIP \texttt{ViT-B/16} and \texttt{ViT-L/14@336}. For each cell in the table, the optimal mixing coefficient $\alpha$ is chosen individually such that the corresponding metric is maximized.
Results for all mixing coefficients are available in Tables \ref{tab:breakdown1} and \ref{tab:breakdown2}. \textit{Avg shifts} displays the mean performance among the five distribution shifts, while \textit{Avg reference, shifts} shows the average of ImageNet (reference) and Avg shifts.
}
\end{center}
\end{table*}

\begin{figure}[h]
    \centering
    \includegraphics[width=.9\textwidth]{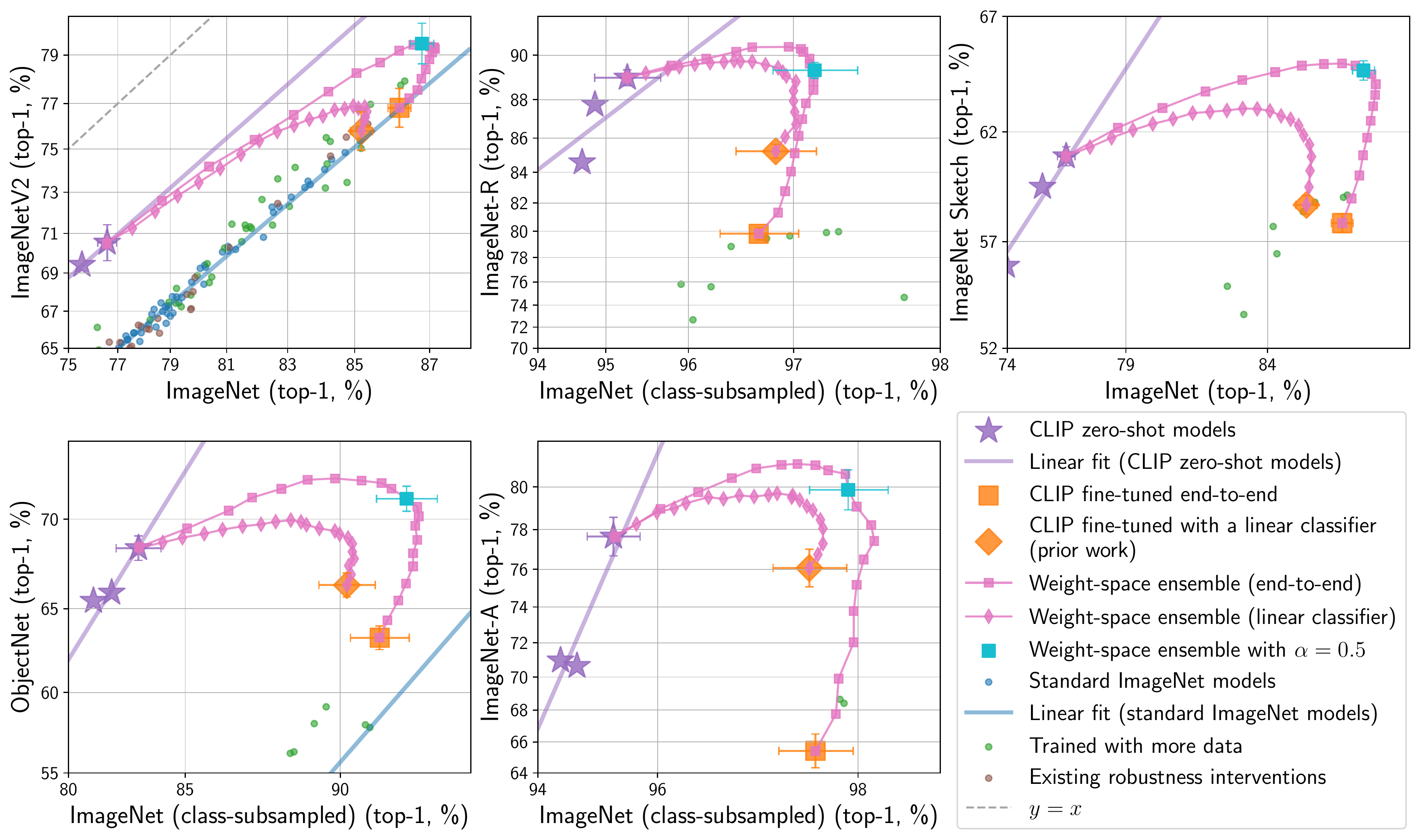}
    \caption{A per-dataset breakdown of the key experimental results (Figure~\ref{fig:fig1}). WiSE-FT improves accuracy on ImageNet and five derived distribution shifts.  Standard ImageNet models, models trained with more data, and existing robustness interventions are from the testbed of \citet{taori2020measuring}.}
    \label{fig:main_breakdown}
\end{figure}

\begin{figure}[h]
    \centering
    \includegraphics[width=.9\textwidth]{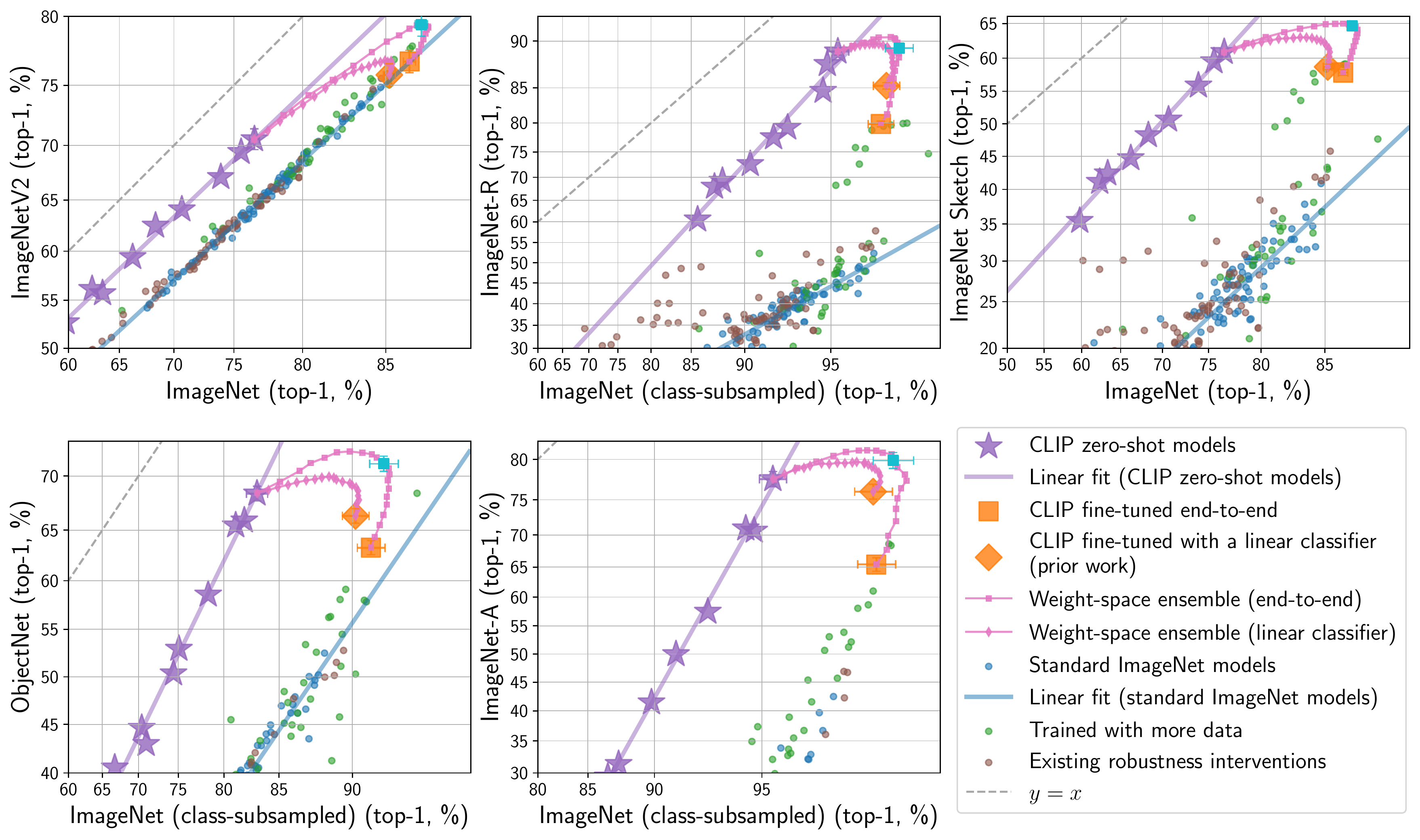}
    \caption{A zoomed-out version of Figure~\ref{fig:main_breakdown}. WiSE-FT improves accuracy on ImageNet and five derived distribution shifts.  Standard ImageNet models, models trained with more data, and existing robustness interventions are from the testbed of \citet{taori2020measuring}.}
    \label{fig:main_zoomout}
\end{figure}

In contrast to Figures \ref{fig:fig1} and \ref{fig:beyondclip}, where our key experimental results for ImageNet and five derived distribution shifts are averaged, we now display the results separately for each distribution shift. Results are provided in Figures~\ref{fig:main_breakdown}, \ref{fig:main_zoomout}. 

To assist in contextualizing the results, the scatter plots we display also show a wide range of machine learning models from a comprehensive testbed of evaluations \cite{taori2020measuring, miller21b}, including:
models trained on $\mathcal{S}^\text{tr}_\mathcal{D}$ (\textit{standard training}); models trained on additional data and fine-tuned using $\mathcal{S}^\text{tr}_\mathcal{D}$  
(\textit{trained with more data}); and models trained using various \textit{existing robustness interventions}, e.g. special data augmentation 
\cite{devries2017improved,engstrom2019exploring,geirhos2018imagenet,hendrycks2019augmix} or adversarially robust models \cite{madry2017towards, cohen2019certified,salman2019provably,shafahi2019adversarial}.

Additionally, Tables \ref{tab:breakdown1} and \ref{tab:breakdown2} show the performance of WiSE-FT for various values of the mixing coefficient $\alpha$ on ImageNet and five derived distribution shifts, for CLIP \texttt{ViT-L/14@336} and the \texttt{ViT-B/16} model.

\FloatBarrier

\FloatBarrier

\begin{table*}
\setlength\tabcolsep{5.1pt}
\small
\begin{center}
\begin{tabular}{lc|ccccc|cc}
\toprule
{} &            &             \multicolumn{5}{c|}{Distribution shifts}             & Avg &     Avg\\
{} &           IN (ref.) &             IN-V2 &              IN-R &                 IN-Sketch &                 ObjectNet &              IN-A & shifts &     ref., shifts\\
\midrule
WiSE-FT, end-to-end & & & & & & & &\\
\quad$\alpha{=}0.00$ & 76.6 & 70.5 & 89.0 & 60.9 & 68.5 & 77.6 & 73.3 & 74.9 \\
\quad$\alpha{=}0.05$ & 78.7 & 72.6 & 89.6 & 62.2 & 69.5 & 79.0 & 74.6 & 76.7 \\
\quad$\alpha{=}0.10$ & 80.4 & 74.2 & 89.9 & 63.1 & 70.4 & 79.8 & 75.5 & 78.0 \\
\quad$\alpha{=}0.15$ & 81.9 & 75.4 & 90.1 & 63.8 & 71.1 & 80.4 & 76.2 & 79.1 \\
\quad$\alpha{=}0.20$ & 83.2 & 76.5 &  \dunderline{1pt}{90.3} & 64.3 & 71.6 & 80.8 & 76.7 & 80.0 \\
\quad$\alpha{=}0.25$ & 84.2 & 77.5 &  \dunderline{1pt}{90.3} & 64.6 &  \dunderline{1pt}{72.1} &  \dunderline{1pt}{81.0} & 77.1 & 80.7 \\
\quad$\alpha{=}0.30$ & 85.1 & 78.3 &  \dunderline{1pt}{90.3} & 64.9 &  \dunderline{1pt}{72.1} &  \dunderline{1pt}{81.0} & 77.3 & 81.2 \\
\quad$\alpha{=}0.35$ & 85.7 & 78.7 & 90.1 &  \dunderline{1pt}{65.0} & 72.0 &  \dunderline{1pt}{81.0} &  \dunderline{1pt}{77.4} & 81.6 \\
\quad$\alpha{=}0.40$ & 86.2 & 79.2 & 89.9 &  \dunderline{1pt}{65.0} & 71.9 & 80.7 & 77.3 & 81.8 \\
\quad$\alpha{=}0.45$ & 86.6 & 79.4 & 89.6 & 64.9 & 71.6 & 80.6 & 77.2 &  \dunderline{1pt}{81.9} \\
\quad$\alpha{=}0.50$ & 86.8 &  \dunderline{1pt}{79.5} & 89.4 & 64.7 & 71.1 & 79.9 & 76.9 & 81.8 \\
\quad$\alpha{=}0.55$ & 87.0 & 79.3 & 88.9 & 64.5 & 70.7 & 79.1 & 76.5 & 81.8 \\
\quad$\alpha{=}0.60$ &  \dunderline{1pt}{87.1} & 79.2 & 88.5 & 64.1 & 70.1 & 78.2 & 76.0 & 81.5 \\
\quad$\alpha{=}0.65$ &  \dunderline{1pt}{87.1} & 79.3 & 87.8 & 63.6 & 69.6 & 77.4 & 75.5 & 81.3 \\
\quad$\alpha{=}0.70$ &  \dunderline{1pt}{87.1} & 79.1 & 87.0 & 63.1 & 68.9 & 76.5 & 74.9 & 81.0 \\
\quad$\alpha{=}0.75$ & 87.0 & 78.8 & 86.1 & 62.5 & 68.1 & 75.2 & 74.1 & 80.5 \\
\quad$\alpha{=}0.80$ & 86.9 & 78.4 & 85.1 & 61.7 & 67.4 & 73.8 & 73.3 & 80.1 \\
\quad$\alpha{=}0.85$ & 86.8 & 78.0 & 84.0 & 61.0 & 66.4 & 72.0 & 72.3 & 79.5 \\
\quad$\alpha{=}0.90$ & 86.7 & 77.6 & 82.8 & 60.0 & 65.5 & 69.9 & 71.2 & 79.0 \\
\quad$\alpha{=}0.95$ & 86.5 & 77.2 & 81.3 & 59.0 & 64.3 & 67.7 & 69.9 & 78.2 \\
\quad$\alpha{=}1.00$ & 86.2 & 76.8 & 79.8 & 57.9 & 63.3 & 65.4 &  68.6 & 77.4 \\
\midrule
WiSE-FT, linear classifier & & & & & & & &\\
\quad$\alpha{=}0.00$     & 76.6 & 70.5 & 89.0 & 60.9 & 69.1 & 77.7 & 73.4 & 75.0 \\
\quad$\alpha{=}0.05$     & 77.6 & 71.3 & 89.2 & 61.3 & 69.3 & 78.3 & 73.9 & 75.8 \\
\quad$\alpha{=}0.10$     & 78.4 & 72.1 & 89.4 & 61.7 & 69.6 & 78.8 & 74.3 & 76.3 \\
\quad$\alpha{=}0.15$     & 79.3 & 72.8 & 89.5 & 62.1 & 70.0 & 79.0 & 74.7 & 77.0 \\
\quad$\alpha{=}0.20$     & 80.0 & 73.5 & 89.6 & 62.4 & 70.3 & 79.3 & 75.0 & 77.5 \\
\quad$\alpha{=}0.25$     & 80.8 & 74.1 & 89.7 & 62.6 & 70.5 & 79.5 & 75.3 & 78.0 \\
\quad$\alpha{=}0.30$     & 81.5 & 74.8 & 89.7 & 62.8 &  \dunderline{1pt}{70.7} & 79.5 & 75.5 & 78.5 \\
\quad$\alpha{=}0.35$     & 82.1 & 75.4 &  \dunderline{1pt}{89.8} & 62.9 &  \dunderline{1pt}{70.7} & 79.6 & 75.7 & 78.9 \\
\quad$\alpha{=}0.40$     & 82.7 & 75.8 & 89.7 &  \dunderline{1pt}{63.0} &  \dunderline{1pt}{70.7} & 79.6 & 75.8 & 79.2 \\
\quad$\alpha{=}0.45$     & 83.2 & 76.1 & 89.7 &  \dunderline{1pt}{63.0} &  \dunderline{1pt}{70.7} & 79.6 & 75.8 & 79.5 \\
\quad$\alpha{=}0.50$     & 83.7 & 76.3 & 89.6 &  \dunderline{1pt}{63.0} &  \dunderline{1pt}{70.7} &  \dunderline{1pt}{79.7} &  \dunderline{1pt}{75.9} & 79.8 \\
\quad$\alpha{=}0.55$     & 84.1 & 76.5 & 89.5 & 62.9 & 70.5 & 79.6 & 75.8 & 79.9 \\
\quad$\alpha{=}0.60$     & 84.4 & 76.7 & 89.3 & 62.7 & 70.3 & 79.5 & 75.7 & 80.1 \\
\quad$\alpha{=}0.65$     & 84.7 & 76.8 & 89.1 & 62.6 & 70.1 & 79.4 & 75.6 &  \dunderline{1pt}{80.2} \\
\quad$\alpha{=}0.70$     & 85.0 &  \dunderline{1pt}{76.9} & 88.9 & 62.3 & 69.9 & 79.1 & 75.4 &  \dunderline{1pt}{80.2} \\
\quad$\alpha{=}0.75$     & 85.1 & 76.8 & 88.4 & 61.9 & 69.7 & 78.9 & 75.1 & 80.1 \\
\quad$\alpha{=}0.80$     &  \dunderline{1pt}{85.3} &  \dunderline{1pt}{76.9} & 87.9 & 61.4 & 69.3 & 78.5 & 74.8 & 80.0 \\
\quad$\alpha{=}0.85$     &  \dunderline{1pt}{85.3} & 76.7 & 87.4 & 60.9 & 68.8 & 78.1 & 74.4 & 79.8 \\
\quad$\alpha{=}0.90$     &  \dunderline{1pt}{85.3} & 76.4 & 86.8 & 60.3 & 68.4 & 77.3 & 73.8 & 79.5 \\
\quad$\alpha{=}0.95$     &  \dunderline{1pt}{85.3} & 76.2 & 86.1 & 59.5 & 67.7 & 76.8 & 73.3 & 79.3 \\
\quad$\alpha{=}1.00$     & 85.2 & 75.8 & 85.3 & 58.7 & 67.2 & 76.1 & 72.6 & 78.9 \\

\bottomrule
\end{tabular}

\caption{\label{tab:breakdown1}
WiSE-FT accuracy on the reference and shifted distributions for various values of the mixing coefficient $\alpha$. Results shown for CLIP \texttt{ViT-L/14@336}. Note that $\alpha{=}0.0$ corresponds to the zero-shot model, while $\alpha=1.0$ corresponds to standard fine-tuning. \textit{Avg shifts} displays the mean performance among the five distribution shifts, while \textit{Avg reference, shifts} shows the average of ImageNet (reference) and Avg shifts.
}
\end{center}
\end{table*}

\begin{table*}
\setlength\tabcolsep{5.1pt}
\small
\begin{center}
\begin{tabular}{lc|ccccc|cc}
\toprule
{} &            &             \multicolumn{5}{c|}{Distribution shifts}             & Avg &     Avg\\
{} &           IN (ref.) &             IN-V2 &              IN-R &                 IN-Sketch &                 ObjectNet &              IN-A & shifts &     ref., shifts\\
\midrule
WiSE-FT, end-to-end & & & & & & & &\\
\quad$\alpha{=}0.00$ & 68.3 & 61.9 & 77.6 & 48.2 & 53.0 & 49.8 &              
     58.1 & 63.2 \\
\quad$\alpha{=}0.05$ & 70.7 & 64.0 & 78.6 & 49.6 & 54.5 & 51.5 &              
     59.6 & 65.2 \\
\quad$\alpha{=}0.10$ & 72.9 & 65.7 & 79.4 & 50.8 & 55.7 & 52.5 &              
     60.8 & 66.8 \\
\quad$\alpha{=}0.15$ & 74.8 & 67.2 & 79.9 & 51.7 & 56.6 & 53.5 &              
     61.8 & 68.3 \\
\quad$\alpha{=}0.20$ & 76.4 & 68.7 &  \dunderline{1pt}{80.1} & 52.5 & 57.1 & 54.2 &              
     62.5 & 69.5 \\
\quad$\alpha{=}0.25$ & 77.8 & 69.9 &  \dunderline{1pt}{80.1} & 53.1 & 57.4 &  \dunderline{1pt}{54.6} &              
     63.0 & 70.4 \\
\quad$\alpha{=}0.30$ & 78.9 & 70.6 &  \dunderline{1pt}{80.1} & 53.6 & 57.5 &  \dunderline{1pt}{54.6} &              
     63.3 & 71.1 \\
\quad$\alpha{=}0.35$ & 79.7 & 71.5 & 79.9 & 53.9 & 57.6 & 54.3 &              
     63.4 & 71.5 \\
\quad$\alpha{=}0.40$ & 80.5 & 72.1 & 79.6 &  \dunderline{1pt}{54.1} &  \dunderline{1pt}{57.7} & 53.8 &  \dunderline{1pt}{63.5} & 72.0 \\
\quad$\alpha{=}0.45$ & 81.2 & 72.4 & 79.3 & 54.0 & 57.5 & 53.2 &              
     63.3 & 72.2 \\
\quad$\alpha{=}0.50$ & 81.7 & 72.8 & 78.7 & 53.9 & 57.3 & 52.2 &              
     63.0 &  \dunderline{1pt}{72.3} \\
\quad$\alpha{=}0.55$ & 82.1 & 73.0 & 78.0 & 53.8 & 56.6 & 51.4 &              
     62.6 &  \dunderline{1pt}{72.3} \\
\quad$\alpha{=}0.60$ & 82.4 & 72.9 & 77.2 & 53.4 & 56.2 & 50.0 &              
     61.9 & 72.2 \\
\quad$\alpha{=}0.65$ &  \dunderline{1pt}{82.6} & 73.1 & 76.3 & 53.0 & 55.5 & 48.9 &              
     61.4 & 72.0 \\
\quad$\alpha{=}0.70$ &  \dunderline{1pt}{82.6} &  \dunderline{1pt}{73.2} & 75.2 & 52.4 & 55.0 & 47.4 &              
     60.6 & 71.6 \\
\quad$\alpha{=}0.75$ &  \dunderline{1pt}{82.6} & 73.1 & 73.9 & 51.8 & 54.3 & 46.0 &              
     59.8 & 71.2 \\
\quad$\alpha{=}0.80$ & 82.5 & 72.8 & 72.7 & 51.0 & 53.5 & 44.6 &              
     58.9 & 70.7 \\
\quad$\alpha{=}0.85$ & 82.3 & 72.4 & 71.1 & 50.0 & 52.7 & 42.9 &              
     57.8 & 70.0 \\
\quad$\alpha{=}0.90$ & 82.1 & 72.0 & 69.5 & 48.9 & 51.7 & 40.9 &              
     56.6 & 69.3 \\
\quad$\alpha{=}0.95$ & 81.7 & 71.5 & 67.7 & 47.6 & 50.7 & 38.8 &              
     55.3 & 68.5 \\
\quad$\alpha{=}1.00$ & 81.3 & 70.9 & 65.6 & 46.3 & 49.6 & 36.7 &              
     53.8 & 67.5 \\
\midrule
WiSE-FT, linear classifier & & & & & & & &\\
\quad$\alpha{=}0.00$     & 68.4 & 62.6 & 77.6 & 48.2 & 53.8 & 50.0 & 58.4 & 63.4 \\
\quad$\alpha{=}0.05$     & 69.9 & 63.7 & 77.9 & 48.9 & 54.2 & 50.6 & 59.1 & 64.5 \\
\quad$\alpha{=}0.10$     & 71.3 & 64.8 & 78.2 & 49.5 & 54.7 & 51.0 & 59.6 & 65.5 \\
\quad$\alpha{=}0.15$     & 72.5 & 65.8 &  \dunderline{1pt}{78.4} & 50.0 & 55.1 & 51.1 & 60.1 & 66.3 \\
\quad$\alpha{=}0.20$     & 73.6 & 66.6 &  \dunderline{1pt}{78.4} & 50.5 & 55.3 & 51.5 & 60.5 & 67.0 \\
\quad$\alpha{=}0.25$     & 74.7 & 67.4 &  \dunderline{1pt}{78.4} & 50.8 & 55.3 &  \dunderline{1pt}{51.8} & 60.7 & 67.7 \\
\quad$\alpha{=}0.30$     & 75.6 & 68.0 & 78.3 & 51.1 & 55.4 & 51.7 & 60.9 & 68.2 \\
\quad$\alpha{=}0.35$     & 76.4 & 68.8 & 78.2 &  \dunderline{1pt}{51.3} &  \dunderline{1pt}{55.5} & 51.6 &  \dunderline{1pt}{61.1} & 68.8 \\
\quad$\alpha{=}0.40$     & 77.1 & 69.0 & 77.8 &  \dunderline{1pt}{51.3} &  \dunderline{1pt}{55.5} & 51.4 & 61.0 & 69.0 \\
\quad$\alpha{=}0.45$     & 77.7 & 69.4 & 77.6 &  \dunderline{1pt}{51.3} & 55.4 & 51.3 & 61.0 & 69.3 \\
\quad$\alpha{=}0.50$     & 78.2 & 69.9 & 77.2 & 51.2 & 55.3 & 51.2 & 61.0 & 69.6 \\
\quad$\alpha{=}0.55$     & 78.6 & 70.1 & 76.7 & 51.0 & 55.0 & 50.9 & 60.7 & 69.7 \\
\quad$\alpha{=}0.60$     & 79.0 & 70.2 & 76.1 & 50.8 & 54.7 & 50.5 & 60.5 &  \dunderline{1pt}{69.8} \\
\quad$\alpha{=}0.65$     & 79.3 & 70.4 & 75.7 & 50.4 & 54.5 & 50.1 & 60.2 &  \dunderline{1pt}{69.8} \\
\quad$\alpha{=}0.70$     & 79.6 & 70.4 & 75.2 & 50.1 & 54.2 & 49.9 & 60.0 &  \dunderline{1pt}{69.8} \\
\quad$\alpha{=}0.75$     & 79.7 & 70.4 & 74.6 & 49.7 & 53.9 & 49.5 & 59.6 & 69.7 \\
\quad$\alpha{=}0.80$     & 79.8 &  \dunderline{1pt}{70.5} & 73.9 & 49.3 & 53.6 & 49.0 & 59.3 & 69.5 \\
\quad$\alpha{=}0.85$     & 79.9 & 70.4 & 73.2 & 48.7 & 53.3 & 48.6 & 58.8 & 69.3 \\
\quad$\alpha{=}0.90$     &  \dunderline{1pt}{80.0} & 70.3 & 72.4 & 48.1 & 52.8 & 47.8 & 58.3 & 69.2 \\
\quad$\alpha{=}0.95$     & 79.9 & 70.1 & 71.7 & 47.5 & 52.6 & 46.9 & 57.8 & 68.8 \\
\quad$\alpha{=}1.00$     & 79.9 & 69.8 & 70.8 & 46.9 & 52.1 & 46.4 & 57.2 & 68.6 \\
\bottomrule
\end{tabular}
\caption{\label{tab:breakdown2}
WiSE-FT accuracy on the reference and shifted distributions for various values of the mixing coefficient $\alpha$. Results shown for CLIP \texttt{ViT-B/16}. Note that $\alpha{=}0.0$ corresponds to the zero-shot model, while $\alpha=1.0$ corresponds to standard fine-tuning. \textit{Avg shifts} displays the mean performance among the five distribution shifts, while \textit{Avg reference, shifts} shows the average of ImageNet (reference) and Avg shifts.
}
\end{center}
\end{table*}

\FloatBarrier
\clearpage

\begin{figure*}[h]
    \centering
    \includegraphics[width=0.9\textwidth]{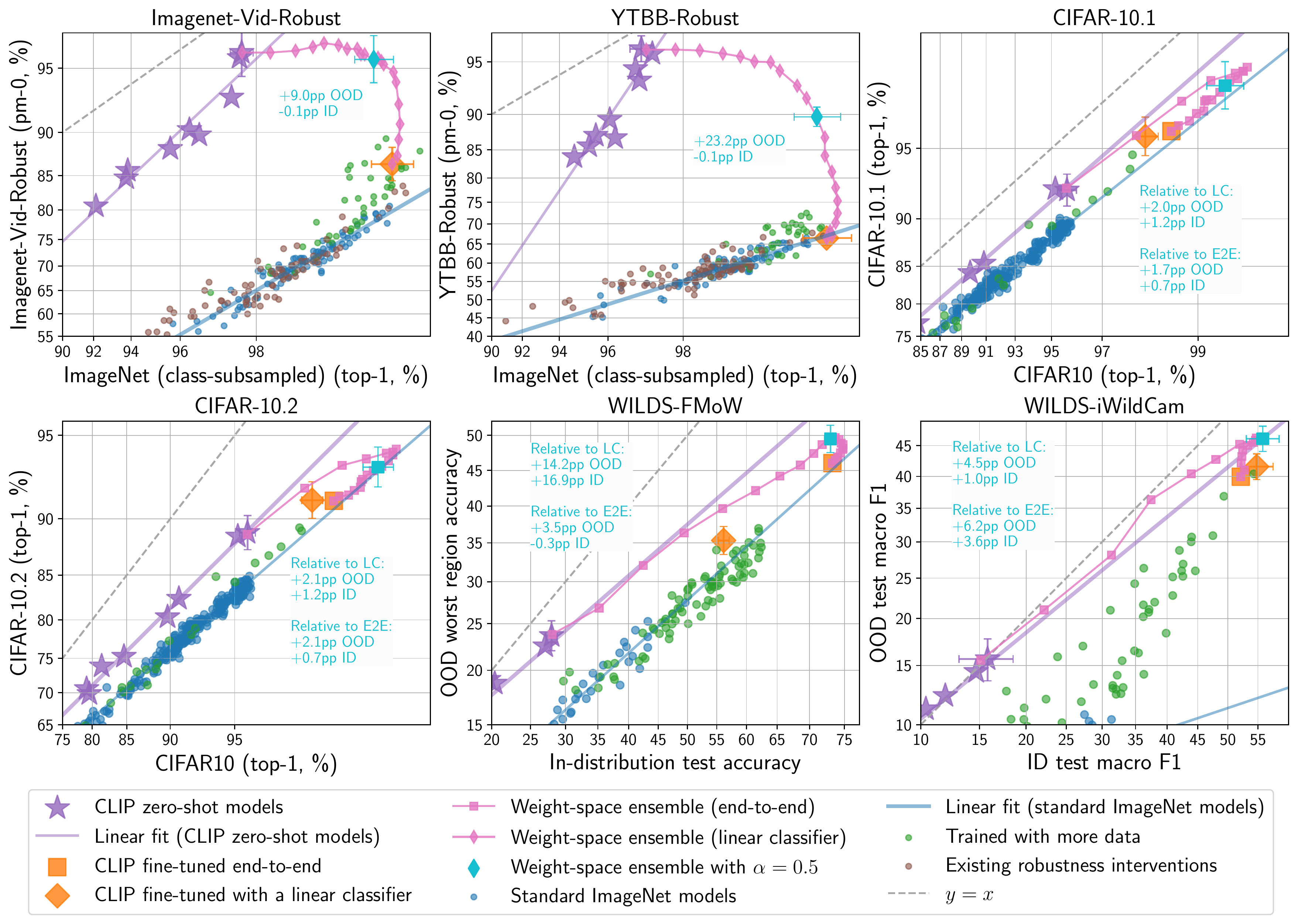}
    \caption{WiSE-FT improves accuracy under distribution shift relative to standard fine-tuning on ImageNet-Vid-Robust, YTBB-Robust \cite{vidrobust}, CIFAR-10.1 \cite{pmlr-v97-recht19a}, CIFAR-10.2 \cite{lu2020harder}, WILDS-FMoW \cite{wilds2021, christie2018functional}, and WILDS-iWildCam \cite{wilds2021, beery2021iwildcam}.} %
    \label{fig:fig_more_shifts}
\end{figure*}

\subsection{Robustness on additional distribution shifts}
\label{sec:moreshift}

Figure~\ref{fig:fig_more_shifts} displays the effective robustness scatter plots for the six additional distribution shifts discussed in Section~\ref{sec:results} (analogous results provided in Table~\ref{tab:moreshifts}).

Concretely, we consider: (i) ImageNet-Vid-Robust and YTBB-Robust, datasets with distribution shift induced by temporal perturbations in videos \cite{vidrobust};
(ii) CIFAR-10.1 \cite{pmlr-v97-recht19a} and CIFAR-10.2 \cite{lu2020harder}, reproductions of the popular image classification dataset CIFAR-10 \cite{krizhevsky2009learning} with a distribution shift;
(iii) WILDS-FMoW, a satellite image recognition task where the test set has a geographic and temporal distribution shift \cite{wilds2021, christie2018functional};
(iv) WILDS-iWildCam, a wildlife recognition task where the test set has a geographic distribution shift \cite{wilds2021, beery2021iwildcam}.

\begin{table*}
\setlength\tabcolsep{5pt}
\small
\begin{center}

\begin{tabular}{lrrrrr}
\toprule

{} &  Zero-shot &  Fine-tuned &  WiSE-FT, $\alpha{=}0.5$ &  WiSE-FT, optimal $\alpha$\\
\midrule
ImageNet-Vid-Robust  (pm-0)           &       95.9 &        86.5 &                  95.5 &               96.5 \\
YTBBRobust (pm-0)                   &       95.8 &        66.5 &                  89.7 &  96.0 \\
CIFAR-10.1 (top-1) &       92.5 &        95.9 &                  97.6 &               98.0 \\
CIFAR-10.2 (top-1) &       88.8 &        91.3 &                  93.4 &                   94.4 \\
WILDS-FMoW: ID test (accuracy)                &       28.0 &        73.3 &                  73.0 &           74.8 \\
WILDS-FMoW: OOD worst region accuracy       &       23.8 &        46.0 &                  49.5 &                49.7 \\
WILDS-iWildCam: ID test macro F1       &       15.1 &        52.1 &                  55.8 & 55.8 \\
WILDS-iWildCam: OOD test macro F1     &       15.5 &        39.9 &                  46.1 &                 46.4 \\
\bottomrule
\end{tabular}
\caption{\label{tab:moreshifts}
WiSE-FT improves results on ImageNet-Vid-Robust, YTBB-Robust \cite{vidrobust}, CIFAR-10.1 \cite{pmlr-v97-recht19a}, CIFAR-10.2 \cite{lu2020harder}, WILDS-FMoW \cite{wilds2021, christie2018functional}, and WILDS-iWildCam \cite{wilds2021, beery2021iwildcam}. Reported numbers are percentages. This is the corresponding table for Figure~\ref{fig:fig_more_shifts}. This table displays results for fine-tuning only a linear classifier for ImageNet-Vid-Robust and YTBBRobust and end-to-end fine-tuning for the remainder. }
\end{center}
\end{table*}

\subsection{Comparison with alternative methods}
\label{sec:baselines-appendix}

We now extend Section~\ref{sec:results} and compare WiSE-FT to additional methods of fine-tuning. We begin with contrasting the weight-space and output-space ensemble.
Next, we show the that varying the decay parameter of an exponential moving average also moves along the curve produced by WiSE-FT.
Finally, we compare with additional methods when fine-tuning only a linear classifier including distillation and various forms
of regularization.

\subsubsection{Output-space ensembles}

\begin{figure*}
    \centering
    \includegraphics[width=0.7\textwidth]{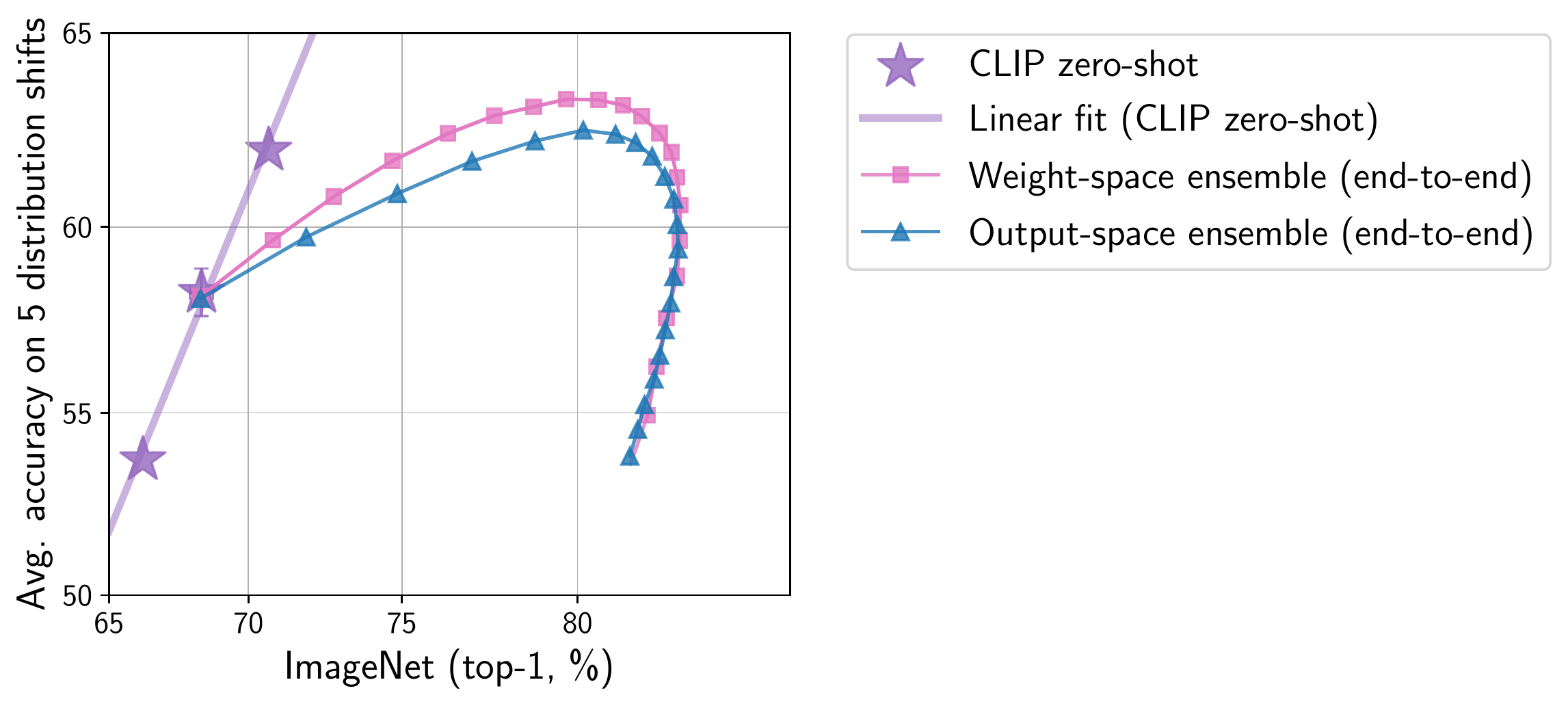}
    \caption{Comparing the weight-space ensemble $f\mleft(x, (1-\alpha)\cdot \theta_0 + \alpha \cdot\theta_1 \mright)$ with the output-space ensemble $(1-\alpha)f\mleft(x,  \theta_0\mright) + \alpha \cdot f\mleft(x, \theta_1 \mright)$ when fine-tuning end-to-end with learning rate $3 \cdot 10^{-5}$. Note that the output-space ensemble requires 2x compute.} %
    \label{fig:ose}
\end{figure*}

 Figure~\ref{fig:ose} compares the weight-space ensemble
  $f\mleft(x, (1-\alpha)\cdot \theta_0 + \alpha \cdot\theta_1 \mright)$ with the output-space ensemble $(1-\alpha)f\mleft(x,  \theta_0\mright) + \alpha \cdot f\mleft(x, \theta_1 \mright)$. 
  Both exhibit a favorable trend, though the output-space ensemble requires twice as much compute. Section~\ref{sec:ntk} further explores the relation between the weight-space and output-space ensemble.

\FloatBarrier
\clearpage

\begin{figure*}
    \centering
    \includegraphics[width=0.75\textwidth]{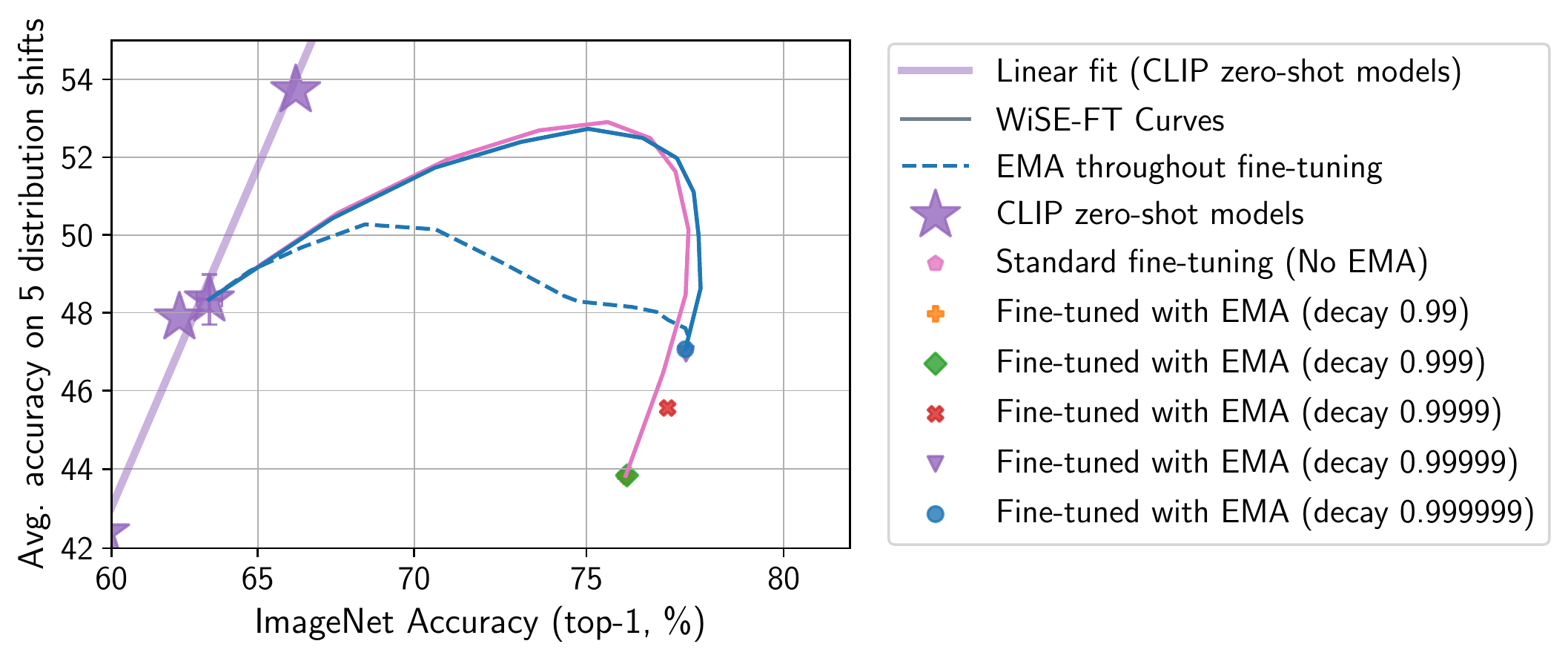}
    \caption{Results for the debiased variant of EMA described in Appendix~\ref{sec:ema}. EMA improves accuracy on both ImageNet and on the distribution shifts, and further applying WiSE-FT to EMA solutions can improve robustness.
    The solutions with no EMA, decay 0.99, and decay 0.999 are overlapping in the plot, as are the solutions with decay 0.99999 and 0.999999.
    }%
    \label{fig:ema0}
\end{figure*}

\begin{figure*}
    \centering
    \includegraphics[width=\textwidth]{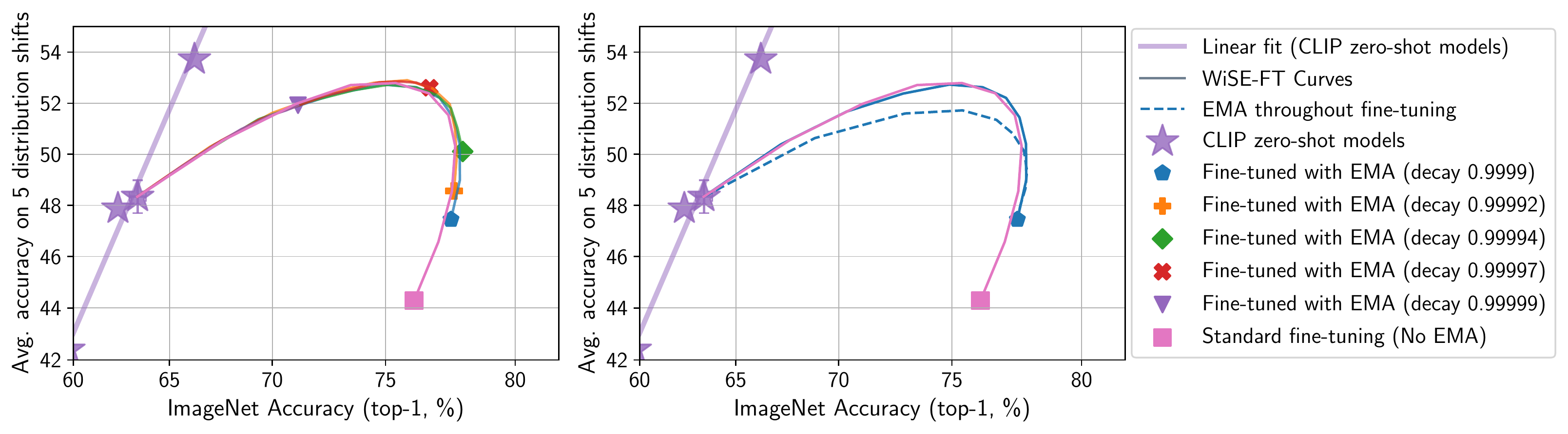}
    \caption{Results for the variant of EMA biased towards the initialization, described in Appendix~\ref{sec:ema}.
    Varying the EMA decay $\beta$ moves
    along the curve produced by WiSE-FT. Applying WiSE-FT to EMA solutions moves further along the curve
    produced by WiSE-FT.
    }%
    \label{fig:ema1}
\end{figure*}

\subsubsection{Comparison to exponential moving averages}\label{sec:ema}

Weight-averaging along the trajectory can improve the performance of models.
For instance, \citet{szegedy2016rethinking} use a running average of the model parameters for their Inception-v2 model.
The exponential moving average (EMA) is a standard technique for keeping a running average of model parameters and
is implemented in libraries such as Optax~\cite{optax2020github} and Pytorch ImageNet Models~\cite{rw2019timm}.

This section explores two variants of EMA for model parameters $\theta \in \mathbb{R}^n$. The first variant is a debiased EMA, where debiasing is done as in \citet{kingma2014adam} (Algorithm 1).
For each iteration $t \in \{1,...,T\}$ let $\theta_t \in \mathbb{R}^n$ be the model parameters at step $t$ and let $\mu_t \in \mathbb{R}^n$ be the EMA at step $t$. For $t = 0$, $\mu_0 \gets 0$, otherwise $\mu_t \gets \beta \cdot \mu_{t - 1} + (1-\beta) \cdot \theta_t$ where $\beta$ is a decay hyperparameter. 
The final debiased EMA is given by $\mu_T / (1 - \beta^T)$. Results for various decay hyperparameters are illustrated by Figure~\ref{fig:ema0}.

Next, we explore a variant of EMA that is biased towards the initialization $\theta_0$. As before, $\mu_t \gets \beta \cdot \mu_{t - 1} + (1-\beta) \cdot \theta_t$. However $\mu_0$ is now initialized to be $\theta_0$, instead of zeros. Moreover, at the end of fine-tuning we use the biased estimate $\mu_T$. Results for this variant are illustrated by Figure~\ref{fig:ema1}.

Section~\ref{sec:main_results} (Figure~\ref{fig:hparams}) showed that decreasing learning rate, training epochs, or early stopping leads to solutions that lie below the curve produced by WiSE-FT.
On the other hand, using an exponential moving average (EMA) and varying the EMA decay $\beta$ can move along or slightly outside or along the curve produced
by WiSE-FT.
For instance, solutions using the second EMA variant follow the WiSE-FT curve.
Indeed, applying WiSE-FT with mixing coefficient $1-\beta^T$ to the debiased EMA variant exactly recovers the second EMA variant described above.
Moreover, further applying WiSE-FT to EMA solutions (i.e., interpolating the weights of the zero-shot model with the EMA solution)
can lead to additional robustness. %
We also evaluate EMA along the fine-tuning trajectory,
finding improved performance under distribution shift for the variant biased towards the initialization.
For the debiased EMA, each model along the trajectory is debiased by $1 / (1 - \beta^t)$.
As shown in Figures~\ref{fig:ema0},\ref{fig:ema1}, evaluations along the trajectory underperform solutions generated by applying WiSE-FT.

\subsubsection{Additional comparisons when fine-tuning a linear classifier}
\label{sec:subsec_additional}

\begin{figure}
    \centering
    \includegraphics[width=\textwidth]{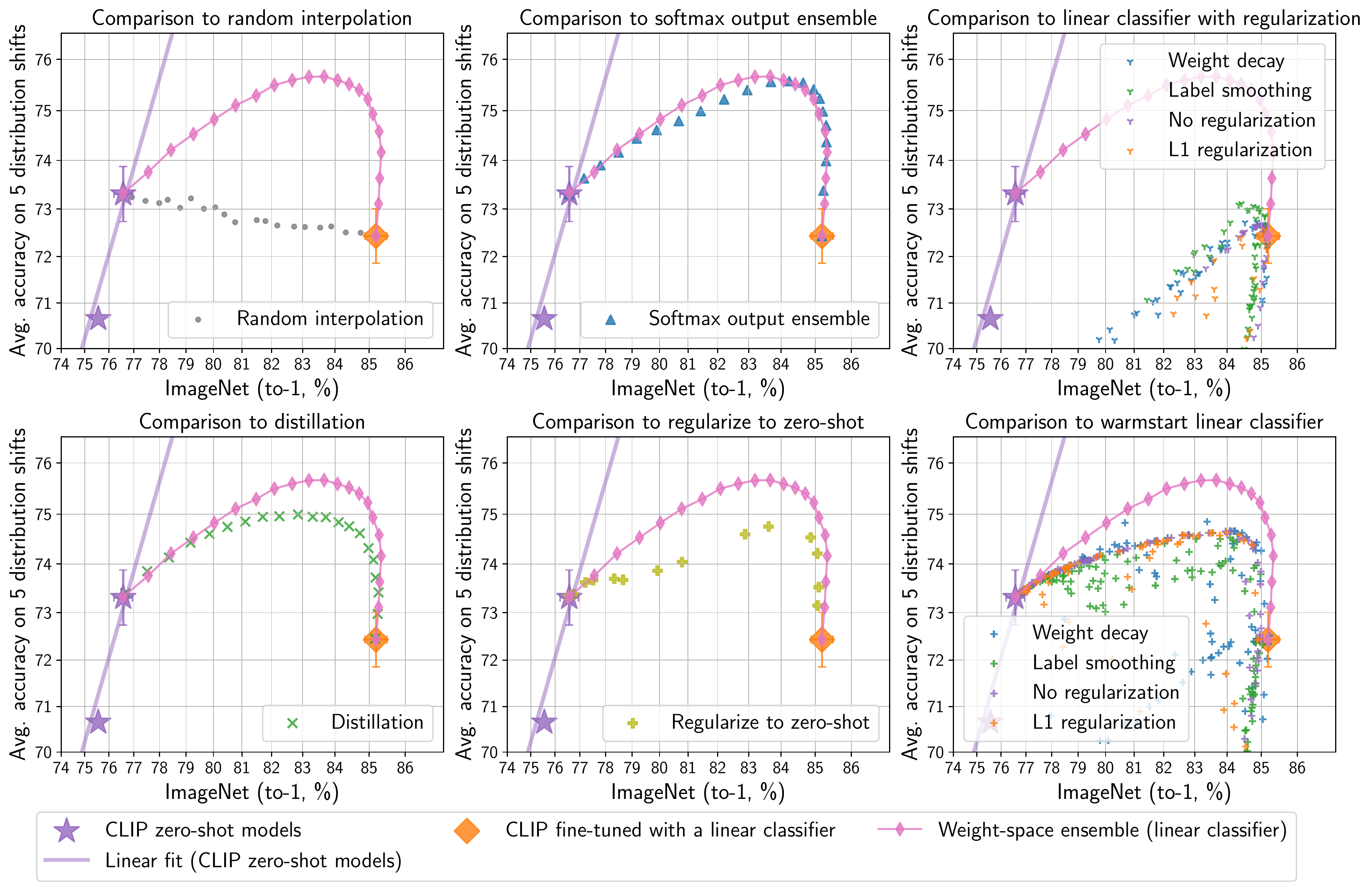}
    \caption{Accuracy on the reference and shifted distributions of WiSE-FT and the alternatives described in Section~\ref{sec:subsec_additional}.}
    \label{fig:baselines}
\end{figure}

We compare against several additional alternatives when fine-tuning only a linear classifier. As this setting is computationally cheaper compared to end-to-end, it allows for comprehensive experimentation. Many of the examined approaches exhibit a concave trend in effective robustness plots, although WiSE-FT matches methods requiring more compute or offers better performance (Figure~\ref{fig:baselines}).

\paragraph{Random interpolation.} This method uses either the zero-shot or fine-tuned linear classifier depending on a (biased) coin flip.
For hyperparameter $\alpha \in [0,1]$ outputs are computed as $(1- \xi)\cdot f(x, \theta_0)  + \xi\cdot  f(x, \theta_1)$ where $\xi$ is a Bernoulli$(\alpha)$ random variable.
For this method and all others with a hyperparameter $\alpha \in [0,1]$ we evaluate models for $\alpha \in \{0, 0.05, 0.1, ..., 1\}$.

\paragraph{Ensembling softmax outputs.} Instead of ensembling in weight space, this method combines softmax probabilities assigned by the zero-shot and fine-tuned linear classifier.
Concretely, for hyperparameter $\alpha \in [0,1]$ outputs are computed as $ (1-\alpha)\cdot \mathsf{softmax} \mleft( f(x, \theta_0) \mright)  + \alpha\cdot  \mathsf{softmax} \mleft( f(x, \theta_1) \mright)$.
This method performs comparably to weight-space ensembling but requires slightly more compute.

\paragraph{Linear classifier with various regularizers.} We explore fine-tuning linear classifiers with four regularization strategies: no regularization, weight decay, L1 regularization, and label smoothing \cite{muller2019does}. Linear classifiers are trained with mini-batch optimization, using the AdamW optimizer \cite{loshchilov2018decoupled, paszke2019pytorch} with a cosine-annealing learning rate schedule \cite{loshchilov2016sgdr}.
This method is significantly faster and less memory-intensive than the L-BFGS implementation used by \citet{radford2021learning} at ImageNet scale with similar accuracy.
Additional details on hyperparameters and more analyses are provided in Appendix~\ref{sec:moreclf}. 

Two variants of this method are shown in Figure~\ref{fig:baselines}, one for which the the linear classifier is initialized randomly and another for which the linear classifier is initialized with the zero-shot weights (denoted \emph{warmstart}). If the convex problem is solved then the initialization does not play a role. However we are using mini-batch optimization and, in certain cases, terminating training before an optimum is reached.

\FloatBarrier

\paragraph{Distillation.} Network distillation \cite{hinton2015distilling} trains one network to match the outputs of another.
We use this technique to fine-tune while matching the outputs of the zero-shot model with weights $\theta_0$.
For a hyperparameter $\alpha \in [0,1]$ and cross-entropy loss $\ell$ we fine-tune $\theta$ according to the minimization  objective
\begin{align} \label{eq:distill}
     \sum_{(x_i, y_i) \in \mathcal{S}^\text{tr}_\mathcal{D}} (1- \alpha) \cdot \ell\mleft(f(x_i,\theta), y_i\mright) + \alpha \cdot \ell\mleft( f(x_i,\theta), f(x_i,\theta_0)\mright)\ .
\end{align}

\paragraph{Regularization towards zero-shot.} We train a linear classifier with an additional regularization term which penalizes movement from the zero-shot classifier's weights.
For a hyperparameter $\lambda \in \{1\cdot10^{-8}, 5\cdot10^{-8}, 1\cdot10^{7},..., 5\cdot10^{2}\}$ we add the regularization term $\lambda \left\| \mathbf{W} - \mathbf{W}_\text{zero-shot} \right\|_F^2$ where $\mathbf{W}$ is the linear classifier being fine-tuned.
In most cases this method performs slightly worse than distillation.

\begin{figure*}
    \centering
    \includegraphics[width=.9\textwidth]{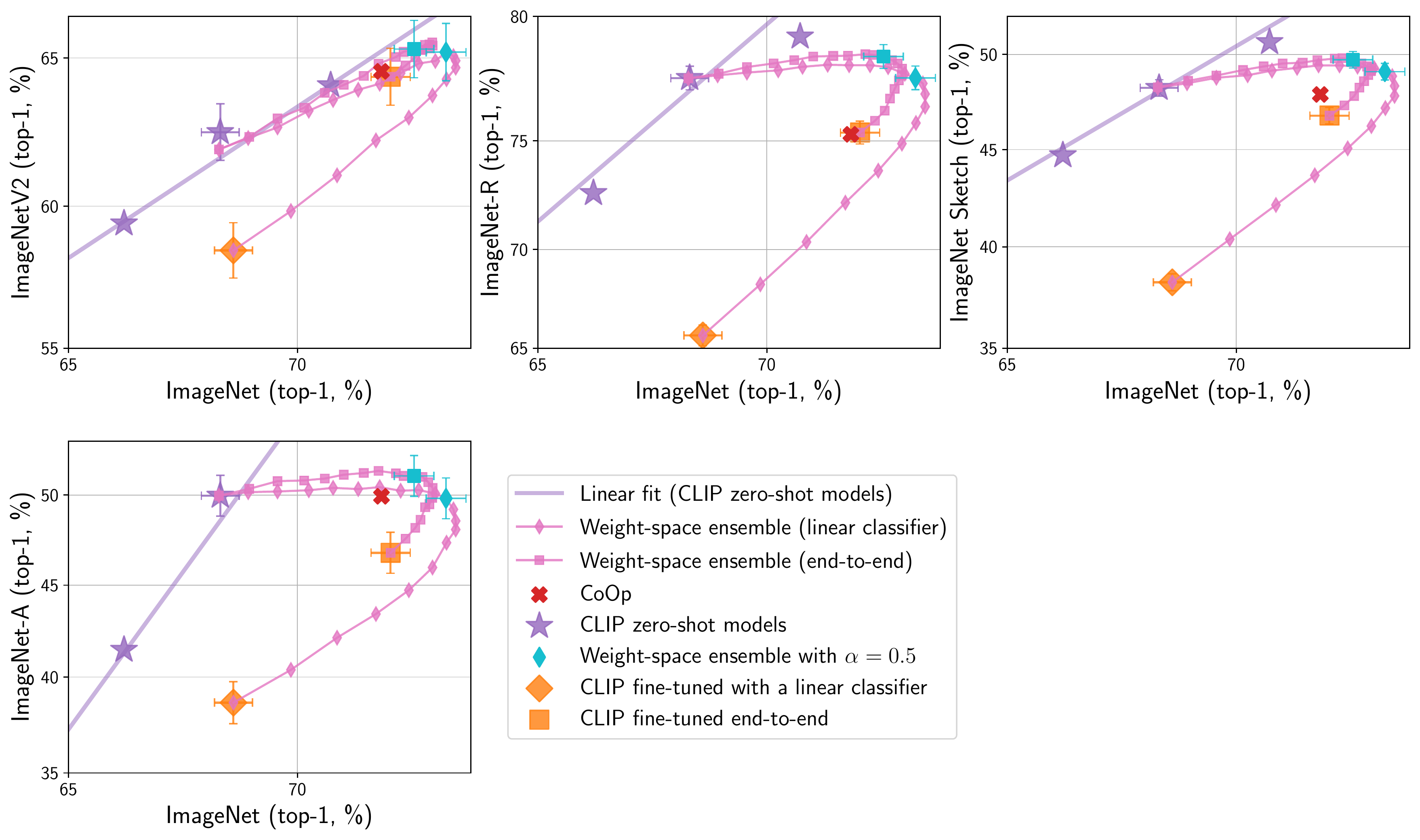}
    \caption{Comparing WiSE-FT with CoOp \cite{coop}. Both methods fine-tune the \texttt{ViT-B/16} CLIP model on 16 examples per class of ImageNet.}
    \label{fig:coop}
\end{figure*}

Finally, Figure~\ref{fig:coop} and Table \ref{tab:coop} demonstrate that WiSE-FT achieves better accuracy than the recently proposed CoOp method \cite{coop} on ImageNet and four derived distribution shifts.
Instead of fine-tuning network parameters, CoOp instead learns continuous embedding for the language prompts. We note that CoOp and WiSE-FT could be used in conjunction in future work. We compare with the \texttt{ViT-B/16} section in Table 7 of \citet{coop}. For comparison we use the same CLIP model as CoOp and also train only on 16 images per class. When end-to-end fine-tuning we use 10 epochs and learning rate $10^{-5}$. %

\begin{table*}[]
    \centering
\begin{tabular}{lccccc}
\toprule
{} &  ImageNet (IN) &  INV2 &  IN-R &  IN-A &  IN Sketch \\
\midrule
CoOp \cite{coop} &     71.73 &       64.56 &       75.28 &       49.93 &            47.89 \\
WiSE-FT (linear classifiere, $\alpha = 0.5$)   &     73.02 &       65.19 &       77.63 &       49.81 &            49.09 \\
WiSE-FT (end-to-end, $\alpha = 0.5$)  &     72.38 &       65.29 &       78.47 &       51.07 &            49.72 \\
\bottomrule
\end{tabular}
    \caption{Comparing WiSE-FT with CoOp \cite{coop}. Both methods fine-tune the \texttt{ViT-B/16} CLIP model on 16 examples per class of ImageNet. Also see Figure~\ref{fig:coop}.}
    \label{tab:coop}
\end{table*}

\subsection{Changes in data augmentation}\label{sec:dataaug}

In the majority of our experiments we follow \citet{radford2021learning} in using minimal data augmentation. However,
Figure~\hyperlink{fig:hparamsaug}{14} recreates Figure~\ref{fig:hparams} with the default ImageNet train augmentation used in PyTorch ImageNet Models \cite{rw2019timm}, which includes random cropping, horizontal flipping and color jitter. As shown in Figure~\hyperlink{fig:hparamsaug}{14}, we find similar trends with this stronger data augmentation. Further investigating the effect of data augmentation remains an interesting direction for future work.

\begin{figure*}
\begin{tikzpicture}
    \draw (0, 0) node[inner sep=0] {\includegraphics[width=\textwidth]{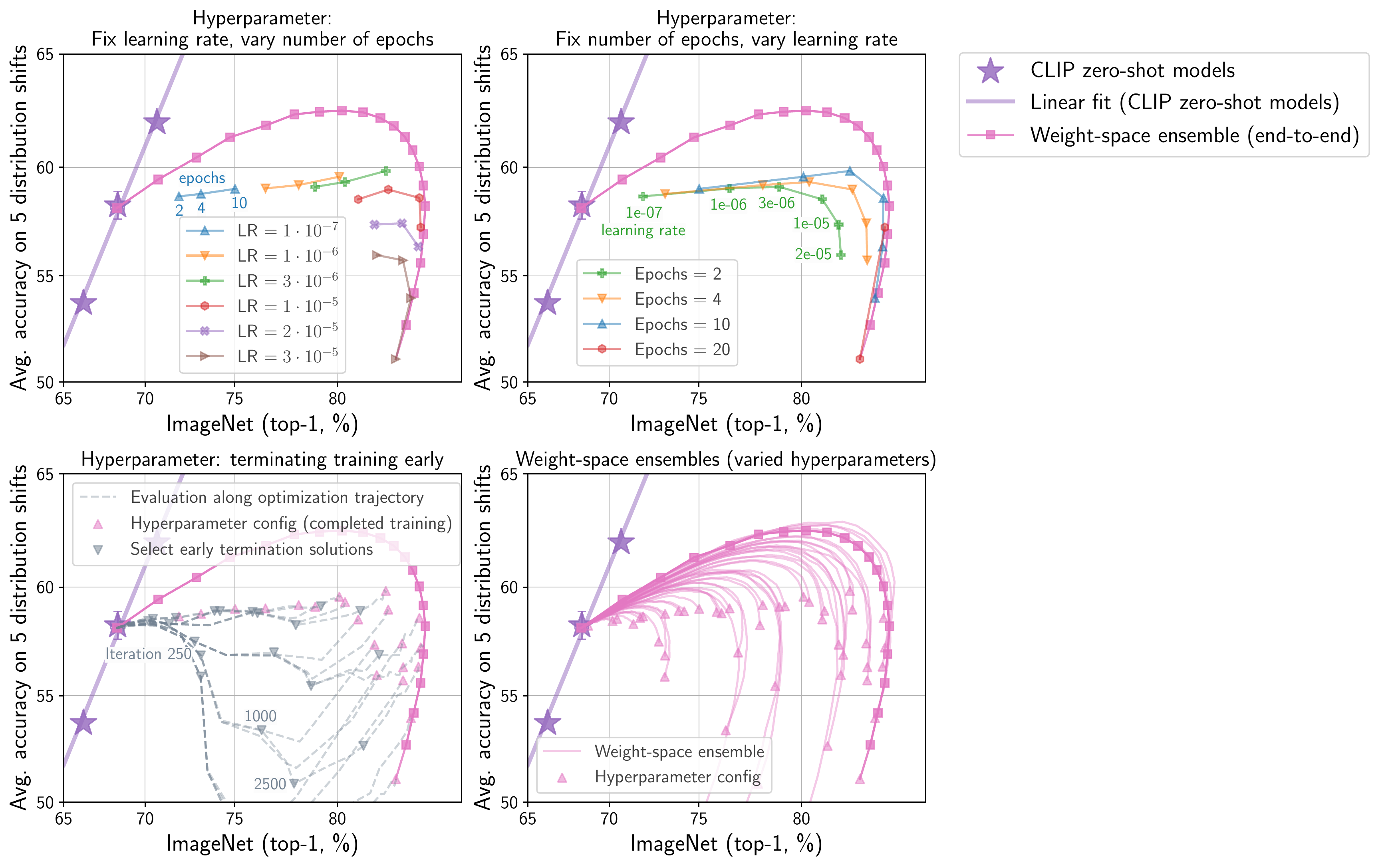}};
    \draw (3., 0) node[right,scale=0.9,inner sep=0pt,outer sep=0pt,text width = 6.3cm] {Figure 14.
    The robustness of fine-tuned models varies substantially under even small changes in hyperparameters.
    Applying WiSE-FT addresses this brittleness and can remove the trade-off between accuracy on the reference and shifted distributions.
    Results shown for CLIP \texttt{ViT-B/16} fine-tuned with cosine-annealing learning rate schedule and ImageNet data augmentation from Pytorch ImageNet Models \cite{rw2019timm}.
    };
\end{tikzpicture}
\captionlistentry{}
\hypertarget{fig:hparamsaug}{}
\end{figure*}

\subsection{Accuracy improvements on reference datasets}
\label{sec:low-data}
\begin{figure*}
    \centering
    \includegraphics[width=\textwidth]{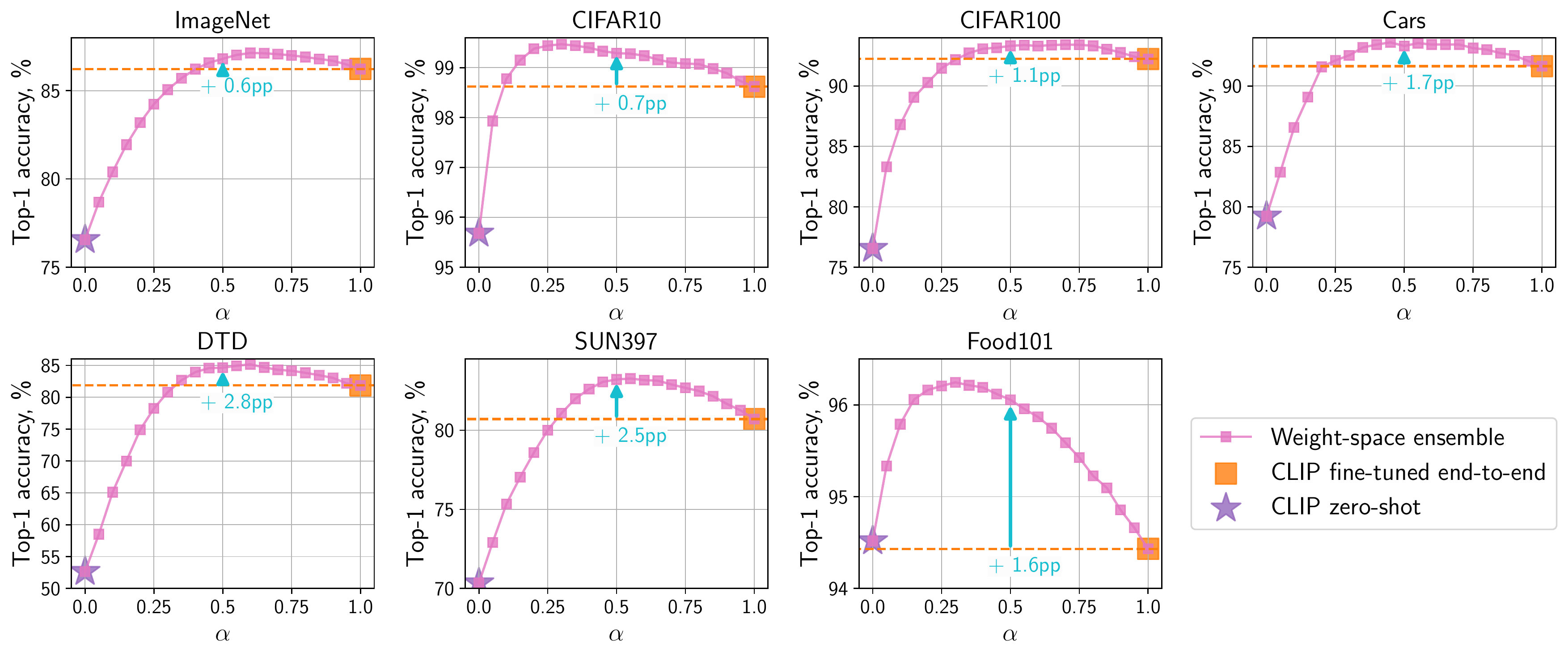}
    \caption{The accuracy of WiSE-FT (end-to-end) with mixing coefficient $\alpha$ on ImageNet and a number of datasets considered by \citet{kornblith2019better}: CIFAR-10, CIFAR-100 \cite{krizhevsky2009learning}, Describable Textures \cite{dtd}, Food-101 \cite{food101}, SUN397 \cite{sun397}, and Stanford Cars \cite{cars}.}
    \label{fig:1d}
\end{figure*}

Beyond robustness, Figure~\ref{fig:1d} demonstrates that WiSE-FT can provide accuracy improvements on ImageNet and a number of datasets considered by \citet{kornblith2019better}: CIFAR-10, CIFAR-100 \cite{krizhevsky2009learning}, Describable Textures \cite{dtd}, Food-101 \cite{food101}, SUN397 \cite{sun397}, and Stanford Cars \cite{cars}.
This is surprising as standard fine-tuning optimizes for low error on the reference distribution.
Figure~\ref{fig:1d} supplements Table~\ref{tab:id_gains} by providing accuracy information for all mixing coefficients $\alpha$.

\begin{figure*}
    \centering
    \includegraphics[width=\textwidth]{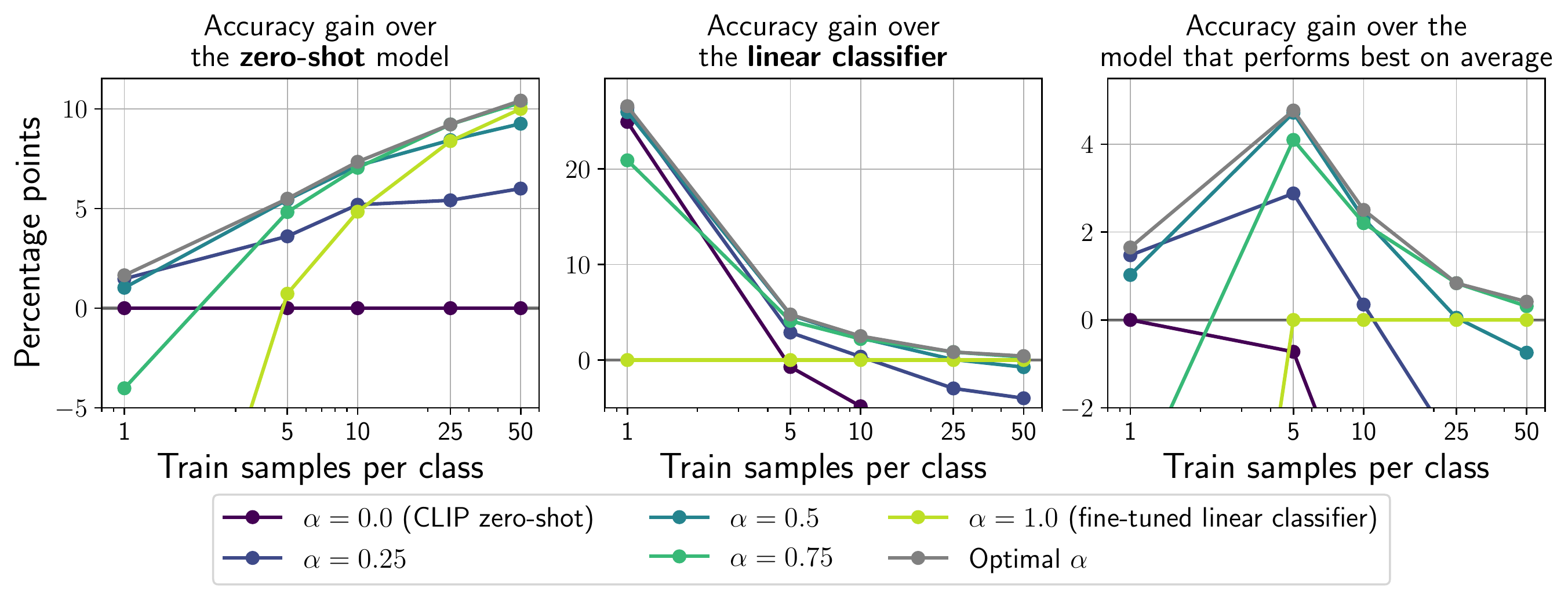}
    \caption{WiSE-FT can improve accuracy over the linear classifier and zero-shot model in the low data regime. On the $x$-axis we consider $k = \{1,5,10,25,50\}$ examples per class for fine-tuning. On the $y$-axis we display accuracy improvements of WiSE-FT averaged over seven datasets \cite{deng2009imagenet,krizhevsky2009learning,dtd,food101,sun397,cars}. For $k=1$, the zero-shot model outperforms the fine-tuned linear classifier, and ensembles closer to the zero-shot model (small $\alpha$) yield high performance. When more data is available, the reverse is true, and higher values of $\alpha$ improve performance. Figures \ref{fig:figntrain} and \ref{fig:figntrainv2} display a breakdown for all datasets.}
    \label{fig:figntrain_avg}
\end{figure*}

\begin{figure*}
    \centering
    \includegraphics[width=\textwidth]{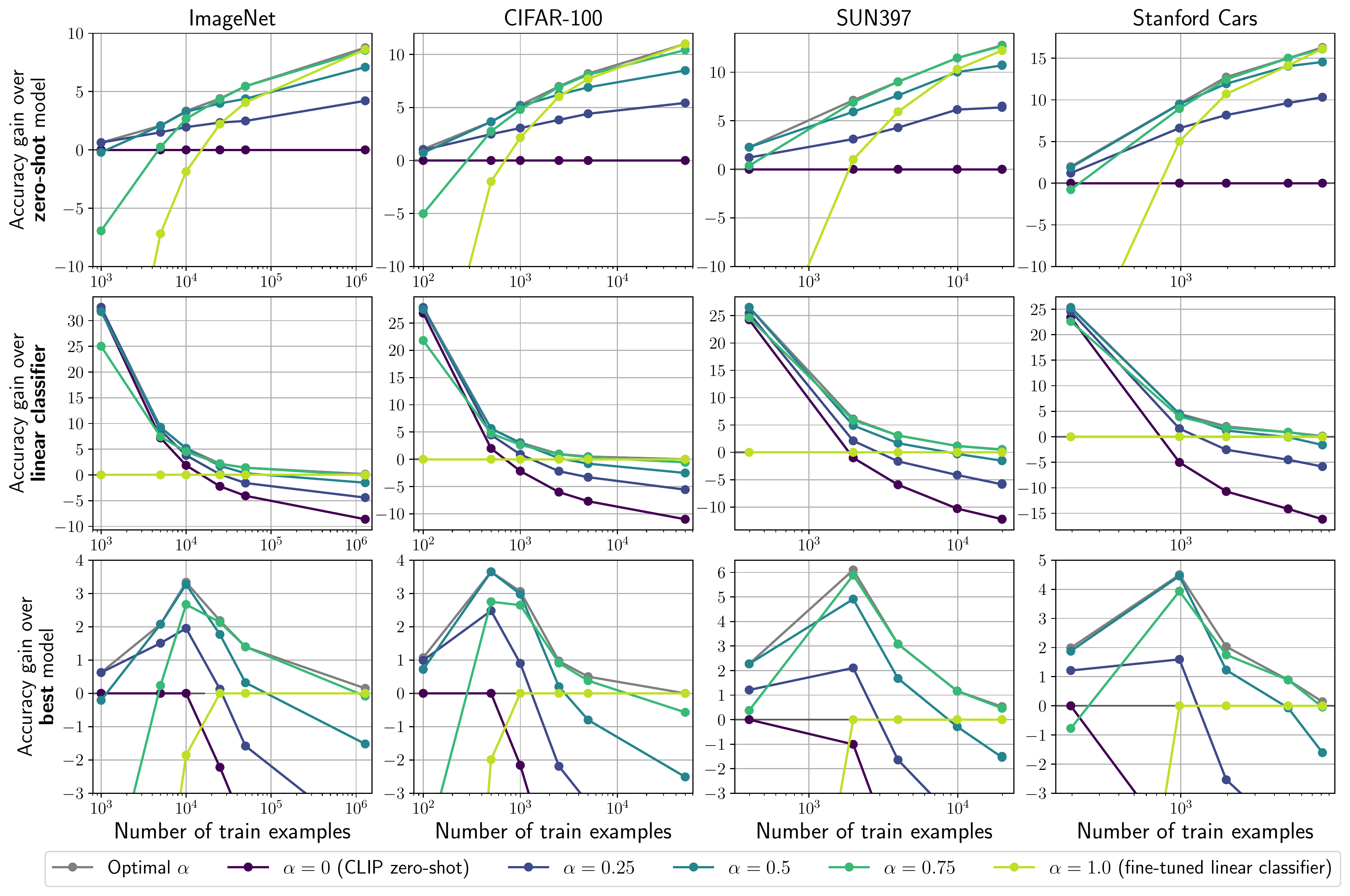}
    \caption{WiSE-FT improves accuracy over the linear classifier and zero-shot model in the low data regime. On the $x$-axis we consider $k = \{1,5,10,25,50\}$ examples per class and the full training set. On the $y$-axis we consider the accuracy improvement of WiSE-FT over the \textbf{(top)} zero-shot model, \textbf{(middle)} fine-tuned linear classifier, and \textbf{(bottom)} best of the zero-shot and fine-tuned linear classifier.}
    \label{fig:figntrain}
\end{figure*}

\begin{figure*}
    \centering
    \includegraphics[width=0.85\textwidth]{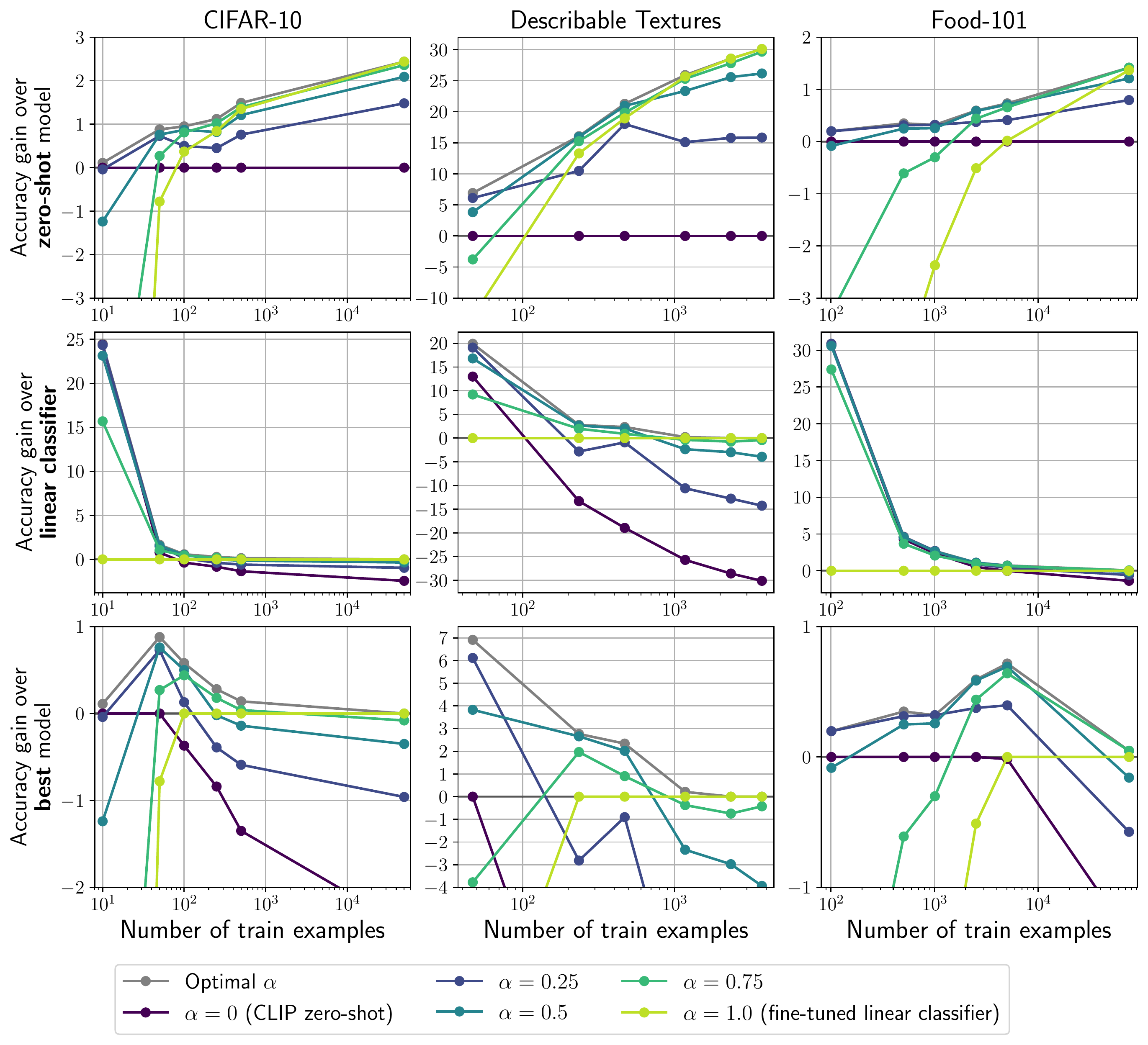}
    \caption{WiSE-FT improves accuracy over the linear classifier and zero-shot model in the low data regime. On the $x$-axis we consider $k = \{1,5,10,25,50\}$ examples per class and the full training set. On the $y$-axis we consider the accuracy improvement of WiSE-FT over the \textbf{(top)} zero-shot model, \textbf{(middle)} fine-tuned linear classifier, and \textbf{(bottom)} best of the zero-shot and fine-tuned linear classifier.}
    \label{fig:figntrainv2}
\end{figure*}

In many application-specific scenarios, only a small amount of data is available for fine-tuning.
Accordingly, we examine the performance of WiSE-FT when only $k$ examples per class are used for fine-tuning on the seven aforementioned datasets ($k = \{1,5,10,25,50\}$). In contrast with Figure~\ref{fig:1d}, we now fine-tune only the linear classifier allowing for comprehensive experiments. 
Average results are shown in Figure~\ref{fig:figntrain_avg}, while Figures~\ref{fig:figntrain} and \ref{fig:figntrainv2} provide a breakdown for all datasets.

\FloatBarrier
\clearpage

\begin{figure}[h]
    \centering
    \includegraphics[width=\textwidth]{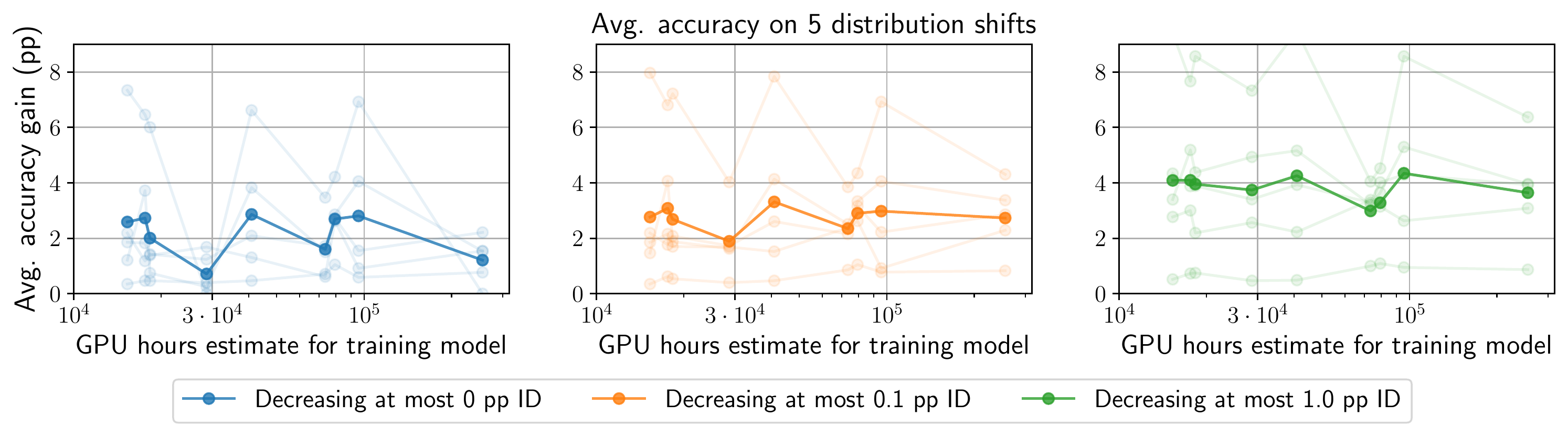}
    \caption{WiSE-FT provides benefits for all CLIP models. Accuracy under distribution shift can be improved relative to the linear classifier with less than $\epsilon \in \{0, 0.1, 1\}$ percentage points (pp) loss in accuracy on the reference distribution, across orders of magnitude of training compute. The CLIP model \texttt{RN50x64} requires the most GPU hours to train.}
    \label{fig:fig_scale}
    
\end{figure}

\begin{figure}[h]
    \centering
    \includegraphics[width=.85\textwidth]{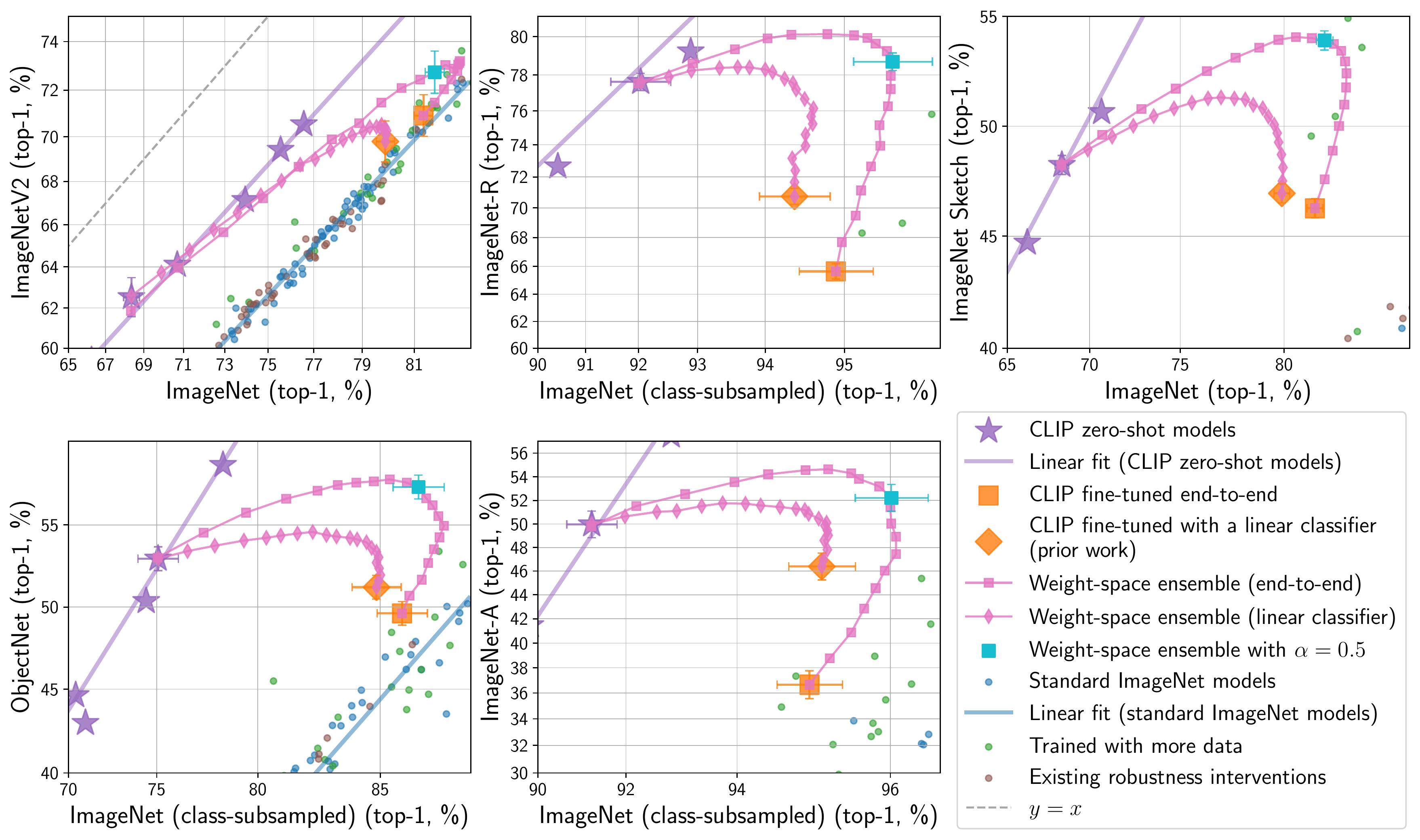}
    \caption{WiSE-FT improves accuracy on the reference and shifted distributions for numerous distribution shifts with a smaller CLIP \texttt{ViT-B/16} model.}
    \label{fig:b16}
\end{figure}

\subsection{Robustness across scales of pre-training compute}
\label{sec:scaling}

The strong correlation between standard test accuracy and accuracy under distribution shift holds from low to high performing models.
This offers the opportunity to explore robustness for smaller, easy to run models.
Our exploration began with the lowest accuracy CLIP models and similar trends held at scale.
Figure~\ref{fig:fig_scale} shows improved accuracy under distribution shift with minimal loss on reference performance across orders of magnitude of pre-training compute with WiSE-FT when fine-tuning a linear classifier.
Moreover, in Figure~\ref{fig:b16} we recreate the experimental results for ImageNet and five associated distribution shifts with a smaller CLIP \texttt{ViT-B/16} model, finding similar trends. Recall that unless otherwise mentioned our experiments use the larger CLIP model (\texttt{ViT-L/14@336px}).

\FloatBarrier

\begin{table*}
\setlength\tabcolsep{5.1pt}
\small
\begin{center}
\begin{tabular}{lc|ccccc|cc}
\toprule
{} &            &             \multicolumn{5}{c|}{Distribution shifts}             & Avg &     Avg\\
{} &           IN (ref.) &             IN-V2 &              IN-R &                 IN-Sketch &                 ObjectNet &              IN-A & shifts &     ref., shifts\\
\midrule
CLIP \texttt{ViT-B/16} \cite{radford2021learning} &  &  &  &  &  &  &  &  \\
\quad Zero-shot & 68.3 & 61.9 & 77.6 & 48.2 & 53.0 & 49.8 & 58.1 & 63.2 \\
\quad Standard fine-tuning & 81.3 & 70.9 & 65.6 & 46.3 & 49.6 & 36.7 & 53.8 & 67.5 \\
\quad WiSE-FT ($\alpha{=}0.5$) & 81.7 & 72.8 & 78.7 & 53.9 & 57.3 & 52.2 & 63.0 &  72.3 \\
\quad WiSE-FT (opt. $\alpha$) & 82.6 & 73.2 & 80.1 & 54.1 & 57.7 & 54.6 & 63.5 &  72.3 \\
CLIP \texttt{ViT-L/14@336px} \cite{radford2021learning} &  &  &  &  &  &  &  &  \\
\quad Zero-shot & 76.6 & 70.5 & 89.0 & 60.9 & 68.5 & 77.6 & 73.3 & 74.9 \\
\quad Standard fine-tuning & 86.2 & 76.8 & 79.8 & 57.9 & 63.3 & 65.4 &  68.6 & 77.4 \\
\quad WiSE-FT ($\alpha{=}0.5$) & 86.8 &  79.5 & 89.4 & 64.7 & 71.1 & 79.9 & 76.9 & 81.8 \\
\quad WiSE-FT (opt. $\alpha$) & 87.1 & 79.5 & 90.3 & 65.0 & 72.1 & 81.0 & 77.4 &  81.9 \\
ALIGN \cite{jia2021scaling} & & & & & & & & \\
\quad Zero-shot & 76.4 & 70.1 & 92.1 & 67.9 & 67.2 & 75.9 & 74.6 & 75.5 \\
\quad Standard fine-tuning & 88.2 & 80.1 & 88.5 & 69.1 & 61.0 & 76.3 & 75.0 & 81.6 \\
\quad WiSE-FT ($\alpha{=}0.5$) & 86.3 & 79.2 & 93.0 & 71.1 & 67.8 & 81.0 & 78.4 & 82.3 \\
\quad WiSE-FT (opt. $\alpha$) & 88.3 & 80.4 & 93.3 & 71.1 & 68.6 & 81.0 & 78.4 &  82.8 \\
JFT pre-trained \texttt{ViT-H} \cite{dosovitskiy2021an} &  &  &  &  &  &  &  &  \\
\quad Zero-shot & 72.9 & 66.1 & 85.9 & 57.0 & 59.2 & 58.4 & 65.3 & 69.1 \\
\quad Standard fine-tuning & 85.4 & 77.6 & 84.9 & 62.8 & 63.1 & 60.8 & 69.8 & 77.6 \\
\quad WiSE-FT ($\alpha{=}0.5$) & 82.9 & 75.4 &  89.3 & 63.8 & 65.8 & 66.2 & 72.1 & 77.5 \\
\quad WiSE-FT (opt. $\alpha$) & 85.4 & 77.8 & 89.3 & 64.5 & 66.0 & 66.6 & 72.5 &  78.6 \\
BASIC-M \cite{pham2021scaling} &  &  &  &  &  &  &  &  \\
\quad Zero-shot & 81.4 & 74.1 & 90.6 & 67.4 & 73.5 & 66.7 & 74.5 & 78.0 \\
\quad Standard fine-tuning & 86.2 & 77.8 & 84.9 & 64.3 & 75.3 & 63.7 & 73.2 & 79.7 \\
\quad WiSE-FT ($\alpha{=}0.5$) & 85.6 & 78.5 & 90.2 & 68.6 & 78.0 & 71.1 & 77.3 & 81.4 \\
\quad WiSE-FT (opt. $\alpha$) & 86.2 & 78.6 & 91.1 & 68.8 & 78.0 & 71.4 & 77.4 &  81.4 \\
BASIC-L \cite{pham2021scaling} &  &  &  &  &  &  &  &  \\
\quad Zero-shot & 85.6 & 80.5 & 95.7 & 76.2 & 82.3 & 85.7 & 84.1 & 84.8 \\
\quad Standard fine-tuning & 87.5 & 79.8 & 84.3 & 68.0 & 77.4 & 72.1 & 76.3 & 81.9 \\
\quad WiSE-FT ($\alpha{=}0.5$) & 87.9 & 81.6 & 94.5 & 73.6 & 84.1 & 83.2 & 83.4 & 85.7 \\
\quad WiSE-FT (opt. $\alpha$) & \textbf{87.9} & \textbf{82.1} & \textbf{96.0} & \textbf{76.5} & \textbf{84.9} & \textbf{86.5} & \textbf{85.0} &  \textbf{86.2} \\
\bottomrule
\end{tabular}
\caption{\label{tab:merged}
WiSE-FT accuracy on ImageNet and derived distribution shifts for various models fine-tuned end-to-end. \textit{Avg shifts} displays the mean performance among the five distribution shifts, while \textit{Avg reference, shifts} shows the average of ImageNet (reference) and Avg shifts. For optimal $\alpha$, we choose the single mixing coefficient that maximizes the column.
}
\end{center}
\end{table*}

\begin{figure*}
    \centering
    \includegraphics[width=.9\textwidth]{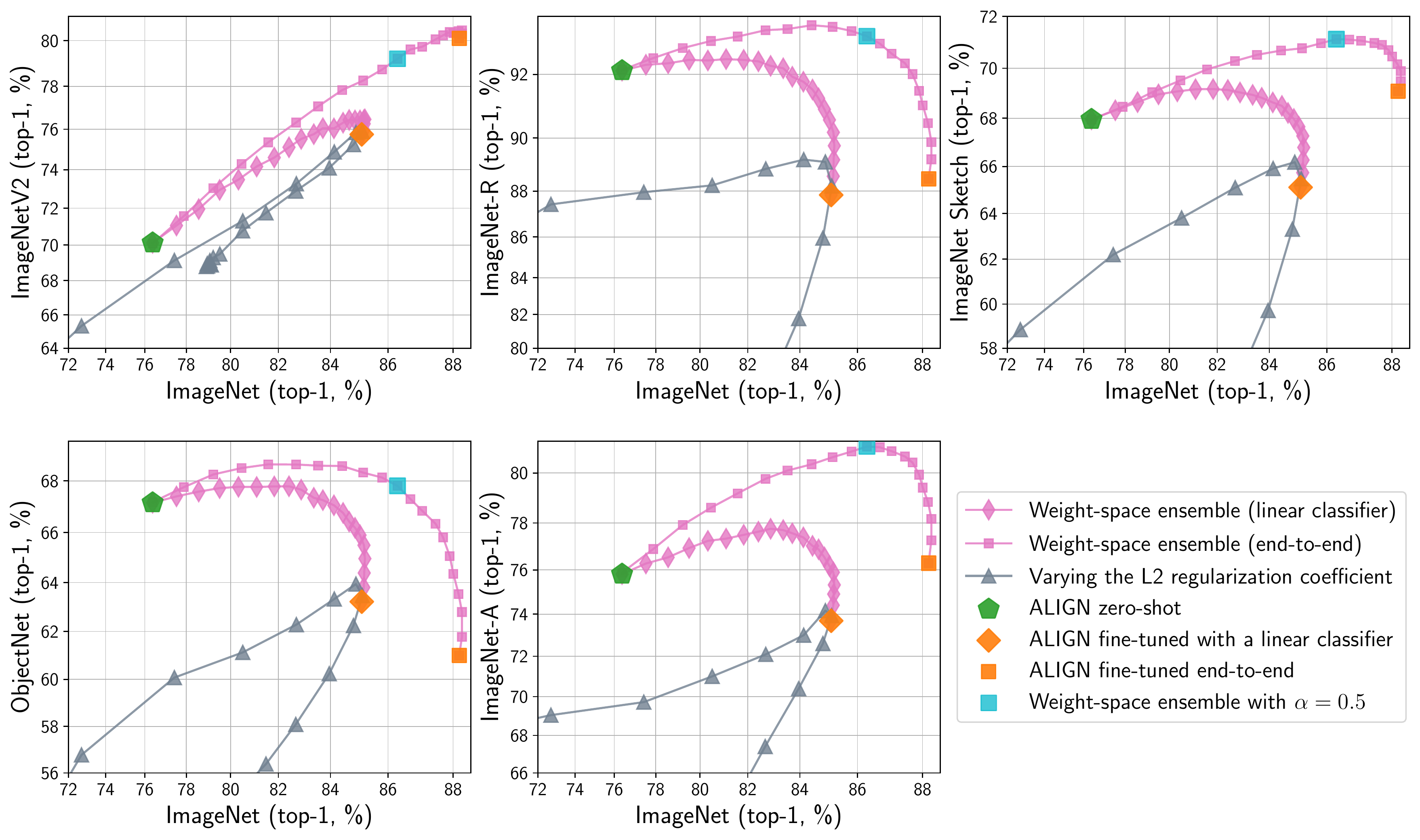}
    \caption{WiSE-FT applied to ALIGN \cite{jia2021scaling}. We also show the effect of varying the L2 regularization strength for linear classifier fine-tuning.}
    \label{fig:align}
\end{figure*}

\begin{table*}
\setlength\tabcolsep{5.1pt}
\small
\begin{center}
\begin{tabular}{lc|ccccc|cc}
\toprule
{} &            &             \multicolumn{5}{c|}{Distribution shifts}             & Avg &     Avg\\
{} &           IN (reference) &             IN-V2 &              IN-R &                 IN-Sketch &                 ObjectNet &              IN-A & shifts &     reference, shifts\\
\midrule
WiSE-FT, end-to-end & & & & & & & &\\
\quad $\alpha{=}0.00$ & 76.4 & 70.1 & 92.1 & 67.9 & 67.2 & 75.9 & 74.6 & 75.5 \\
\quad $\alpha{=}0.05$ & 77.9 & 71.6 & 92.5 & 68.5 & 67.8 & 76.9 & 75.5 & 76.7 \\
\quad $\alpha{=}0.10$ & 79.2 & 73.0 & 92.7 & 69.0 & 68.2 & 77.9 & 76.2 & 77.7 \\
\quad $\alpha{=}0.15$ & 80.5 & 74.3 & 92.9 & 69.5 & 68.5 & 78.6 & 76.8 & 78.7 \\
\quad $\alpha{=}0.20$ & 81.6 & 75.4 & 93.0 & 70.0 &  \dunderline{1pt}{68.6} & 79.2 & 77.2 & 79.4 \\
\quad $\alpha{=}0.25$ & 82.7 & 76.3 & 93.2 & 70.3 &  \dunderline{1pt}{68.6} & 79.8 & 77.6 & 80.2 \\
\quad $\alpha{=}0.30$ & 83.5 & 77.1 & 93.2 & 70.5 &  \dunderline{1pt}{68.6} & 80.1 & 77.9 & 80.7 \\
\quad $\alpha{=}0.35$ & 84.4 & 77.8 &  \dunderline{1pt}{93.3} & 70.7 &  \dunderline{1pt}{68.6} & 80.3 & 78.1 & 81.2 \\
\quad $\alpha{=}0.40$ & 85.2 & 78.3 &  \dunderline{1pt}{93.3} & 70.8 & 68.3 & 80.6 & 78.3 & 81.8 \\
\quad $\alpha{=}0.45$ & 85.8 & 78.8 & 93.2 & 71.0 & 68.1 & 80.8 &  \dunderline{1pt}{78.4} & 82.1 \\
\quad $\alpha{=}0.50$ & 86.3 & 79.2 & 93.0 &  \dunderline{1pt}{71.1} & 67.8 &  \dunderline{1pt}{81.0} &  \dunderline{1pt}{78.4} & 82.3 \\
\quad $\alpha{=}0.55$ & 86.7 & 79.6 & 92.8 &  \dunderline{1pt}{71.1} & 67.3 &  \dunderline{1pt}{81.0} &  \dunderline{1pt}{78.4} & 82.6 \\
\quad $\alpha{=}0.60$ & 87.1 & 79.7 & 92.6 &  \dunderline{1pt}{71.1} & 66.8 & 80.8 & 78.2 & 82.7 \\
\quad $\alpha{=}0.65$ & 87.5 & 80.0 & 92.3 & 71.0 & 66.3 & 80.6 & 78.0 &  \dunderline{1pt}{82.8} \\
\quad $\alpha{=}0.70$ & 87.7 & 80.2 & 92.0 & 70.9 & 65.8 & 80.4 & 77.9 &  \dunderline{1pt}{82.8} \\
\quad $\alpha{=}0.75$ & 87.9 &  \dunderline{1pt}{80.4} & 91.5 & 70.7 & 65.1 & 79.9 & 77.5 & 82.7 \\
\quad $\alpha{=}0.80$ & 88.0 & 80.3 & 91.1 & 70.5 & 64.3 & 79.4 & 77.1 & 82.5 \\
\quad $\alpha{=}0.85$ & 88.2 &  \dunderline{1pt}{80.4} & 90.5 & 70.2 & 63.5 & 78.9 & 76.7 & 82.5 \\
\quad $\alpha{=}0.90$ &  \dunderline{1pt}{88.3} &  \dunderline{1pt}{80.4} & 89.9 & 69.9 & 62.8 & 78.2 & 76.2 & 82.2 \\
\quad $\alpha{=}0.95$ &  \dunderline{1pt}{88.3} & 80.3 & 89.2 & 69.5 & 61.8 & 77.3 & 75.6 & 81.9 \\
\quad $\alpha{=}1.00$ & 88.2 & 80.1 & 88.5 & 69.1 & 61.0 & 76.3 & 75.0 & 81.6 \\\midrule
WiSE-FT, linear classifier & & & & & & & &\\
\quad $\alpha{=}0.00$ & 76.4 & 70.1 & 92.1 & 68.0 & 67.2 & 75.8 & 74.6 & 75.5 \\
\quad $\alpha{=}0.05$ & 77.5 & 71.1 & 92.3 & 68.3 & 67.4 & 76.3 & 75.1 & 76.3 \\
\quad $\alpha{=}0.10$ & 78.6 & 72.0 & 92.3 & 68.6 & 67.6 & 76.5 & 75.4 & 77.0 \\
\quad $\alpha{=}0.15$ & 79.5 & 73.0 &  \dunderline{1pt}{92.4} & 69.0 & 67.7 & 76.9 & 75.8 & 77.7 \\
\quad $\alpha{=}0.20$ & 80.3 & 73.5 &  \dunderline{1pt}{92.4} & 69.1 &  \dunderline{1pt}{67.8} & 77.3 & 76.0 & 78.2 \\
\quad $\alpha{=}0.25$ & 81.1 & 74.2 &  \dunderline{1pt}{92.4} &  \dunderline{1pt}{69.2} &  \dunderline{1pt}{67.8} & 77.3 & 76.2 & 78.7 \\
\quad $\alpha{=}0.30$ & 81.8 & 74.6 &  \dunderline{1pt}{92.4} &  \dunderline{1pt}{69.2} &  \dunderline{1pt}{67.8} & 77.5 & 76.3 & 79.0 \\
\quad $\alpha{=}0.35$ & 82.4 & 75.1 &  \dunderline{1pt}{92.4} & 69.1 &  \dunderline{1pt}{67.8} & 77.6 &  \dunderline{1pt}{76.4} & 79.4 \\
\quad $\alpha{=}0.40$ & 82.9 & 75.5 & 92.2 & 69.0 & 67.7 &  \dunderline{1pt}{77.8} &  \dunderline{1pt}{76.4} & 79.7 \\
\quad $\alpha{=}0.45$ & 83.4 & 75.8 & 92.2 & 68.9 & 67.4 & 77.7 &  \dunderline{1pt}{76.4} & 79.9 \\
\quad $\alpha{=}0.50$ & 83.7 & 76.1 & 91.9 & 68.8 & 67.3 & 77.6 & 76.3 & 80.0 \\
\quad $\alpha{=}0.55$ & 84.1 & 76.0 & 91.8 & 68.6 & 67.1 & 77.4 & 76.2 &  \dunderline{1pt}{80.2} \\
\quad $\alpha{=}0.60$ & 84.5 & 76.3 & 91.6 & 68.5 & 66.8 & 77.0 & 76.0 &  \dunderline{1pt}{80.2} \\
\quad $\alpha{=}0.65$ & 84.7 & 76.4 & 91.3 & 68.2 & 66.4 & 76.9 & 75.8 &  \dunderline{1pt}{80.2} \\
\quad $\alpha{=}0.70$ & 84.9 & 76.4 & 91.0 & 68.0 & 66.2 & 76.5 & 75.6 &  \dunderline{1pt}{80.2} \\
\quad $\alpha{=}0.75$ & 85.1 & 76.4 & 90.6 & 67.6 & 65.9 & 76.2 & 75.3 &  \dunderline{1pt}{80.2} \\
\quad $\alpha{=}0.80$ &  \dunderline{1pt}{85.2} & 76.4 & 90.2 & 67.3 & 65.5 & 75.9 & 75.1 &  \dunderline{1pt}{80.2} \\
\quad $\alpha{=}0.85$ &  \dunderline{1pt}{85.2} &  \dunderline{1pt}{76.5} & 89.7 & 66.8 & 65.0 & 75.3 & 74.7 & 80.0 \\
\quad $\alpha{=}0.90$ &  \dunderline{1pt}{85.2} & 76.3 & 89.2 & 66.3 & 64.4 & 74.9 & 74.2 & 79.7 \\
\quad $\alpha{=}0.95$ &  \dunderline{1pt}{85.2} & 76.0 & 88.6 & 65.7 & 63.8 & 74.4 & 73.7 & 79.5 \\
\quad $\alpha{=}1.00$ & 85.1 & 75.7 & 87.8 & 65.1 & 63.2 & 73.7 & 73.1 & 79.1 \\
\bottomrule

\end{tabular}
\caption{\label{tab:align}
 WiSE-FT accuracy on the reference and shifted distributions for various values of the mixing coefficient $\alpha$. Results shown for ALIGN, fine-tuned end-to-end (top) and with a linear classifier (bottom). Note that $\alpha{=}0.0$ corresponds to the zero-shot model, while $\alpha=1.0$ corresponds to standard fine-tuning. \textit{Avg shifts} displays the mean performance among the five distribution shifts, while \textit{Avg reference, shifts} shows the average of ImageNet (reference) and Avg shifts.
}
\end{center}
\end{table*}

\subsection{WiSE-FT and additional models}
\label{sec:more-models}

Table \ref{tab:merged} summarizes the results for the main models we study, CLIP, ALIGN, BASIC and a ViT model pre-trained on JFT. Details are provided in the subsequent sections.

\subsubsection{ALIGN}
\label{sec:align}

In addition to CLIP, we show WiSE-FT to be effective for an additional zero-shot model, ALIGN \cite{jia2021scaling}.  Results are shown in Figure \ref{fig:align} and Table \ref{tab:align}. End-to-end fine-tuning is performed using AdamW, which we found to perform slightly better than SGD + momentum. The model is fine-tuned for 40,000 steps with a batch size of 512, a maximum learning rate of $5\times 10^{-6}$, and weight decay of 0.1. The learning rate schedule consisted of 500 steps of linear warmup followed by cosine decay. The linear classifier is trained using L-BFGS and no label smoothing. All models are evaluated on $360\times 360$ pixel crops obtained by taking the central 87.5\% square region of the test set images. For end-to-end fine-tuning, we take $299\times 299$ pixel Inception-style random crops from the original ImageNet images during training; for linear classifier training, we use the same preprocessing as at evaluation time. The weights of the zero-shot model are calibrated using temperature scaling on the ImageNet training set before performing WiSE-FT.

\FloatBarrier
\clearpage

\begin{figure*}[h]
    \centering
    \includegraphics[width=.9\textwidth]{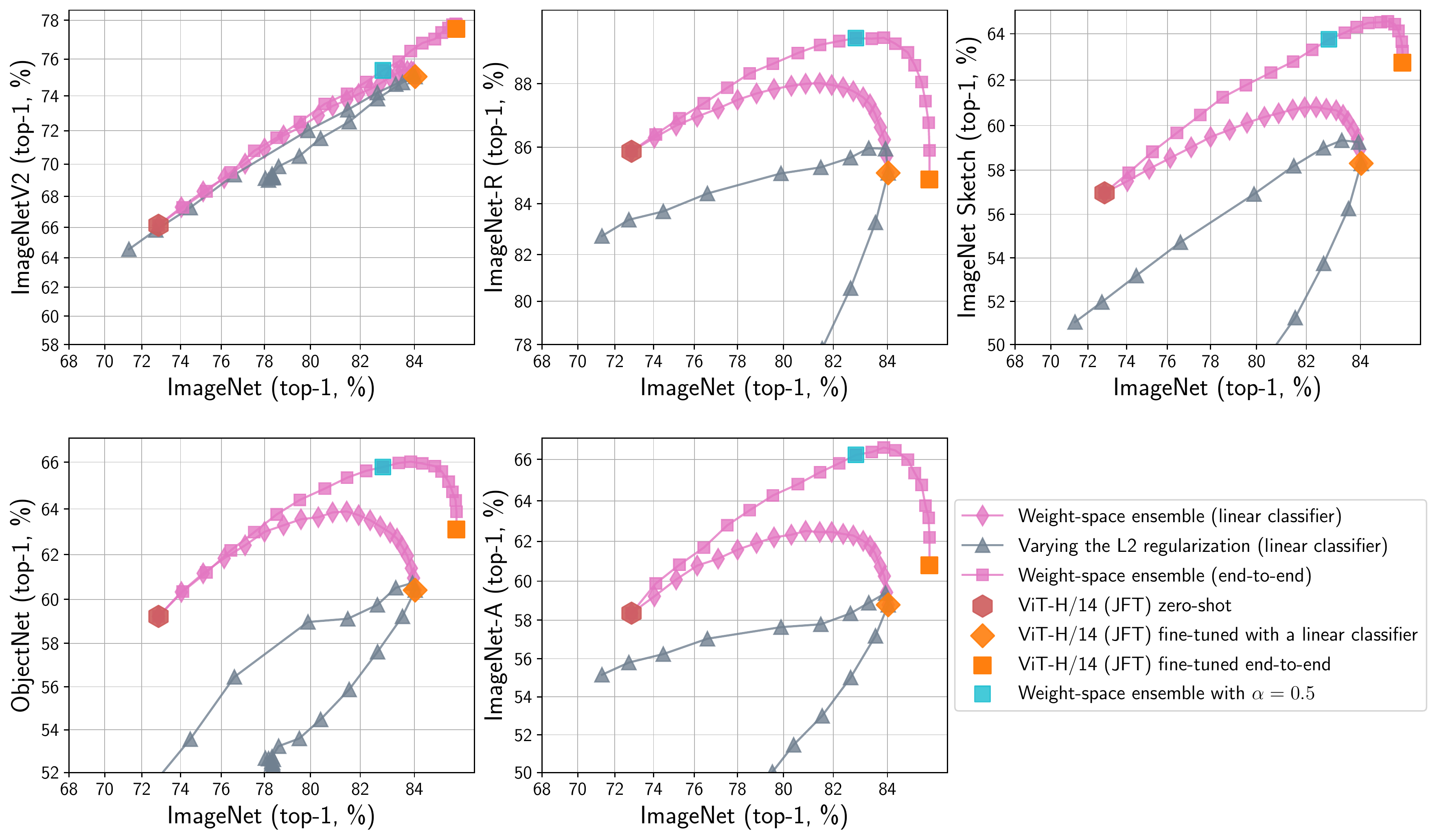}
    \caption{WiSE-FT applied to \texttt{ViT-H/14} \cite{dosovitskiy2021an} pre-trained on JFT. We also show the effect of varying the L2 regularization strength for linear classifier fine-tuning.}
    \label{fig:jft}
\end{figure*}

\subsubsection{JFT pre-training}
\label{sec:JFT}

\begin{table*}
\setlength\tabcolsep{5.1pt}
\small
\begin{center}
\begin{tabular}{lc|ccccc|cc}
\toprule
{} &            &             \multicolumn{5}{c|}{Distribution shifts}             & Avg &     Avg\\
{} &           IN (ref.) &             IN-V2 &              IN-R &                 IN-Sketch &                 ObjectNet &              IN-A & shifts &     ref., shifts\\
\midrule
WiSE-FT, edn-to-end & & & & & & & &\\
\quad $\alpha{=}0.00$ & 72.9 & 66.1 & 85.9 & 57.0 & 59.2 & 58.4 & 65.3 & 69.1 \\
\quad $\alpha{=}0.05$ & 74.1 & 67.3 & 86.4 & 57.9 & 60.4 & 59.9 & 66.4 & 70.2 \\
\quad $\alpha{=}0.10$ & 75.3 & 68.3 & 86.9 & 58.8 & 61.2 & 60.8 & 67.2 & 71.2 \\
\quad $\alpha{=}0.15$ & 76.5 & 69.5 & 87.4 & 59.7 & 62.2 & 61.7 & 68.1 & 72.3 \\
\quad $\alpha{=}0.20$ & 77.5 & 70.8 & 87.9 & 60.5 & 63.0 & 62.8 & 69.0 & 73.2 \\
\quad $\alpha{=}0.25$ & 78.5 & 71.6 & 88.3 & 61.2 & 63.8 & 63.5 & 69.7 & 74.1 \\
\quad $\alpha{=}0.30$ & 79.6 & 72.5 & 88.6 & 61.8 & 64.4 & 64.3 & 70.3 & 74.9 \\
\quad $\alpha{=}0.35$ & 80.6 & 73.5 & 88.9 & 62.3 & 64.9 & 64.8 & 70.9 & 75.8 \\
\quad $\alpha{=}0.40$ & 81.5 & 74.1 & 89.1 & 62.8 & 65.3 & 65.4 & 71.3 & 76.4 \\
\quad $\alpha{=}0.45$ & 82.2 & 74.8 & 89.2 & 63.3 & 65.6 & 65.8 & 71.7 & 77.0 \\
\quad $\alpha{=}0.50$ & 82.9 & 75.4 &  \dunderline{1pt}{89.3} & 63.8 & 65.8 & 66.2 & 72.1 & 77.5 \\
\quad $\alpha{=}0.55$ & 83.4 & 75.9 &  \dunderline{1pt}{89.3} & 64.0 &  \dunderline{1pt}{66.0} & 66.3 & 72.3 & 77.8 \\
\quad $\alpha{=}0.60$ & 83.9 & 76.4 &  \dunderline{1pt}{89.3} & 64.3 &  \dunderline{1pt}{66.0} &  \dunderline{1pt}{66.6} &  \dunderline{1pt}{72.5} & 78.2 \\
\quad $\alpha{=}0.65$ & 84.3 & 76.8 & 89.1 &  \dunderline{1pt}{64.5} & 65.9 & 66.4 &  \dunderline{1pt}{72.5} & 78.4 \\
\quad $\alpha{=}0.70$ & 84.7 & 77.1 & 88.9 &  \dunderline{1pt}{64.5} & 65.8 & 66.0 &  \dunderline{1pt}{72.5} &  \dunderline{1pt}{78.6} \\
\quad $\alpha{=}0.75$ & 84.9 & 77.4 & 88.5 &  \dunderline{1pt}{64.5} & 65.6 & 65.3 & 72.3 &  \dunderline{1pt}{78.6} \\
\quad $\alpha{=}0.80$ & 85.2 & 77.6 & 88.1 & 64.4 & 65.2 & 64.8 & 72.0 &  \dunderline{1pt}{78.6} \\
\quad $\alpha{=}0.85$ & 85.3 &  \dunderline{1pt}{77.8} & 87.5 & 64.1 & 64.7 & 63.8 & 71.6 & 78.4 \\
\quad $\alpha{=}0.90$ &  \dunderline{1pt}{85.4} &  \dunderline{1pt}{77.8} & 86.8 & 63.7 & 64.4 & 63.2 & 71.2 & 78.3 \\
\quad $\alpha{=}0.95$ &  \dunderline{1pt}{85.4} &  \dunderline{1pt}{77.8} & 85.9 & 63.3 & 63.9 & 62.2 & 70.6 & 78.0 \\
\quad $\alpha{=}1.00$ &  \dunderline{1pt}{85.4} & 77.6 & 84.9 & 62.8 & 63.1 & 60.8 & 69.8 & 77.6 \\\midrule
WiSE-FT, linear classifier & & & & & & & &\\
\quad $\alpha{=}0.00$ & 72.9 & 66.1 & 85.9 & 57.0 & 59.2 & 58.4 & 65.3 & 69.1 \\
\quad $\alpha{=}0.05$ & 74.0 & 67.3 & 86.3 & 57.5 & 60.3 & 59.2 & 66.1 & 70.0 \\
\quad $\alpha{=}0.10$ & 75.1 & 68.3 & 86.7 & 58.1 & 61.2 & 60.1 & 66.9 & 71.0 \\
\quad $\alpha{=}0.15$ & 76.1 & 69.1 & 87.0 & 58.5 & 61.8 & 60.8 & 67.4 & 71.8 \\
\quad $\alpha{=}0.20$ & 77.1 & 70.0 & 87.3 & 59.0 & 62.4 & 61.1 & 68.0 & 72.5 \\
\quad $\alpha{=}0.25$ & 78.0 & 71.0 & 87.5 & 59.5 & 63.0 & 61.6 & 68.5 & 73.2 \\
\quad $\alpha{=}0.30$ & 78.8 & 71.7 & 87.7 & 59.8 & 63.3 & 61.9 & 68.9 & 73.8 \\
\quad $\alpha{=}0.35$ & 79.6 & 72.2 & 87.8 & 60.1 & 63.6 & 62.2 & 69.2 & 74.4 \\
\quad $\alpha{=}0.40$ & 80.3 & 72.9 & 87.9 & 60.4 & 63.6 & 62.3 & 69.4 & 74.8 \\
\quad $\alpha{=}0.45$ & 80.9 & 73.4 &  \dunderline{1pt}{88.0} & 60.5 & 63.8 &  \dunderline{1pt}{62.5} & 69.6 & 75.2 \\
\quad $\alpha{=}0.50$ & 81.5 & 73.8 &  \dunderline{1pt}{88.0} & 60.7 &  \dunderline{1pt}{63.9} &  \dunderline{1pt}{62.5} &  \dunderline{1pt}{69.8} & 75.7 \\
\quad $\alpha{=}0.55$ & 81.9 & 74.1 &  \dunderline{1pt}{88.0} &  \dunderline{1pt}{60.8} & 63.7 &  \dunderline{1pt}{62.5} &  \dunderline{1pt}{69.8} & 75.8 \\
\quad $\alpha{=}0.60$ & 82.4 & 74.4 & 87.9 &  \dunderline{1pt}{60.8} & 63.5 & 62.4 &  \dunderline{1pt}{69.8} & 76.1 \\
\quad $\alpha{=}0.65$ & 82.8 & 74.7 & 87.8 & 60.7 & 63.2 & 62.3 & 69.7 & 76.2 \\
\quad $\alpha{=}0.70$ & 83.1 & 75.0 & 87.6 & 60.7 & 63.0 & 62.0 & 69.7 & 76.4 \\
\quad $\alpha{=}0.75$ & 83.4 & 75.2 & 87.4 & 60.5 & 62.7 & 61.8 & 69.5 &  \dunderline{1pt}{76.5} \\
\quad $\alpha{=}0.80$ & 83.6 &  \dunderline{1pt}{75.4} & 87.1 & 60.2 & 62.4 & 61.4 & 69.3 & 76.4 \\
\quad $\alpha{=}0.85$ & 83.7 &  \dunderline{1pt}{75.4} & 86.7 & 59.8 & 61.9 & 60.7 & 68.9 & 76.3 \\
\quad $\alpha{=}0.90$ & 83.9 &  \dunderline{1pt}{75.4} & 86.3 & 59.4 & 61.4 & 60.3 & 68.6 & 76.2 \\
\quad $\alpha{=}0.95$ &  \dunderline{1pt}{84.0} & 75.3 & 85.7 & 58.9 & 61.0 & 59.4 & 68.1 & 76.0 \\
\quad $\alpha{=}1.00$ &  \dunderline{1pt}{84.0} & 75.1 & 85.1 & 58.3 & 60.4 & 58.8 & 67.5 & 75.8 \\
\bottomrule
\end{tabular}
\caption{\label{tab:jft}
WiSE-FT accuracy on the reference and shifted distributions for various values of the mixing coefficient $\alpha$. Results shown for \texttt{ViT-H/14} pre-trained on JFT-300M, fine-tuned end-to-end (top) and with a linear classifier (bottom). Note that $\alpha{=}0.0$ corresponds to the zero-shot model, while $\alpha=1.0$ corresponds to standard fine-tuning. \textit{Avg shifts} displays the mean performance among the five distribution shifts, while \textit{Avg reference, shifts} shows the average of ImageNet (reference) and Avg shifts.
}
\end{center}
\end{table*}

We also investigate whether WiSE-FT can provide gains for models trained using a standard image classification objective on the JFT-300M dataset \cite{sun2017revisiting}. Results are shown in Figure \ref{fig:jft} and Table \ref{tab:jft}. For 973/1000 ImageNet classes, we were able to manually identify a corresponding class from the 18K classes in JFT. We use this mapping between ImageNet and JFT classes to obtain zero-shot ImageNet weights from the final layer weights of the pre-trained \texttt{ViT-H/14} model from Dosovitskiy et al.~\cite{dosovitskiy2021an}. We also train a linear classifier on the fixed penultimate layer of the same \texttt{ViT-H/14} model using L-BFGS without label smoothing with softmax cross-entropy loss, and fine-tune end-to-end using AdamW with maximum learning rate $5 \cdot 10^{-6}$ and weight decay 0.1 for 20k iterations at batch size 512 with sigmoid cross-entropy loss. As for CLIP models, our learning rate schedule consists of 500 steps of linear warmup followed by cosine decay. All ViT-H/14 models are trained and evaluated on $224\times 224$ pixel images. For fair evaluation, we prevent fine-tuned solutions from predicting the 27 classes with no plausible corresponding JFT class at all points on the WiSE-FT curve but still include these points in the denominator when computing accuracy. 

\FloatBarrier

\subsubsection{BASIC}
\label{sec:basic}

\begin{figure}[h]
    \centering
    \includegraphics[width=\textwidth]{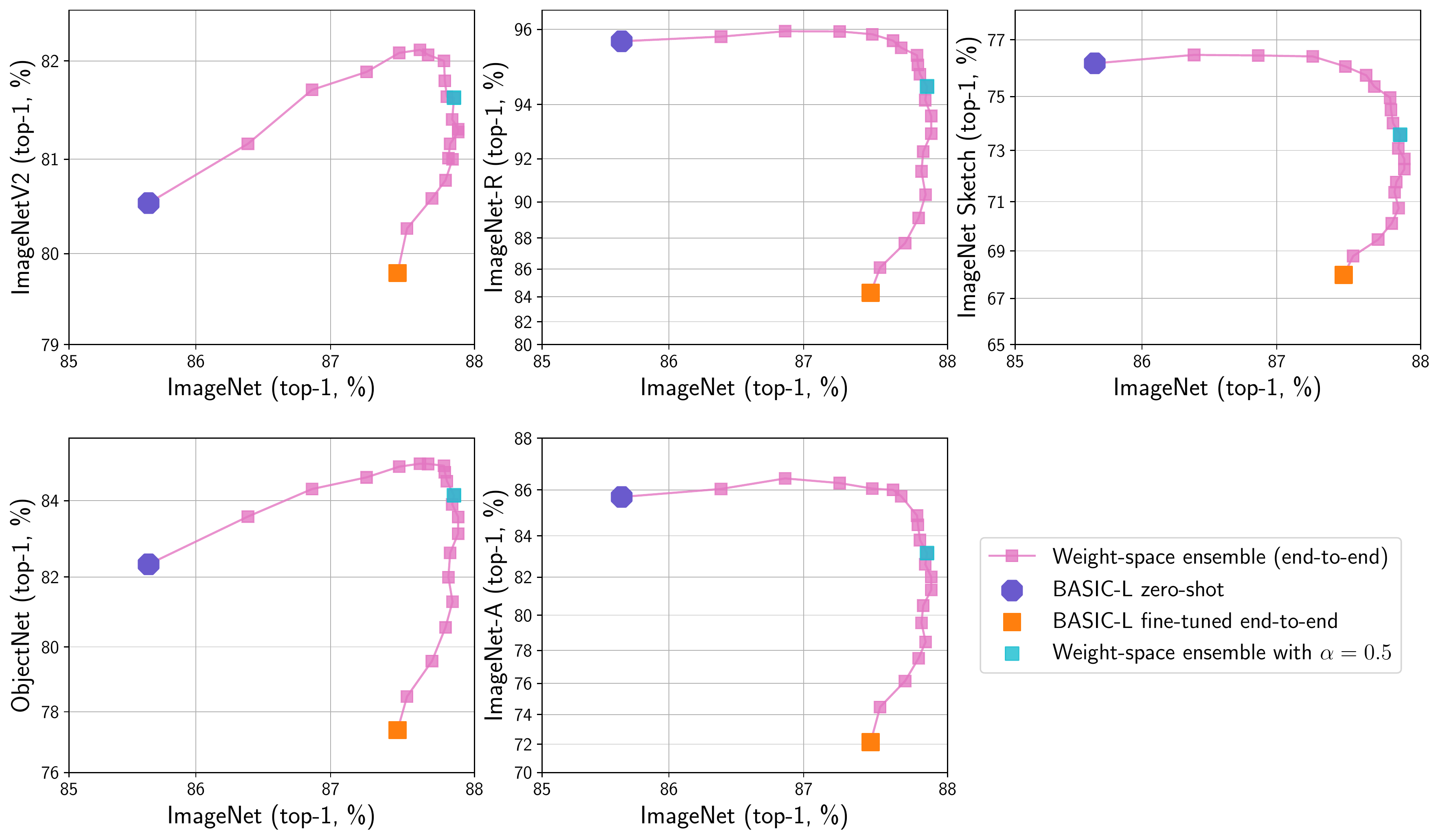}
    \caption{WiSE-FT improves accuracy relative to the fine-tuned model on ImageNet and five derived distribution shifts for BASIC-L \cite{pham2021scaling} using ImageNet class names to construct the zero-shot classifier.}
    \label{fig:basicbreakdown}
\end{figure}

We apply WiSE-FT to BASIC \cite{pham2021scaling}, fine-tuning both the image and text encoder with a contrastive loss on half of the ImageNet training data, as in \citet{pham2021scaling}. Results are shown in Figure \ref{fig:basicbreakdown} and Tables \ref{tab:basic_m} and \ref{tab:basic_l}.

\begin{table*}
\setlength\tabcolsep{5.1pt}
\small
\begin{center}
\begin{tabular}{lc|ccccc|cc}
\toprule
{} &            &             \multicolumn{5}{c|}{Distribution shifts}             & Avg &     Avg\\
{} &           IN (ref.) &             IN-V2 &              IN-R &                 IN-Sketch &                 ObjectNet &              IN-A & shifts &     ref., shifts\\
\midrule
$\alpha{=}0.00$ & 81.4 & 74.1 & 90.6 & 67.4 & 73.5 & 66.7 & 74.5 &    
                78.0 \\         
$\alpha{=}0.05$ & 82.2 & 75.0 & 90.8 & 67.9 & 74.6 & 67.8 & 75.2 &    
                78.7 \\         
$\alpha{=}0.10$ & 82.8 & 75.9 & 90.9 & 68.2 & 75.4 & 68.5 & 75.8 &    
                79.3 \\         
$\alpha{=}0.15$ & 83.3 & 76.4 & 91.0 & 68.4 & 76.2 & 69.3 & 76.3 &    
                79.8 \\         
$\alpha{=}0.20$ & 83.8 & 76.8 & 91.0 & 68.6 & 76.9 & 70.0 & 76.7 &    
                80.2 \\         
$\alpha{=}0.25$ & 84.1 & 77.1 &  \dunderline{1pt}{91.1} & 68.7 & 77.4 & 70.5 & 77.0 &    
                80.5 \\         
$\alpha{=}0.30$ & 84.5 & 77.4 & 91.0 &  \dunderline{1pt}{68.8} & 77.7 & 70.8 & 77.1 &    
                80.8 \\         
$\alpha{=}0.35$ & 84.9 & 77.9 & 90.8 &  \dunderline{1pt}{68.8} & 77.8 & 71.3 & 77.3 &    
                81.1 \\         
$\alpha{=}0.40$ & 85.2 & 78.1 & 90.7 & 68.7 & 77.9 & 71.3 & 77.3 &    
                81.2 \\         
$\alpha{=}0.45$ & 85.4 & 78.3 & 90.5 & 68.7 &  \dunderline{1pt}{78.0} &  \dunderline{1pt}{71.4} &  \dunderline{1pt}{77.4} &  \dunderline{1pt}{81.4} \\         
$\alpha{=}0.50$ & 85.6 & 78.5 & 90.2 & 68.6 &  \dunderline{1pt}{78.0} & 71.1 & 77.3 &  \dunderline{1pt}{81.4} \\         
$\alpha{=}0.55$ & 85.8 & 78.5 & 89.9 & 68.4 &  \dunderline{1pt}{78.0} & 70.6 & 77.1 &  \dunderline{1pt}{81.4} \\         
$\alpha{=}0.60$ & 85.9 & 78.4 & 89.5 & 68.1 &  \dunderline{1pt}{78.0} & 70.5 & 76.9 &  \dunderline{1pt}{81.4} \\         
$\alpha{=}0.65$ & 86.0 & 78.5 & 89.1 & 67.7 & 77.8 & 70.3 & 76.7 &    
                81.3 \\         
$\alpha{=}0.70$ & 86.1 & 78.5 & 88.8 & 67.3 & 77.6 & 69.7 & 76.4 &    
                81.2 \\         
$\alpha{=}0.75$ &  \dunderline{1pt}{86.2} &  \dunderline{1pt}{78.6} & 88.4 & 67.0 & 77.3 & 69.2 & 76.1 &    
                81.2 \\         
$\alpha{=}0.80$ &  \dunderline{1pt}{86.2} & 78.5 & 87.8 & 66.6 & 77.1 & 68.3 & 75.7 &    
                81.0 \\         
$\alpha{=}0.85$ &  \dunderline{1pt}{86.2} & 78.5 & 87.2 & 66.0 & 76.7 & 67.5 & 75.2 &    
                80.7 \\         
$\alpha{=}0.90$ &  \dunderline{1pt}{86.2} & 78.4 & 86.5 & 65.5 & 76.2 & 66.4 & 74.6 &    
                80.4 \\         
$\alpha{=}0.95$ &  \dunderline{1pt}{86.2} & 78.2 & 85.7 & 65.0 & 75.8 & 65.3 & 74.0 &    
                80.1 \\         
$\alpha{=}1.00$ &  \dunderline{1pt}{86.2} & 77.8 & 84.9 & 64.3 & 75.3 & 63.7 & 73.2 &    
                79.7 \\                  
\bottomrule
\end{tabular}
\caption{\label{tab:basic_m}
WiSE-FT accuracy on the reference and shifted distributions for various values of the mixing coefficient $\alpha$. Results shown for BASIC-M using ImageNet class names. Note that $\alpha{=}0.0$ corresponds to the zero-shot model, while $\alpha=1.0$ corresponds to standard fine-tuning. \textit{Avg shifts} displays the mean performance among the five distribution shifts, while \textit{Avg reference, shifts} shows the average of ImageNet (reference) and Avg shifts.
}
\end{center}
\end{table*}

\begin{table*}
\setlength\tabcolsep{5.1pt}
\small
\begin{center}
\begin{tabular}{lc|ccccc|cc}
\toprule
{} &            &             \multicolumn{5}{c|}{Distribution shifts}             & Avg &     Avg\\
{} &           IN (ref.) &             IN-V2 &              IN-R &                 IN-Sketch &                 ObjectNet &              IN-A & shifts &     ref., shifts\\
\midrule
$\alpha{=}0.00$ & 85.6 & 80.5 & 95.7 & 76.2 & 82.3 & 85.7 & 84.1 & 84.8 \\
$\alpha{=}0.05$ & 86.4 & 81.2 & 95.8 &  \dunderline{1pt}{76.5} & 83.6 & 86.0 & 84.6 & 85.5 \\
$\alpha{=}0.10$ & 86.9 & 81.7 &  \dunderline{1pt}{96.0} &  \dunderline{1pt}{76.5} & 84.3 &  \dunderline{1pt}{86.5} &  \dunderline{1pt}{85.0} & 86.0 \\
$\alpha{=}0.15$ & 87.3 & 81.9 &  \dunderline{1pt}{96.0} & 76.4 & 84.6 & 86.3 &  \dunderline{1pt}{85.0} &  \dunderline{1pt}{86.2} \\
$\alpha{=}0.20$ & 87.5 &  \dunderline{1pt}{82.1} & 95.9 & 76.1 & 84.8 & 86.1 &  \dunderline{1pt}{85.0} &  \dunderline{1pt}{86.2} \\
$\alpha{=}0.25$ & 87.6 &  \dunderline{1pt}{82.1} & 95.7 & 75.8 &  \dunderline{1pt}{84.9} & 86.0 & 84.9 &  \dunderline{1pt}{86.2} \\
$\alpha{=}0.30$ & 87.7 &  \dunderline{1pt}{82.1} & 95.6 & 75.4 &  \dunderline{1pt}{84.9} & 85.7 & 84.7 &  \dunderline{1pt}{86.2} \\
$\alpha{=}0.35$ & 87.8 & 82.0 & 95.4 & 75.0 &  \dunderline{1pt}{84.9} & 84.9 & 84.4 & 86.1 \\
$\alpha{=}0.40$ & 87.8 & 81.8 & 95.1 & 74.5 & 84.7 & 84.5 & 84.1 & 85.9 \\
$\alpha{=}0.45$ & 87.8 & 81.6 & 94.9 & 74.0 & 84.5 & 83.8 & 83.8 & 85.8 \\
$\alpha{=}0.50$ &  \dunderline{1pt}{87.9} & 81.6 & 94.5 & 73.6 & 84.1 & 83.2 & 83.4 & 85.7 \\
$\alpha{=}0.55$ & 87.8 & 81.4 & 94.1 & 73.1 & 83.9 & 82.6 & 83.0 & 85.4 \\
$\alpha{=}0.60$ &  \dunderline{1pt}{87.9} & 81.3 & 93.6 & 72.7 & 83.6 & 82.0 & 82.6 & 85.2 \\
$\alpha{=}0.65$ &  \dunderline{1pt}{87.9} & 81.3 & 93.0 & 72.3 & 83.2 & 81.3 & 82.2 & 85.1 \\
$\alpha{=}0.70$ & 87.8 & 81.2 & 92.3 & 71.8 & 82.7 & 80.5 & 81.7 & 84.8 \\
$\alpha{=}0.75$ & 87.8 & 81.0 & 91.5 & 71.4 & 82.0 & 79.6 & 81.1 & 84.4 \\
$\alpha{=}0.80$ &  \dunderline{1pt}{87.9} & 81.0 & 90.4 & 70.7 & 81.3 & 78.5 & 80.4 & 84.2 \\
$\alpha{=}0.85$ & 87.8 & 80.8 & 89.1 & 70.1 & 80.6 & 77.5 & 79.6 & 83.7 \\
$\alpha{=}0.90$ & 87.7 & 80.6 & 87.7 & 69.5 & 79.6 & 76.1 & 78.7 & 83.2 \\
$\alpha{=}0.95$ & 87.5 & 80.3 & 86.1 & 68.8 & 78.5 & 74.5 & 77.6 & 82.5 \\
$\alpha{=}1.00$ & 87.5 & 79.8 & 84.3 & 68.0 & 77.4 & 72.1 & 76.3 & 81.9 \\
\bottomrule
\end{tabular}
\caption{\label{tab:basic_l}
WiSE-FT accuracy on the reference and shifted distributions for various values of the mixing coefficient $\alpha$. Results shown for BASIC-L using ImageNet class names. Note that $\alpha{=}0.0$ corresponds to the zero-shot model, while $\alpha=1.0$ corresponds to standard fine-tuning. \textit{Avg shifts} displays the mean performance among the five distribution shifts, while \textit{Avg reference, shifts} shows the average of ImageNet (reference) and Avg shifts.
}
\end{center}
\end{table*}

\FloatBarrier
\clearpage

\begin{figure*}[h]
    \centering
    \includegraphics[width=\textwidth]{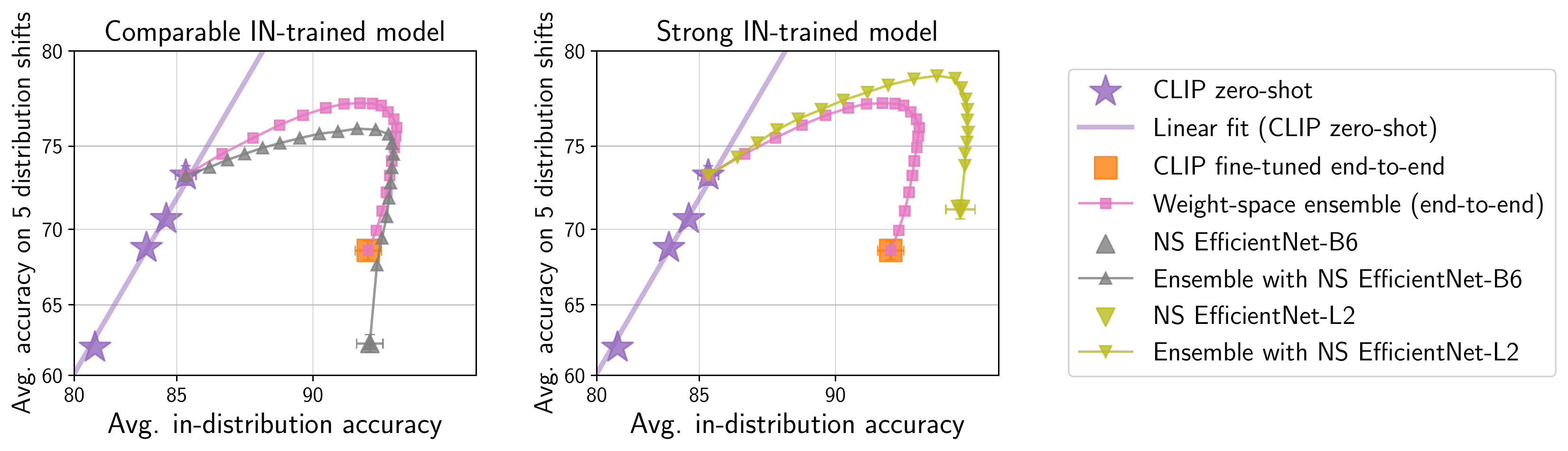}
    \caption{\textbf{Ensembling with a zero-shot model improves accuracy under distribution shift of an independently trained model}. \textbf{(Left)} Output-space ensembling with an independently trained model (NoisyStudent EfficientNet-B6) with comparable performance to the end-to-end fine-tuned model on the reference distribution. \textbf{(Right)} Output-space ensembling with an independently trained model with strong performance on the reference distribution (NoisyStudent EfficientNet-L2). Results averaged over the five distribution shifts as in Figure~\ref{fig:fig1}.}
    \label{fig:beyond}
\end{figure*}

\begin{table*}
\setlength\tabcolsep{3.2pt}
\small
\begin{center}
\begin{tabular}{lc|ccccc|cc}
\toprule
{} &            &             \multicolumn{5}{c|}{Distribution shifts}             & Avg &     Avg\\
{} &           IN (reference) &             IN-V2 &              IN-R &                 IN-Sketch &                 ObjectNet &              IN-A & shifts &     reference, shifts\\
\midrule
CLIP &            &  & & & &  &  & \\
\quad End-to-end fine-tuned                 &           86.2 &           76.8 &           79.8 &           57.9 &           63.3 &           65.4 &           68.6 &           77.4 \\  
\quad WSE ($\alpha{=}0.75$)     &           87.0 &           78.8 &           86.1 &           62.5 &           68.1 &           75.2 &           74.1 &           80.5 \\             
\quad WSE ($\alpha{=}0.5$)      &           86.8 &           79.5 &           89.4 &           64.7 &           71.1 &           79.9 &           76.9 &           81.8 \\             
\quad WSE ($\alpha{=}0.4$)      &           86.2 &           79.2 &           89.9 &  \textbf{65.0} &           71.9 &           80.7 &           77.3 &           81.8 \\             
\quad WSE (optimal $\alpha$)    &           87.1 &           79.5 &           90.3 &  \textbf{65.0} &           72.1 &           81.0 &           77.6 &           82.3 \\               
NS EfficientNet-B6  &            &  & & & &  &  & \\
\quad No ensemble      &           86.5 &           77.7 &           65.6 &           47.8 &           58.3 &           62.3 &           62.3 &           74.4 \\  
\quad OSE ($\alpha{=}0.75$)  &           87.0 &           78.8 &           86.4 &           56.7 &           66.5 &           75.9 &           72.9 &           80.0 \\                
\quad OSE ($\alpha{=}0.5$)   &           86.2 &           78.7 &           89.2 &           63.8 &           69.3 &           78.6 &           75.9 &           81.1 \\                
\quad OSE ($\alpha{=}0.4$)   &           84.3 &           77.2 &           89.5 &           63.8 &           69.7 &           79.0 &           75.8 &           80.0 \\                
\quad OSE (optimal $\alpha$) &           87.1 &           79.3 &           89.7 &           63.8 &           69.7 &           79.3 &           76.4 &           81.8 \\                 

NS EfficientNet-L2   &            &  & & & &  &  & \\
\quad No ensemble      &           88.3 &           80.8 &           74.6 &           47.6 &           69.8 &           84.7 &           71.5 &           79.9 \\  
\quad OSE ($\alpha{=}0.75$)  &  \textbf{88.6} &           81.6 &           88.0 &           53.4 &           72.2 &  \textbf{87.1} &           76.5 &           82.5 \\                
\quad OSE ($\alpha{=}0.5$)   &           87.4 &           80.6 &           90.2 &           63.4 &  \textbf{73.1} &           86.5 &           78.8 &           83.1 \\                
\quad OSE ($\alpha{=}0.4$)   &           85.2 &           78.5 &  \textbf{90.5} &           63.9 &           72.6 &           86.0 &           78.3 &           81.8 \\                
\quad OSE (optimal $\alpha$) &  \textbf{88.6} &  \textbf{81.7} &  \textbf{90.5} &           63.9 &  \textbf{73.1} &  \textbf{87.1} &  \textbf{79.3} &  \textbf{83.9} \\

\bottomrule
\end{tabular}
\caption{\label{tab:beyond}
Accuracy of various independently trained models ensembled with CLIP on ImageNet and derived distribution shifts. OSE denotes output-space ensembling.  \textit{Avg shifts} displays the mean performance among the five distribution shifts, while \textit{Avg reference, shifts} shows the average of ImageNet (reference) and Avg shifts.
}
\end{center}
\end{table*}

\subsection{Ensembling zero-shot CLIP with independently trained models}
\label{sec:beyond}

So far we have shown that a zero-shot model can be used to improve performance under distribution shift of the derived fine-tuned model. 
Here, we investigate whether this improvement is specific to fine-tuned models. On the contrary, we find that the performance under distribution shift of \textit{independently trained models} improves when ensembling with robust models. Note that in the general case where the models being ensembled have different architectures, we are unable to perform weight-space ensembling;
instead, we ensemble the outputs of each model. This increases the computational cost of inference, in contrast to the results shown in Section \ref{sec:results}.

Concretely, we ensemble zero-shot CLIP with two Noisy Student EfficientNet models \cite{xie2020self,tan2019efficientnet}: (i) EfficientNet-B6 (Figure \ref{fig:beyond}, left), with performance on the reference distribution comparable to the end-to-end fine-tuned CLIP model;
 and (ii) EfficientNet-L2 (Figure \ref{fig:beyond}, right), the strongest model available on PyTorch ImageNet Models \cite{rw2019timm}.
 In both cases, we observe substantial improvements from ensembling---13.6 pp and 6.9 pp in average accuracy under distribution shift without reducing performance on the reference dataset. Further results are shown in Table \ref{tab:beyond}.

\section{Experimental details}
\label{sec:appendix_hparam}

\subsection{CLIP zero-shot}\label{sec:morezs}

This section extends Section \ref{sec:expsetup} with more details on inference with the CLIP zero-shot model. First, in all settings we use the CLIP model \texttt{ViT-L/14@336px}, except when explicitly mentioned otherwise.
Second, CLIP learns a temperature parameter which is factored into the learned weight matrix $\textbf{W}_{\text{zero-shot}}$ described in Section~\ref{sec:expsetup}. Finally, to construct $\textbf{W}_{\text{zero-shot}}$ we ensemble the 80 prompts provided by CLIP at \url{https://github.com/openai/CLIP}. However, we manually engineer prompts for five datasets: WILDS-FMoW, WILDS-iWildCam, Stanford Cars, Describable Textures and Food-101, which are found in the code.

\subsection{End-to-end fine-tuning}
\label{sec:e2e-ft}

Two important experimental details for end-to-end fine-tuning are as follows:
\begin{itemize}
    \item We initialize the final classification layer with the zero-shot classifier used by CLIP. We scale the zero-shot classifier weights by the temperature parameter of the pre-trained CLIP model at initialization, and do not include a temperature parameter during fine-tuning.
    \item As the zero-shot classifier expects the outputs of the image-encoder $g$ to be normalized, we continue to normalize the outputs of $g$ during fine-tuning.
\end{itemize}

When fine-tuning end-to-end, unless otherwise mentioned, we use the AdamW optimizer \cite{loshchilov2018decoupled, paszke2019pytorch} and choose the largest batch size such that the model fits into 8 GPUs (512 for \texttt{ViT-B/16}). Unless otherwise mentioned, we use the default PyTorch AdamW hyperparameters $\beta_1=0.9$, $\beta_2=0.999$, $\epsilon=10^{-8}$, weight decay of 0.1 and a cosine-annealing learning rate schedule \cite{loshchilov2016sgdr} with 500 warm-up steps. Unless otherwise mentioned we use a learning rate of $3\times10^{-5}$, gradient clipping at global norm 1 and fine-tune for a total of 10 epochs. Additionally, unless otherwise mentioned we use the same data augmentations as \cite{radford2021learning}, randomly cropping a square from resized images with the largest dimension being 336 pixels for \texttt{ViT-L/14@336px} and 224 for the remaining models.

\subsection{Fine-tuning a linear classifier}\label{sec:moreclf}

This section extends the description of linear classifier training from Appendix~\ref{sec:baselines-appendix} with details on hyperparameters and additional analyses. In each of the four regularization strategies---no regularization, weight decay, L1 regularization, and label smoothing---we run 64 hyperparameter configurations. For each trial, mini-batch size is drawn uniformly from $\{64, 128, 256\}$ and learning rate is set to $10^{-\beta}$ with $\beta$ chosen uniformly at random from the range $[0,6]$.
Hyperparameters for each regularization strategy are as follows: (i) The weight decay coefficient is set to $10^{-\lambda}$ where $\lambda$ is chosen uniformly at random from $[0,4]$ for each trial; (ii) The L1 regularization coefficient is set to $10^{-\lambda}$ where $\lambda$ is chosen uniformly at random from $[4,8]$ for each trial; (iii) The label smoothing \cite{muller2019does} coefficient $\lambda$ is chosen uniformly at random from $[0, 0.25]$ for each trial. The linear classifier used for ensembling attains the best performance in-distribution. The hyperparameters from this trial are then used in the distillation and regularization experiments described in Appendix~\ref{sec:baselines-appendix}. In the low-data regime (Section~\ref{sec:low-data}), this process is repeated for each $k$ and dataset.

When training linear classifiers with $k$ images per class as in Section \ref{sec:low-data} the maximum number of epochs $T$ is scaled approximately inversely proportional to the amount of data removed (e.g., with half the data we train for twice as many epochs so the number of iterations is consistent). To choose the $T$ we use default PyTorch AdamW hyperparameters (learning rate 0.001, weight decay 0.01) and double the number of epochs until performance saturates. For each random hyperparameter run we choose the epochs uniformly from $\{1,...,T\}$.

\subsection{ObjectNet}\label{sec:objectnet}

\begin{figure*}
    \centering
    \includegraphics[width=0.85\textwidth]{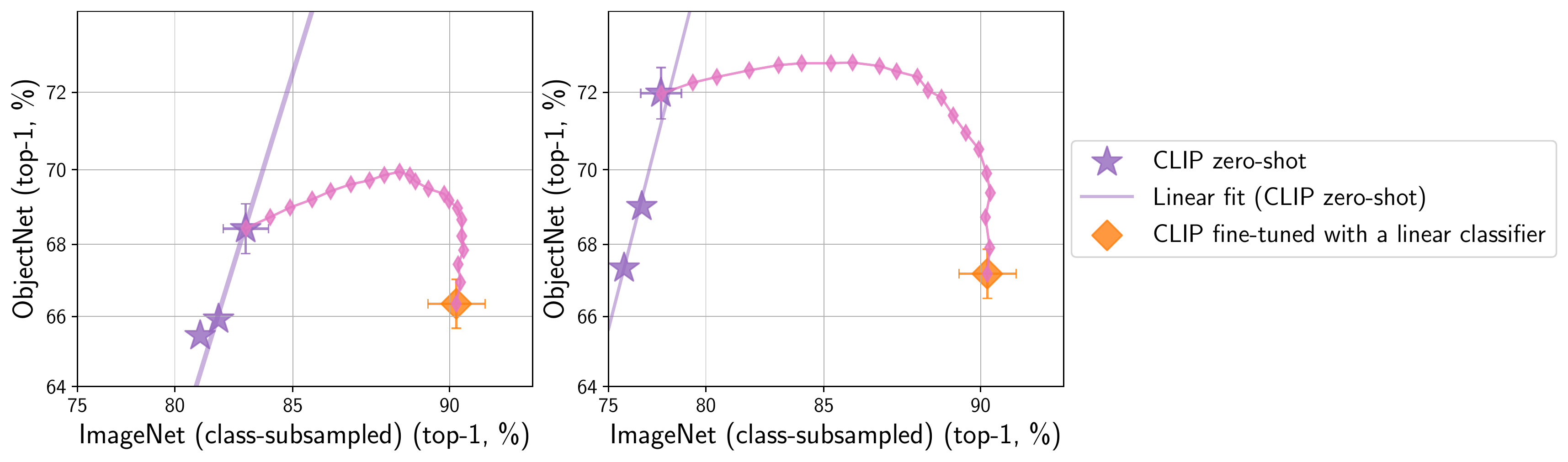}
    \caption{Effective robustness scatter plots for ObjectNet, with and without adapting to class shift. \textbf{Left:} Using ImageNet class names to construct the zero-shot classifier. \textbf{Right:} Using ObjectNet class names to construct the zero-shot classifier.}
    \label{fig:fig_objectnet}
\end{figure*}

The zero-shot models in Table \ref{tab:main} use the ImageNet class names instead of the ObjectNet class names. However, this \textit{adaptation to class shift} improves performance by 2.3\% \cite{radford2021learning}. Out of the five datasets used for the majority of the experiments in Section~\ref{sec:main}, ObjectNet is the only dataset for which this is possible. In Figure \ref{fig:fig_objectnet} we compare weight-space ensembles with and without adaptation to class shift.

\section{Diversity measures}
\label{sec:diversity_defs}

Let $\mathcal{S} = \{(x^{(i)}, y^{(i)}),  1 \le i \le N\}$ be a classification set with input data $x^{(i)}$ and labels $y^{(i)} \in \{1,...,C\}$, where $C$ is the number of classes. A classifier $f$ is a function that maps inputs $x$ to logits $f(x) \in \mathbb{R}^C$, yielding predictions $\hat{y} = \argmax_{1\le c\le C}f(x)_c$. We consider measures of diversity $\mathcal{M}(f,g, \mathcal{S})$ between two classifiers $f$ and $g$ and the dataset $\mathcal{S}$. For simplicity, $\hat{y}_f^{(i)}$ is used to denote the predictions from classifier $f$ given inputs $x^{(i)}$ (and similarly for $g$).

\paragraph{Prediction Diversity (PD).} One of the most intuitive ways to measure diversity between pairs of classifiers is to compute the fraction of samples where they disagree while one is correct \cite{ho1998random,skalak1996sources}. Formally, the prediction diversity $\textrm{PD}$ is defined as:
\begin{equation}
    \textrm{PD}(f, g, \mathcal{S}) = \frac{1}{N}\sum_{1 \le i \le N} \mathbbm{1}\left[d_f \vee d_g \right],
    \label{eq:pd}
\end{equation}
where
\begin{equation}
    d_{f} = \left(\hat{y}_f^{(i)} = y^{(i)} \wedge \hat{y}_g^{(i)} \ne y^{(i)}\right).
\end{equation}
\begin{equation}
    d_{g} = \left(\hat{y}_f^{(i)} \ne y^{(i)} \wedge \hat{y}_g^{(i)} = y^{(i)}\right).
\end{equation}

\paragraph{Cohen's Kappa Complement (CC).} Cohen's kappa coefficient is a measure of agreement between two annotators \cite{mchugh2012interrater}. Here, we use it's complement as a diversity measure between two classifiers:
\begin{equation}
    \textrm{CC}(f, g, \mathcal{S}) = 1 - \frac{p_o - p_e}{1 - p_e}  = \frac{1-p_o}{1-p_e},
    \label{eq:cc}
\end{equation}
where $p_e$ is the expected agreement between the classifiers and $p_o$ is the empirical probability of agreement. Formally, if $n_{f,k}$ is the number of samples where classifier $f$ predicted label $k$ (i.e. $n_{f,k} = \sum_{1\le i \le N} \mathbbm{1}[\hat{y}^{i}_f = k]$), then:
\begin{equation}
    p_e = \frac{1}{N^2}\sum_{1\le c \le C}n_{f,c}n_{g,c}, \quad p_o = \frac{1}{N}\sum_{1\le i \le N}\mathbbm{1}[\hat{y}^{i}_f = \hat{y}^{i}_g]
\end{equation}

\paragraph{KL Divergence (KL).} The Kullback-Leibler divergence measures how different a probability distribution is from another. Let $p_f^{(i)} = \textrm{softmax}\left(f(x^{(i)})\right)$ for a classifier $f$, and let $p_{f,c}^{(i)}$ be the probability assigned to class $c$. We consider the average KL-divergence over all samples as a diversity measure:
\begin{equation}
    \textrm{KL}(f, g, \mathcal{S}) = \frac{1}{N}\sum_{1 \le i \le N}\sum_{1\le c \le C}p_{f,c}^{(i)}\log\left(\frac{p_{f,c}^{(i)}}{p_{g,c}^{(i)}}\right).
    \label{eq:kl}
\end{equation}

\paragraph{Centered Kernel Alignment Complement (CKAC).} CKA is a similarity measure that compares two different sets of high-dimensional representations \cite{kornblith2019similarity}. It is commonly used for comparing representations of two neural networks, or determining correspondences between two hidden layers of the same network. CKA measures the agreement between two matrices containing the pair-wise similarities of all samples in a dataset, where each matrix is constructed according to the representations of a model. More formally, let $S \in \mathbb{R}^{N\times d}$ denote the $d$-dimensional features for all samples in a dataset $\mathcal{S}$, pre-processed to center the columns. For two models $f$ and $g$ yielding similarity matrices $S_f$ and $S_g$, CKA is defined as:

\begin{equation}
    \mathrm{CKA}(f, g, \mathcal{S}) = \frac{||S_g^\top S_f||^2_F}{||S_f^\top S_f||_F ||S_g^\top S_g||_F},
\end{equation}
where $||S||_F$ denotes the Frobenius norm of the matrix $S$. Larger CKA values indicate larger similarities between the representations of the two models, and thus, smaller diversity. We define the diversity measure CKAC as:
\begin{equation}
    \mathrm{CKAC} = 1 - \mathrm{CKA}.
    \label{eq:ckac}
\end{equation}
Note that CKAC is computationally expensive to compute for large datasets. For this reason, in our experiments with distributions larger than 10,000 samples, we randomly sample 10,000 to compute this measure.

\begin{figure*}
    \centering
    \includegraphics[width=\textwidth]{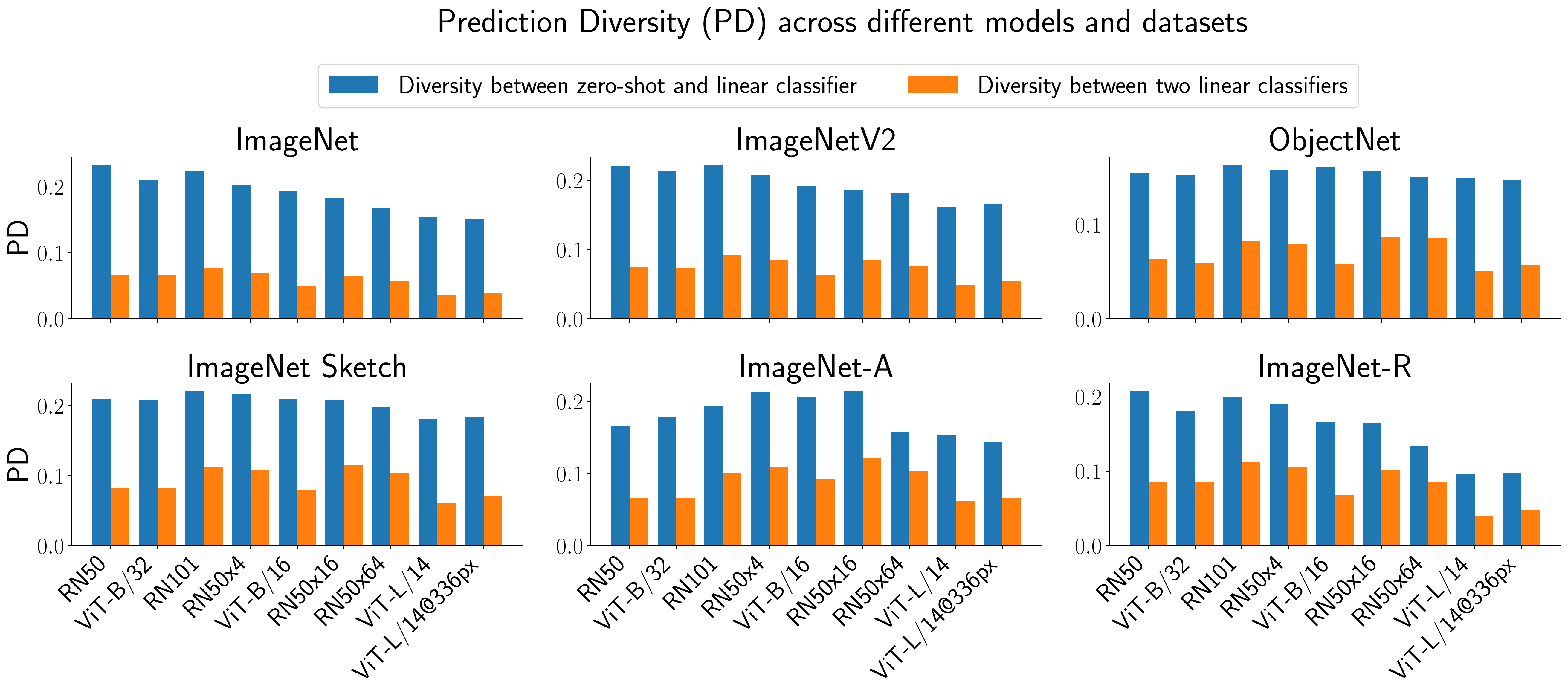}
    \caption{\textbf{Prediction Diversity (PD)} for multiple datasets and CLIP models (Equation \ref {eq:pd}).}
    \label{fig:diversity_breakdown_start}
\end{figure*}

\paragraph{Diversity across different architectures} We extend Figure \ref{fig:diversity_and_confidence} to show results for all combinations of diversity measures, datasets, and CLIP models. Similarly to before, the baselines compares models with the same encoder, with two linear classifiers trained on different subsets of ImageNet with half of the data. Results are shown in Figures \ref{fig:diversity_breakdown_start}-\ref{fig:diversity_breakdown_end}.

\FloatBarrier

\begin{figure*}
    \centering
    \includegraphics[width=\textwidth]{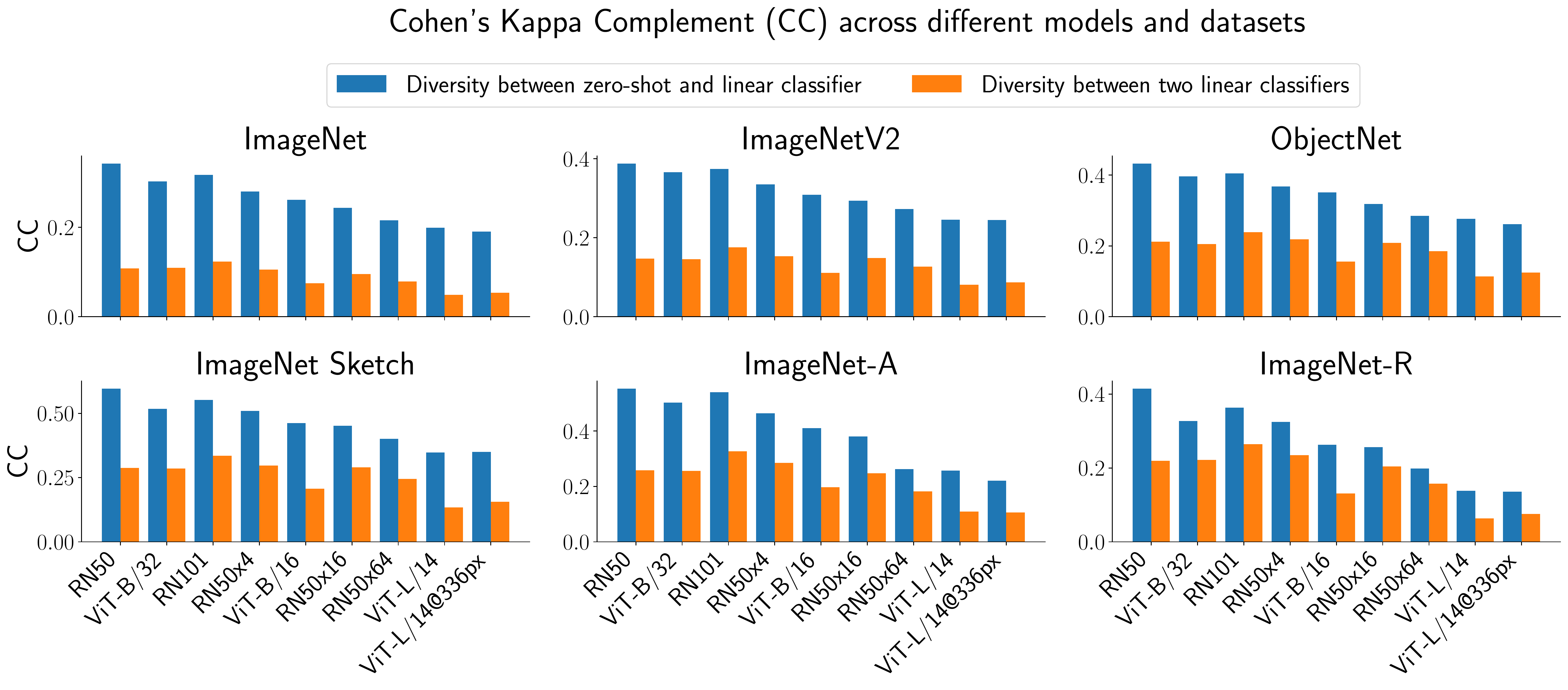}
    \caption{\textbf{Cohen's Kappa Complement (CC)} for multiple datasets and CLIP models (Equation \ref {eq:cc}).}
\end{figure*}

\begin{figure*}
    \centering
    \includegraphics[width=\textwidth]{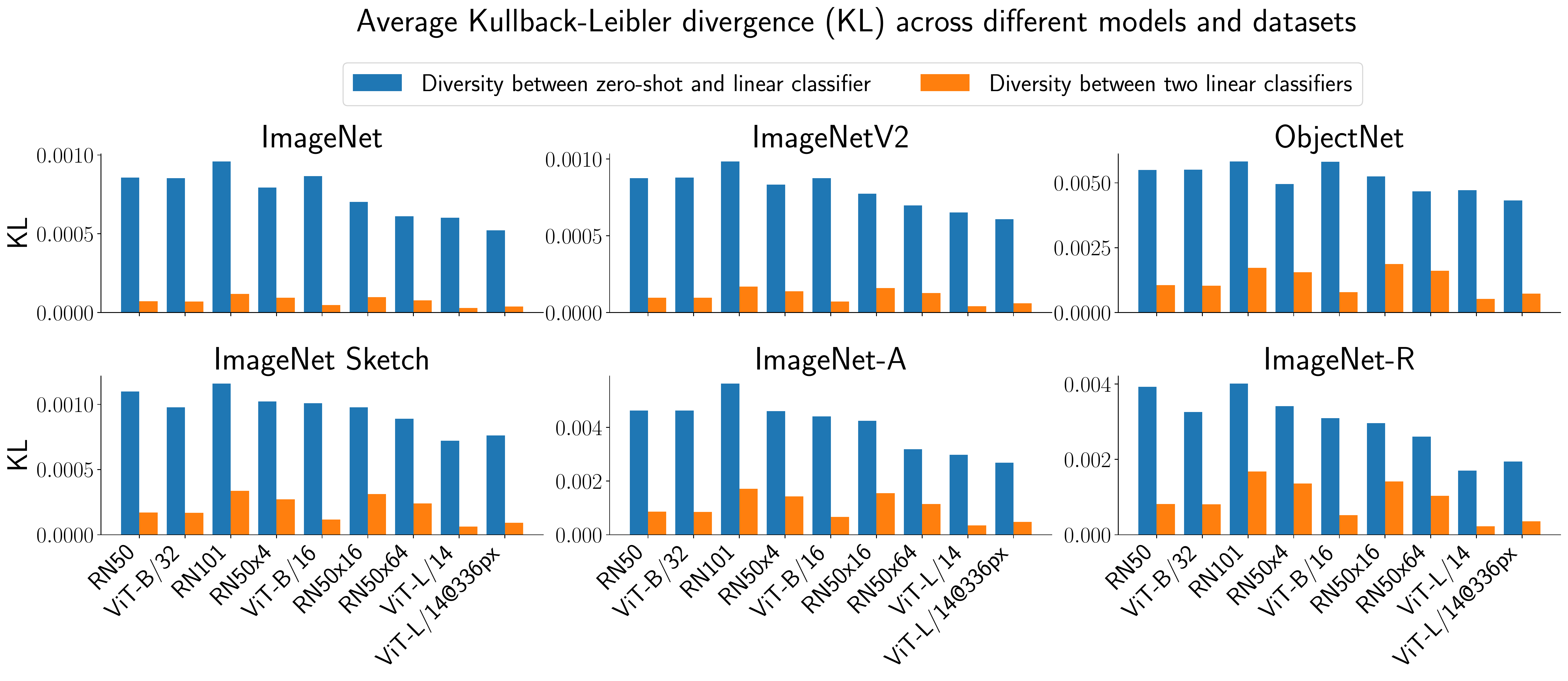}
    \caption{\textbf{Average KL Divergence (KL)} for multiple datasets and CLIP models (Equation \ref {eq:kl}).}
\end{figure*}

\begin{figure*}
    \centering
    \includegraphics[width=\textwidth]{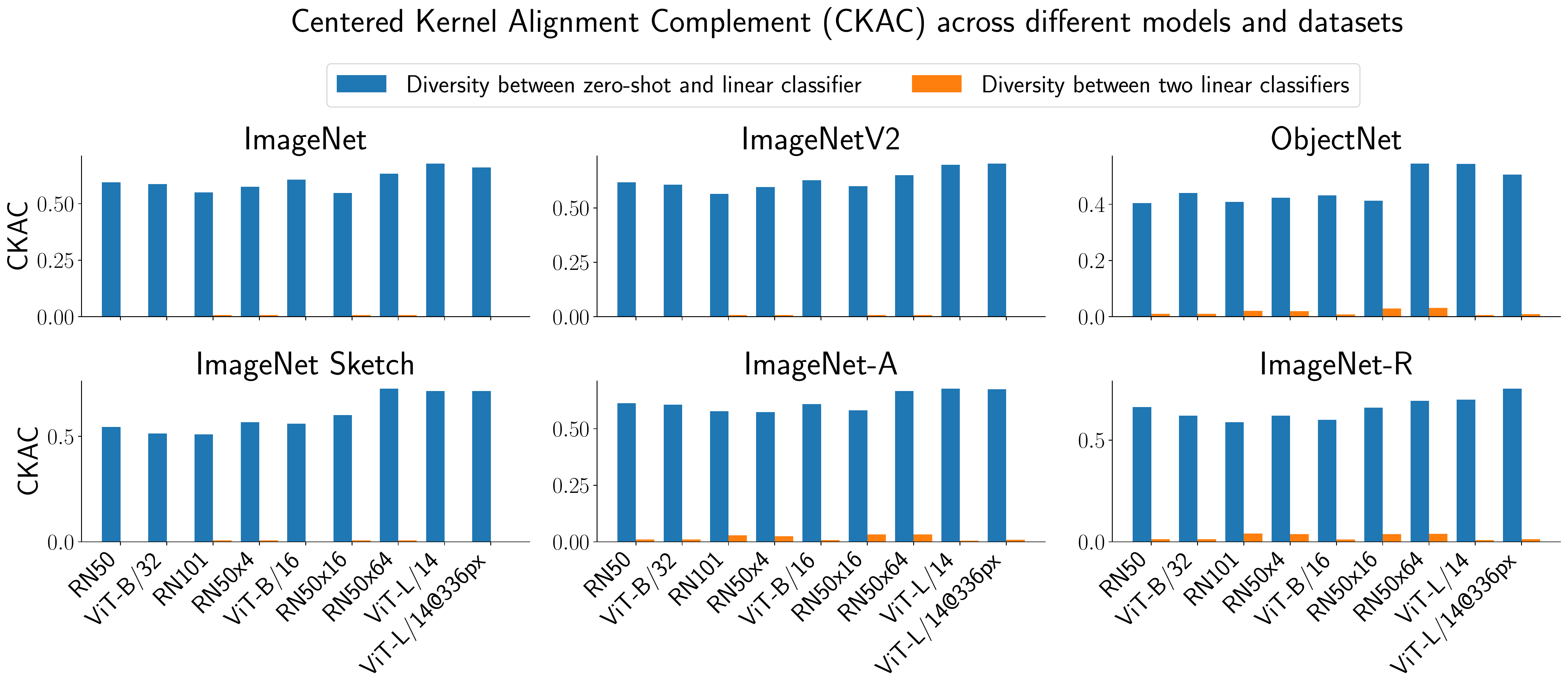}
    \caption{\textbf{Central Kernel Alignment Complement (CKAC)} for multiple datasets and CLIP models (Equation \ref {eq:ckac}).}
    \label{fig:diversity_breakdown_end}
\end{figure*}

\FloatBarrier

\section{When do weight-space ensembles approximate output-space ensembles?}\label{sec:ntk}

In practice we observe a difference between weight-space and output-space ensembling. 
However, it is worth noting that these two methods of ensembling are not as different as they initially appear. In certain regimes a weight-space ensemble approximates the corresponding output-space ensemble---for instance, when training is well approximated by a linear expansion, referred to as the 
NTK regime \cite{jacot2018neural}. \citet{fort2020deep} find that a linear expansion becomes more accurate in the later phase of neural network training, a phase which closely resembles fine-tuning.

Consider the set $\Theta = \mleft\{ (1-\alpha) \theta_0 + \alpha \theta_1 : \alpha \in [0,1] \mright\}$ consisting of all $\theta$ which lie on the linear path between $\theta_0$ and $\theta_1$.

\textbf{Proposition 1.} When $f(\theta) = f(\theta_0) + \nabla f(\theta_0)^\top ( \theta - \theta_0)$ for all $\theta \in \Theta$, the weight- and output-space ensemble of $\theta_0$ and $\theta_1$ are equivalent.

\textit{Proof.} We may begin with the weight-space ensemble and retrieve the output-space ensemble
\begin{align}
    &f\mleft((1-\alpha) \theta_0 + \alpha \theta_1 \mright) \\
    &= f\mleft(\theta_0\mright) + \nabla f\mleft(\theta_0\mright)^\top \mleft( (1-\alpha) \theta_0 + \alpha \theta_1 - \theta_0 \mright) \\
    &= f\mleft(\theta_0\mright) + \alpha \nabla f\mleft(\theta_0\mright)^\top \mleft( \theta_1 - \theta_0 \mright) \\
    &= f\mleft(\theta_0\mright) + \alpha \nabla f\mleft(\theta_0\mright)^\top \mleft( \theta_1 - \theta_0 \mright) + \alpha f\mleft(\theta_0\mright) - \alpha f\mleft(\theta_0\mright) \\
    &= (1-\alpha) f\mleft(\theta_0\mright) + \alpha \mleft(f\mleft(\theta_0\mright) +  \nabla f\mleft(\theta_0\mright)^\top \mleft( \theta_1 - \theta_0 \mright) \mright) \\
    &= (1-\alpha) f\mleft(\theta_0\mright) + \alpha f\mleft(\theta_1 \mright)
\end{align}
where the first and final line follow by the linearity assumption. \qed

\end{document}